\documentclass[10pt,journal,letterpaper,twocolumn]{IEEEtran}

\usepackage[utf8]{inputenc} 
\usepackage[T1]{fontenc}    
\usepackage{hyperref}       
\usepackage{url}            
\usepackage{booktabs}       
\usepackage{amsfonts}       
\usepackage{nicefrac}       
\usepackage{microtype}      
\usepackage{graphicx}
\usepackage{multirow}
\usepackage{subfigure}
\usepackage[ruled]{algorithm2e}
\usepackage{epsfig}
\usepackage{color}
\usepackage{amsthm}
\usepackage{amsmath}
\usepackage{amssymb}

\usepackage{caption2}

\usepackage{footmisc}

\usepackage{array}

\usepackage{epstopdf}

\def\etal{\emph{et al. }}

\def\ie{\emph{i.e.}}

\def\V05{\hspace{-0.075in}}
\def\VV05{\vspace{-0.01in}}

\newtheorem{theorem}{Theorem}

\newtheorem{property}{Property}

\title{ A Novel Tensor Factorization-Based Method with Robustness to   Inaccurate Rank Estimation}

\author{Jingjing Zheng$^\dagger$, Wenzhe Wang$^\dagger$, Xiaoqin Zhang, 
 and Xianta Jiang  

\IEEEcompsocitemizethanks{

\IEEEcompsocthanksitem
(Corresponding author: Xiaoqing Zhang.) 
\IEEEcompsocthanksitem
 J. Zheng, W. Wang and X. Zhang are with the College of Computer Science and Artificial Intelligence, Wenzhou University, Zhejiang, China (e-mail:
		jjzheng233@gmail.com,  woden3702@gmail.com, zhangxiaoqinnan@gmail.com, ).
\IEEEcompsocthanksitem
J. Zheng and X. Jiang are with the Department of Computer Science, Memorial University of Newfoundland, Newfoundland and Labrador, Canada (e-mail: jjzheng233@gmail.com, xiantaj@mun.ca).
\IEEEcompsocthanksitem
$\dagger$ These authors contributed equally to this work.
}
}

\begin{document}

\maketitle


\begin{abstract}
This study aims to solve the over-reliance on the rank estimation strategy in the standard tensor factorization-based tensor recovery and the problem of a large computational cost in the standard t-SVD-based tensor recovery. To this end, we proposes a new tensor norm with a dual low-rank constraint, which utilizes the low-rank prior and  rank information at the same time. In the proposed tensor norm, a series of surrogate functions of the tensor tubal rank can be used to achieve better performance in harness low-rankness within tensor data. It is proven theoretically that the resulting tensor completion model can effectively avoid performance degradation caused by inaccurate rank estimation. Meanwhile, attributed to the proposed dual low-rank constraint, the t-SVD of a smaller tensor instead of the original big one is computed by using a sample trick. Based on this, the total cost at each iteration of the optimization algorithm  is reduced to $\mathcal{O}(n^3\log n +kn^3)$ from $\mathcal{O}(n^4)$ achieved with standard methods, where $k$ is the estimation of the true tensor rank and far less than $n$. Our method was evaluated on synthetic and real-world data, and it demonstrated superior performance and efficiency over several existing state-of-the-art tensor completion methods.

\end{abstract}
\section{Introduction}\label{section:Introduction}

In recent years,  owing to the dramatic advance in hardware, massive high-dimensional data (e.g., images, videos, hyper-spectral data, and 3-D range data) are available \cite{1} and  represented as  tensors. Lots of low-rank tensor recovery  methods have  been developed to exploit  low-dimensional structures in these tensor data and  widely used in the color images and videos denoising \cite{34}, image inpainting \cite{wu2018fused,shi2018feature},  video background modeling \cite{candes2011robust,zhao2015bayesian,34}, background initialization \cite{sobral2017matrix,kajo2018svd,kajo2019self, madathil2018twist}, and hyperspectral image restoration \cite{gandy2011tensor}.   According to various tensor rank \cite{Koldakiers2000towards, kolda2009tensor} \cite{kolda2009tensor,12,zhang2019robust}\cite{liu2016low,guichardet2006symmetric}, many tensor completion methods can be derived.

\begin{table}
	\renewcommand\arraystretch{1}
	\centering
	\caption{The surrogate functions of $\ell_{0}$ norm, where $\gamma > 0$. }
	\begin{tabular}{c|c}\hline\hline
		Name          &      $g(x)$     \\\hline  \hline
		$\ell_p $    \cite{frank1993statistical} &     $  x^{p} $,               $  0< p < 1 $      \\ \hline
		Geman    \cite{geman1992constrained}        &      $ \frac{x}{x+\gamma}$           \\  \hline
		Laplace    \cite{trzasko2008highly}     &      $  (1-\exp(-\frac{x}{\gamma}))$            \\\hline
		LOG  \cite{Malioutov2013Iterative} &  $ \log(\gamma+x)$ \\\hline
		Logarithm \cite{friedman2012fast} &        $\frac{1}{\log(\gamma + 1)} \log (\gamma x +1)$    \\\hline                  
		ETP  \cite{gao2011feasible}	&        $   \frac{1-\exp(-\gamma x)}{1-\exp(-\gamma)} $  \\\hline
	\end{tabular}
	\label{tab1}
\end{table}

Recently,   the convex hull of t-product-based rank, the tensor nuclear norm (TNN)  has attracted much attention because of its effectiveness in tensor recovery  \cite{22, 37, 34}. The resulting TNN-based models can exactly recover the true value under some conditions, as stated in \cite{22,37}. However, the conditions are difficult to be satisfied in practice. In addition, since the information of the data is concentrated in the components corresponding to a few largest singular values \cite{shang2017bilinear,34}, the larger singular values should be penalized mildly. Nevertheless, the TNN-based methods treat the singular values with an equal penalty, leading to the over-punishment for large singular values and therefore performance degradation. 
The non-convex surrogate functions as listed in Table \ref{tab1}  have been extensively used in the field of low-rank recovery.
Since these methods need to compute t-SVD in each iteration, they incur a large computation cost and cannot process large-scale tensor data efficiently.   Zhou \etal \cite{29} have proposed Tensor Completion by Tensor Factorization (TCTF) that reduces running time significantly when the given  data is  low rank. 

Although the t-product-based tensor completion methods have achieved great success, there are still two challenges:
(1) The TCTF \cite{29} is based on a basic hypothesis that the tensor with tensor tubal rank $k$ can be approximately decomposed to the t-product of two skinny tensors $\mathcal{A} \in \mathbb{R}^{n_1 \times k \times n_3}$ and $\mathcal{B}  \in \mathbb{R}^{k \times n_2 \times n_3}$ ($k$ could be obtained by  estimating  the  tensor tubal rank), so it overly rely on the rank estimation strategy. On the other hand, due to the lack of a rank-increasing scheme, the rank estimation strategy given in \cite{29} often underestimates the true rank, causing performance degradation in TCTF.  
(2) Currently, iterative re-weighted tensor nuclear norm algorithms \cite{wang2021generalized,8913528} and generalized tensor singular value thresholding \cite{Zhang2022Tensor} were proposed to solve the  non-convex approximation of tensor recovery. These algorithms require computing the t-SVD of the original large tensor in each iteration, causing a high computational cost. 
Therefore, it is necessary to develop an efficient and effective tensor recovery framework for a wide range of surrogate functions.

To address the issues mentioned above, this paper proposes  a novel tensor completion framework with a new tensor norm that utilizes a dual
low-rank constraint (TCDLR), and its goal is to avoid the high computational cost in the standard t-SVD-based method and achieve superior recovery results. More specifically, the features of the proposed method are as follows: 
\begin{itemize} 
	\item First, a new tensor norm with a dual low-rank constraint is given to utilize  the low-rank prior and the true tubal-rank information at the same time. Based on the proposed tensor norm, a series of surrogate functions (possibly non-convex) of the tensor tubal rank in the resulting tensor completion model (TCDLR) are allowed to achieve better performance in harnessing low-rankness within the tensor data and solving the over-punishment problem in the TNN-based tensor completion.  Besides, Property \ref{Pn1}, Theorem \ref{th1}, and synthetic experiments confirm that TCDLR can be less negatively affected by the mis-estimation of tensor rank compared with standard tensor factorization-based methods. 

 
  
	\item Second, an optimization algorithm is developed to solve TCDLR efficiently, in which the t-SVD of a smaller tensor instead of the big one is computed by using a simple trick. As a result, the total cost at each iteration of the developing algorithm is reduced to $\mathcal{O}(n_1n_2n_3\log n_3 +kn_1n_2n_3)$ from $\mathcal{O}(n_{(1)}n_{(2)}^2n_3+n_1n_2n_3\log n_3)$ achieved with standard t-SVD-based methods, where $n_{(1)}=\max(n_1,n_2)$ and $n_{(2)}=\min(n_1,n_2)$. Here, $k\ll n_{(2)}$ is the estimation of tensor rank. The convergence of the optimization algorithm was analyzed experimentally.


\item Third, since the   tensor rank of real tensor data is unknown, a novel rank estimation method is proposed, which adopts an increasing and decreasing strategy to estimate the tensor rank more precisely. By combining TCDLR with the proposed rank estimation method, an efficient tensor completion framework (TCDLR-RE) is established to accurately and effectively recover the principal components (the low-rank tensor). The experiments  demonstrate the high efficiency and effectiveness of TCDLR-RE.

\end{itemize}


\section{Notations, Preliminaries, and Related Works} 
\subsection{Notations and Preliminaries}

\begin{table}[]
	\centering
	\caption{Notations}\label{notation}
	\vspace{0.1cm}
	\renewcommand\arraystretch{1.5}
	\scalebox{0.49}{
		\resizebox{\textwidth}{!}{%
			\begin{tabular}{|c|c|c|c|}
				\hline
				Notations                                          & Descriptions                                                                                                                       & Notations                                      & Descriptions                                                      \\ \hline
				$\mathbb{R}$                                       & real field                                                                   &                    	$ \mathbb{C}  $                                    & complex field         \\       
 \hline    $ a $    &  scalar  &	         $  \textbf{a} $                                    & vector            \\ \hline
	  	$A $                                     & matrix                 &       $ \mathcal{A} $                                & tensor 	                                 \\\hline
       	 $\textbf{A} $                                   & sets   &	$ \textbf{a}_i,  a_i $                                   & $ i $-th element of $  \textbf{a} $   \\\hline 
		$ \sigma_i( \mathrm{A}) $                          & 
    $ i $-th singular value of matrix $ \mathrm{A} $ 	   &                             	$ \sigma(\mathrm{A}) $                             & $ (\sigma_1( \mathrm{A}),\sigma_2( \mathrm{A}),\dots,\sigma_r( \mathrm{A}))^T$                            \\ \hline
    	   	$ \operatorname{Conj}(A) $                         & conjugate of $A$  &    $ \textbf{0}_{m\times n} $                         & $m\times n $ null matrix   \\ \hline
         $\mathcal{I}_{n \times n  \times n_3} $              & $n \times n  \times n_3$ identity tensor           &                         $ \mathcal{A}^* $                                   & conjugate transpose of $ \mathcal{A}$       	                                            \\ \hline
			                                 $[\mathcal{A}]_{i, j, k}$                        & $ (i,j,k) $-th element in $ \mathcal{A} $                                                                 & $ [\mathcal{A}]_{i,j,:} $                        & $ (i,j) $-th tube                                                 \\ \hline 
	 
		   $ [\mathcal{A}]_{:,:,i} $     & $ i $-th frontal slice   &     $\mathcal{A}^{\dag} $                         & pseudo-inverse of $ \mathcal{A} $                                             \\ \hline
			      $ \mathcal{\bar A} $                           &  $ \mathrm{fft}(\mathcal{A},[],3) $                                                &        $\bar A$                      &     $ \mathrm{bdiag}(\mathcal{\bar A})$           \\ \hline
		    $ \mathrm{\bar A}_{i} $                        & $ i $-th frontal slice of $ \mathcal{\bar A} $        	&                           $\#\textbf{A}$  	&number of elements of  $ \textbf{A} $                    \\ \hline
			\end{tabular}
			
	}}
\vspace{-0.4cm}
\end{table}

The notations used throughout this paper are listed in Table \ref{notation}. 
Specifically, $ \mathcal{\bar A}\in \mathbb{C}^{n_1\times n_2\times n_3} $ represents the resulting tensor by applying the Discrete Fourier Transformation    on all tubes of tensor $ \mathcal{A} $, and 
its inverse transform  can be obtained by  the Matlab command $ \mathcal{A}=\mathrm{ifft}(\bar{\mathcal{A}},[], 3). $  Besides, $ \mathrm{bdiag}(\mathcal{A}) $ is   defined as follows: 
\[ \mathrm{bdiag}(\mathcal{A})=\left(
\begin{array}{cccc}
	\left[\mathcal{A}\right]_{:,:,1}& 	\textbf{0}_{n_1\times n_2}  & \cdots & 	\textbf{0}_{n_1\times n_2}  \\
	\textbf{0}_{n_1\times n_2} &  \left[\mathcal{A}\right]_{:,:,2} & \cdots&  	\textbf{0}_{n_1\times n_2}\\
	\vdots  & \vdots & \ddots & \vdots \\
	\textbf{0}_{n_1\times n_2}  & 	\textbf{0}_{n_1\times n_2} & \cdots &  	 \left[\mathcal{A}\right]_{:,:,n_3}  \\
\end{array}                               \right).\] 

Given   $\mathcal{A}\in \mathbb{R}^{n_1\times n_2 \times n_3}$ and $\mathcal{B}\in \mathbb{R}^{n_2\times l \times n_3 }$, the $\mathrm{t}$-product $\mathcal{A}\ast\mathcal{B} \in \mathbb{R}^{n_1\times l \times n_3}$ \cite{18} is defined as \begin{equation}
		\mathcal{A}\ast\mathcal{B}=\mathrm{fold}(\mathrm{bcirc}(\mathcal{A})\cdot \mathrm{unfold}(\mathcal{B})).
	\end{equation}
Here, $	\operatorname{fold}(\cdot)$ is the  inverse operator 
 of $\mathrm{unfold}(\cdot)$, $ \mathrm{bcirc} (\mathcal{A}) \in \mathbb{R}^{n_1n_3\times n_2n_3}$ and  $\mathrm{unfold}(\mathcal{B})\in \mathbb{R}^{n_1n_3\times n_2}$ are the  matrices given by 
  \[\mathrm{bcirc}(\mathcal{A})=\left(
\begin{array}{cccc}
	\left[\mathcal{A}\right]_{:,:,1}& \left[\mathcal{A}\right]_{:,:,n_3}  & \cdots & \left[\mathcal{A}\right]_{:,:,2}  \\
	\left[\mathcal{A}\right]_{:,:,2} &  \left[\mathcal{A}\right]_{:,:,1}  & \cdots&  \left[\mathcal{A}\right]_{:,:,3} \\
	\vdots  & \vdots & \ddots & \vdots \\
	\left[\mathcal{A}\right]_{:,:,n_3}  & \left[\mathcal{A}\right]_{:,:,n_3-1}   & \cdots &  	 \left[\mathcal{A}\right]_{:,:,1}  \\
\end{array}                               \right),	\]
and 	\[
\mathrm{unfold}(\mathcal{B})=\left(
\begin{array}{c}
	\left[\mathcal{B}\right]_{:,:,1} \\
	\left[\mathcal{B}\right]_{:,:,2} \\
	\vdots \\
	\left[\mathcal{B}\right]_{:,:,n_3} \\
\end{array}
\right)\in \mathbb{R}^{n_1n_3  \times n_2},  \] respectively.





  Let $
	\mathcal{A}=\mathcal{U} \ast \mathcal{S} \ast \mathcal{V}^*	$ be  the t-SVD of $\mathcal{A}$  \cite{34}. The tensor tubal rank \cite{kilmer2013third,21} and the tensor nuclear norm \cite{34} of $\mathcal{A}$ are defined as 	$\mathrm{rank}(\mathcal{A})=\|[\mathcal{S}]_{:,:,1}\|_0$
       and $\|\mathcal{A}\|_*=\sum_{i=1}^{\mathrm{rank}(\mathcal{X})}[\mathcal{S}]_{i,i,1}$, respectively.

\subsection{Related Works}
 
As mentioned in \cite{34}, compared with the CP rank and the tucker rank,  the low tensor tubal rank can be more easily satisfied in practice. The examples given in \cite{34,29} show the low  rank  property of tensor data.
 
Therefore,  the  tensor completion model is formulated as
 \begin{align}\label{TC}
	& \min_{ \mathcal{X}   }\mathrm{rank}(\mathcal{X})\notag\\
	&s.t.~  \mathbf{P}_{\Omega}(\mathcal{M})=\mathbf{P}_{\Omega}(\mathcal{X}),
\end{align}
where     $\mathbf{P}_{\Omega}$ is a linear project operator on the support set $\Omega$ composed of the locations corresponding to the observed entries in $\mathcal{M}$.
Since \eqref{TC} is NP-hard, a common method to solve it is using the convex envelope of tensor tubal rank (\ie,TNN) to replace $\mathrm{rank}(\mathcal{X})$. As a result, a TNN-based tensor completion \cite{22,37} model is  given as 
\begin{equation}\label{TC_tnn}
	 \min_{\mathcal{X}}\|\mathcal{X}\|_{*} ~~~~~~
	s.t.~  \mathbf{P}_{\Omega}(\mathcal{X})=\mathbf{P}_{\Omega}(\mathcal{M}).
\end{equation} 


Although the exact recovery of TNN-based tensor completion methods is guaranteed in theory, the conditions for the exact recovery are difficult to meet in practice.   As a better approximation than the tensor tubal rank \cite{wang2021generalized}, $\|\cdot\|_{*,g}$ has been extensively studied \cite{jiang2020multi,wang2021generalized,kong2018t,xu2019laplace,8913528}, where $\|\mathcal{X}\|_{*,g}=\frac{1}{n_3}\sum_{i=1}^{r}g(\sigma_i(\bar{X}))$, $g$ is an increasing non-negative function, and $r=\mathrm{rank}(\bar{X})$.   
Kong \etal \cite{kong2018t} proposed the tensor Schatten-$p$ norm (defined as $\|\mathcal{X}\|^p_{S_p}=\frac{1}{n_3}\sum_{i=1}^{\min(n_1,n_2)n_3}(\sigma_i(\bar{X}))^p$ ($0<p\leq 1$) to  achieve better tensor completion performance. 
 Kong \etal proved 
\begin{equation}\label{lpDe}
\|\mathcal{X}\|^p_{S_p}=\min_{\mathcal{X}_i:\mathcal{X}=\mathcal{X}_1\ast\cdots \ast\mathcal{X}_I}\sum_{i=1}^I
\frac{1}{p_i}\|\mathcal{X}_i\|^{p_i}_{S_{p_i}} 
\end{equation}
if $\frac{1}{p}=\sum_{i=1}^I
\frac{1}{p_i}$ ($p_i$ can be bigger than 1.), $\mathcal{X}_1 \in \mathbb{R}^{n_1\times d_1 \times n_3}$, $\mathcal{X}_i \in \mathbb{R}^{d_{i-1}\times d_i \times n_3}$  $(i=2,\cdots,I-1)$, and $\mathcal{X}_I \in \mathbb{R}^{d_I\times n_2 \times n_3}$ for $\mathrm{rank}(\mathcal{X})\leq \min\{d_i,i=1,\cdots, I\}$,  
which reduces the computational complexity to $\mathcal{O}(n_1n_2n_3\log n_3 +\min\{d_i,i=1,\cdots, I\}n_1n_2n_3)$. 
It is worth noting that the decomposition given in \eqref{lpDe} depends on the true tubal-rank information $\mathrm{rank}(\mathcal{X})$, which is ignored in \cite{kong2018t} and will cause performance degradation.
 Jiang \etal given a new non-convex approximation called the partial sum of the TNN (PSTNN) to explore the low-rank structure in  the data effectively \cite{jiang2020multi}.  Xu \etal \cite{xu2019laplace} given a  Laplace function-based non-convex surrogate strategy, in which the weight for each singular value is updated adaptively.
The basic idea of these non-convex-based methods is to replace the tensor nuclear norm with its non-convex surrogate function, which is summarized  below:
\begin{equation}\label{TClg}
 \min_{\mathcal{X}}\|\mathcal{X}\|_{*,g} ~~~~~~
	s.t.~  \mathbf{P}_{\Omega}(\mathcal{X})=\mathbf{P}_{\Omega}(\mathcal{M}).
\end{equation}

Iterative re-weighted tensor nuclear norm algorithms \cite{8913528,wang2021generalized} were proposed to solve the  generalized non-convex problem  \eqref{TClg}.  Since these  algorithms require  computing the t-SVD of a tensor of size $n_1\times n_2 \times n_3$, they incur a computational cost of $\mathcal{O}(n_{(1)}n_{(2)}^2n_3+n_1n_2n_3\log n_3)$ and cannot be used to process large scale tensor data efficiently.
	Therefore,   TCTF has been proposed:
\begin{align}\label{TCTF}
	& \min_{\mathcal{A} , \mathcal{B}, \mathcal{X} }\|\mathcal{A}\ast\mathcal{B}-\mathcal{X}\|_F\notag\\
	&s.t.~  \mathbf{P}_{\Omega}(\mathcal{M})=\mathbf{P}_{\Omega}(\mathcal{X}), 
\end{align}
where $\mathcal{A}\in \mathbb{R}^{n_1\times k \times n_3}$, $\mathcal{B}\in \mathbb{R}^{k\times  n_2\times n_3 }$, and $k$ obtained by the rank estimation strategy \cite{29} are the estimation for the tensor tubal rank of $\mathcal{M}$. The total computational costs at each iteration of TCTF is $\mathcal{O}(kn_{1}n_{2}n_3+n_1n_2n_3\log n_3)$, thus achieving a significant performance improvement for the case of $k\ll n_{(2)}$. 
However, due to the lack of a rank-increasing scheme and the over-reliance on  the rank estimation strategy, TCTF often suffers from degraded recovery accuracy.


In \cite{shi2021robust}, Shi \etal given a new rank estimation strategy that estimates the tensor tubal rank more accurately by counting the   non-zero tubal in $\hat{\mathcal{S}}$, where  $\hat{\mathcal{S}}$ is obtained by solving
\begin{align}\label{ShiRe}
	(\hat{\mathcal{S}}, \hat{\mathcal{U}}, \hat{\mathcal{V}},\hat{\mathcal{X}})=& \mathop{\arg\min}_{\mathcal{S},  \mathcal{U}, \mathcal{V}, \mathcal{X}} \alpha\|\mathcal{S}\|_*+\frac{1}{2}\|\mathcal{U}\ast\mathcal{S}\ast\mathcal{V}^T-\mathcal{X}\|_F^2   \notag\\
	&s.t.~  \mathbf{P}_{\Omega}(\mathcal{M})=\mathbf{P}_{\Omega}(\mathcal{X}),\mathcal{I}=\mathcal{U}^T\ast \mathcal{U},\notag\\
 &~~~~\mathcal{I}=\mathcal{V}^T\ast \mathcal{V}. 
\end{align} 
Then, they recover the low-rank tensor   by solving a tensor factorization problem.

\section{Proposed Tensor Completion Framework}

 
\subsection{Formulation of TCDLR}



Although TCTF can well address the issue of high computational complexity caused by t-SVD for large tensors at each iteration, its over-reliance on the rank estimation strategy often leads to degraded recovery accuracy. According to Property \ref{Pn1} (ii), the low-rank estimation of TCTF and TC-RE  will deviate significantly from the true value when the rank estimation deviate from the true rank. Even worse, the true rank is difficult to be estimated accurately, especially under a low sampling rate. Therefore, in addition to achieving a better rank estimation, a new effective tensor completion model needs to be developed.

\begin{property}\label{Pn1} For $\mathcal{Y}, \mathcal{X}\in \mathbb{R}^{n_1 \times n_2 \times n_3}$, then 
\begin{itemize}
\item[(i)] $\|\mathcal{Y}-\mathcal{X}\|^2_F\geq \frac{1}{n_3}\sum_i(\sigma_i(\bar{Y})-\sigma_i(\bar{X}))^2$;
    
    \item[(ii)] $\|\mathcal{Y}-\mathcal{X}\|^2_F\geq \frac{1}{n_3}\sum_{i=1}^{n_3}  \sum_{\mathrm{rank}(\mathcal{Y})<j\leq \mathrm{rank}(\mathcal{X})}\sigma_j(\bar{X}_i)^2$ if $\mathrm{rank}(\mathcal{Y})<\mathrm{rank}(\mathcal{X})$.
\end{itemize}
\end{property}

The fundamental idea of our approach is to calculate the t-SVD of the obtained smaller tensor instead of the original tensor by utilizing tensor tubal rank information that could be provided by rank estimation methods, thus reducing the computational complexity of the original t-SVD-based methods.  
 Therefore, a new tensor norm is introduced, 
\begin{equation}
{\left\| {\mathcal{X}} \right\|_{*,(k,g)}} = \left\{ \begin{array}{l}
	\infty , ~~~~~~~  \text{if} ~ \mathrm{rank}(\mathcal{X})>k;\\
	{\left\| {\mathcal{X}} \right\|_{*,g}}, ~~ \text{if} ~ \mathrm{rank}(\mathcal{X}) \leq k,
\end{array} \right.
\end{equation} 
and the following generalized framework (TCDLR) is established: 
\begin{align}\label{TCDLR}
	& \min_{ \mathcal{X}  }\|\mathcal{X}\|_{*,(k,g)}\notag\\
	&s.t.~  \mathbf{P}_{\Omega}(\mathcal{X})=\mathbf{P}_{\Omega}(\mathcal{M}),
\end{align} 
where $g$ can be any of the surrogate functions listed in Table \ref{tab1}. 
By minimizing  $\|\mathcal{X}\|_{*,(k,g)}$, both the low-rank prior and tensor tubal rank information $\mathrm{rank}(\mathcal{X})\leq k$ are considered. Theorem \ref{th1} shows the 
robustness of TCDLR to inaccurate rank estimations for $k$. Noting that since $\|\cdot\|_{*,g}$ is a better approximation of the tensor tubal rank than the tensor nuclear norm, the optimal solution to \eqref{TClg} is expected to be  low rank. When $\mathrm{rank}(\widetilde{\mathcal{X}}) \leq k <\mathrm{rank}(\mathring{\mathcal{X}})$, the estimation $k$ provides a better prior for the true tensor tubal rank, which helps to recover the low-rank tensor more accurately. Later, the effectiveness of TCDLR in tensor recovery will be demonstrated by experiments  further.



 \begin{theorem}\label{th1}
Let $ \mathring{\mathcal{X}}$,  $\hat{\mathcal{X}}$  and $ \widetilde{\mathcal{X}}$ be the optimal solutions to \eqref{TClg}, \eqref{TCDLR},  and \eqref{TC}, respectively.  
 Then, we have: 
 \begin{itemize}

      \item[(i)]  $\mathrm{rank}(\mathring{\mathcal{X}})\geq \mathrm{rank}(\widetilde{\mathcal{X}})$ holds;
     \item[(ii)] If $k \geq \mathrm{rank}(\mathring{\mathcal{X}})$,  $\hat{\mathcal{X}}$ is an optimal solution of \eqref{TClg} and  $\mathring{\mathcal{X}}$ is an optimal solution of \eqref{TCDLR};
            \item[(iii)] If $k=\mathrm{rank}(\widetilde{\mathcal{X}})$,      $\hat{\mathcal{X}}$ is an optimal solution of \eqref{TC}; 
            \item[(iv)]   $k<\mathrm{rank}(\widetilde{\mathcal{X}})$ holds if and only if    $\|\hat{\mathcal{X}}\|_{*,(k,g)}=\infty$.
             \end{itemize}
\end{theorem}

     As it will be shown later, benefiting from the  proposed dual low-rank constraint, TCDLR can avoid performing the t-SVD operation for the original bigger tensor that causes high time consumption. 
 



\subsection{The Developing Optimization Algorithm for Solving TCDLR}
\subsubsection{A Trick to Efficiently Solve TCDLR}

Since there exist $\mathcal{A}\in \mathbb{R}^{n_1\times k \times n_3}$ and  
 $\mathcal{B}\in \mathbb{R}^{k\times  n_2\times n_3 }$ such that $\mathcal{X}=\mathcal{A}\ast \mathcal{B}$ and $\|\mathcal{X}\|_{*,(k,g)}=\|\mathcal{A}\ast\mathcal{B}\|_{*,g}$ when $\mathrm{rank}(\mathcal{X})\leq k$, the following problem is considered:
\begin{align}\label{AB}
	& \min_{ \mathcal{A}, \mathcal{B}  }\|\mathcal{A}\ast\mathcal{B}\|_{*,g}\notag\\
	&s.t.~  \mathbf{P}_{\Omega}(\mathcal{M})=\mathbf{P}_{\Omega}(\mathcal{A}\ast\mathcal{B}),
\end{align}
 where $\mathcal{A}\in \mathbb{R}^{n_1\times k \times n_3}$ and  
 $\mathcal{B}\in \mathbb{R}^{k\times  n_2\times n_3 }$.
If $(\hat{\mathcal{A}},\hat{\mathcal{B}})$ is an optimal solution to \eqref{AB}, we have $ \|\hat{\mathcal{A}}\ast\hat{\mathcal{B}}\|_{*,(k,g)}=\|\hat{\mathcal{A}}\ast\hat{\mathcal{B}}\|_{*,g} \leq \|\hat{\mathcal{X}}\|_{*,(k,g)}$, where $\hat{\mathcal{X}}$ is the optimal solution to \eqref{TCDLR}. Thus, it can be concluded that $\hat{\mathcal{A}}\ast\hat{\mathcal{B}}$ is an optimal solution to \eqref{TCDLR}.

\begin{theorem}\label{th2}
Let $ \mathcal{Y}=\mathcal{A}\ast\mathcal{B}$, where $ \mathcal{A} \in \mathbb{R}^{n_1 \times k  \times n_3} $ and $ \mathcal{B} \in \mathbb{R}^{k \times n_2  \times n_3}$.
If $  \mathcal{B}^*=\mathcal{Q}^*\ast\mathcal{R}^*$ is  the QR decomposition  of $\mathcal{B}^*$ and $\mathcal{Z}=\mathcal{A}\ast \mathcal{R}$,  $$ \mathcal{Y}=\mathcal{Z}\ast\mathcal{Q}$$ and
$$\mathbb{S}_{\tau,g}(\mathcal{Y})=\mathbb{S}_{\tau,g}(\mathcal{Z})\ast\mathcal{Q}$$ hold,  where $ \mathcal{Z} \in \mathbb{R}^{n_1 \times k  \times n_3} $, $ \mathcal{Q} \in \mathbb{R}^{k \times n_2  \times n_3}$, and  
$$ \mathbb{S}_{\tau,g}(\mathcal{Y})=\mathop{\arg\min}_{\mathcal{X}} \frac{1}{2}\|\mathcal{Y}-\mathcal{X}\|_F^2+\tau\|\mathcal{X}\|_{*,g}.$$ 
\end{theorem} 

According to Theorem \ref{th2}, the t-SVD of the large tensor can be avoided by constructing $\mathbb{S}_{\tau,g}(\mathcal{\mathcal{A}\ast\mathcal{B}})$. To achieve this goal, this paper introduces an auxiliary tensor $\mathcal{X}$ such that $\mathcal{X}=\mathcal{A}\ast\mathcal{B}$. Meanwhile, \eqref{AB} is convert to 
 \begin{align}\label{reinipro2}
	& \min_{ \mathcal{A}, \mathcal{B}, \mathcal{X}} \|\mathcal{X}\|_{*,g} \notag\\
	&s.t.~  \mathbf{P}_{\Omega}(\mathcal{M})=\mathbf{P}_{\Omega}(\mathcal{A}\ast\mathcal{B}),\notag\\
 &~~~~~ \mathcal{X}=\mathcal{A}\ast\mathcal{B},
\end{align} 
which is equivalent to \eqref{reinipro1} when $\mu=+\infty$. 
 \begin{align}\label{reinipro1}
	& \min_{ \mathcal{A}, \mathcal{B}, \mathcal{X}} \|\mathcal{X}\|_{*,g}+ \dfrac{\mu}{2}\|\mathcal{A}\ast\mathcal{B}-\mathcal{X}\|^2_F \notag\\
	&s.t.~  \mathbf{P}_{\Omega}(\mathcal{M})=\mathbf{P}_{\Omega}(\mathcal{A}\ast\mathcal{B}) 
\end{align}

When \eqref{reinipro1} is solved by iterative algorithms such as the Alternating Direction Method of Multiplier (ADMM) \cite{lin2010augmented}, 
 the t-SVD operation is only involved in solving  \begin{equation}\label{ABProx}
  \mathbb{S}_{\tau,g}(\mathcal{\mathcal{A}\ast\mathcal{B}})=\mathop{\arg\min}_{\mathcal{X}} \tau\|\mathcal{X}\|_{*,g}+\frac{1}{2}\|\mathcal{A}\ast \mathcal{B}-\mathcal{X}\|_F^2. \end{equation}  
 According to Theorem \ref{th2},  a fast solver for \eqref{ABProx} is presented in Algorithm \ref{fSolverABProx}.

\subsubsection{The Developing Optimization Algorithm based on ADMM}
By utilizing the above trick, TCDLR can be solved as follows. 
To simplify \eqref{reinipro1}, an auxiliary tensor $ \mathcal{E}$ is introduced. Then, \eqref{reinipro1}  can be rewritten to:
	 \begin{align}\label{TCDLR_auxi}
	 	&\min_{\mathcal{E}\in \textbf{I}, \mathcal{A}, \mathcal{B}, \mathcal{X}  }\dfrac{\mu}{2}\|\mathcal{A}\ast\mathcal{B}-\mathcal{X}\|^2_F+\|\mathcal{X}\|_{*,g}\notag\\
	&s.t.~  \mathbf{P}_{\Omega}(\mathcal{M})=\mathcal{A}\ast\mathcal{B}+\mathcal{E},
	 \end{align}
where $\textbf{I}= \{\mathcal{E}|\mathbf{P}_{\Omega}(\mathcal{E})=\textbf{0} \}$.

\begin{algorithm}[t]
			\caption{Fast Solver to \eqref{ABProx} }\label{fSolverABProx}
			\KwIn{$\mathcal{A}\in \mathbb{R}^{n_1\times k \times n_3 }$, $\mathcal{B}\in \mathbb{R}^{n_1\times k \times n_3 }$,$\lambda>0$.}
			\KwOut{$\mathbb{S}_{\lambda,g}(\mathcal{A}\ast \mathcal{B})$, $\mathcal{Z}$ ,$\mathcal{ Q}$ and $\mathcal{R}$.}
	  1. Compute $\mathcal{R}$ and $\mathcal{ Q}$   by  QR decomposition \cite{kilmer2013third} of $(\mathcal{B}^{(t+1)})^*$:
         $  \mathcal{B}^*=\mathcal{Q}^*\ast\mathcal{R}^*;$    \\ 
          2. Compute $\mathcal{Z}=\mathcal{A}\ast \mathcal{R}$;\\
          3. Obtain $\mathbb{S}_{\lambda,g}(\mathcal{Z})$ by Generalized Tensor Singular Value Thresholding (GTSVT)\cite{Zhang2022Tensor} \\
          4. Compute $\mathbb{S}_{\lambda,g}(\mathcal{A}\ast \mathcal{B})=\mathbb{S}_{\lambda,g}(\mathcal{Z})\ast\mathcal{Q}$. 
    \\ 
		\end{algorithm}

\begin{algorithm}[h]
\caption{Solve \eqref{TCDLR} by the Developing Optimization Algorithm}\label{ourmethod} 
\KwIn{The observed tensor   $ \mathbf{P}_{\Omega}(\mathcal{M}) \in \mathbb{R}^{n_1 \times n_2 \times n_3}$, the support set $ \Omega $, $\rho=1.3$, $\bar{\mu}=10^{14}$, $ \varepsilon=10^{-9}$,  $k$.} 
\KwOut{$\mathcal{X}^{(t+1)}$} 
\textbf{Initialize:} $ \mu^{(0)}$,  $ \mathcal{B}^{(0)}$, $ \mathcal{E}^{(0)} $, $ \mathcal{X}^{(0)}$,  $ \mathcal{Y}^{(0)}$, and $t=0$

\textbf{While not converge do}

1.  Compute $ \mathcal{A}^{(t+1)}$ by  \eqref{updateA};   

2.  Compute $ \mathcal{B}^{(t+1)} $ by \eqref{updateB};

3.  Calculate $ \mathcal{ Q}^{(t+1)}$,  $ \mathcal{Z}^{(t+1)} $ and $ \mathcal{ X}^{(t+1)} $ by Algorithm \ref{fSolverABProx};  

4. Calculate $ \mathcal{P}^{(t+1)} $ by $ \mathcal{P}^{(t+1)}=   \mathcal{Z}^{(t+1)} \ast \mathcal{Q}^{(t+1)}$;

5. Calculate $ \mathcal{ E}^{(t+1)} $ by \eqref{update_e};

6. Calculate  $ \mathcal{ Y}^{(t+1)} $ by \eqref{update_y};

7. Calculate $\mu^{(t+1)}$ by \eqref{update_mu}; 

8. Check the convergence condition: $\|\mathbf{P}_{\Omega}(\mathcal{M})-\mathcal{P}^{(t)}-\mathcal{E}^{(t)}\|_{\infty}<\varepsilon$, $\|\mathcal{P}^{(t+1)}-\mathcal{P}^{(t)}\|_{\infty}<\varepsilon$ ,$\|\mathcal{E}^{(t+1)}-\mathcal{E}^{(t)}\|_{\infty}<\varepsilon$; \\
9. $t=t+1$.\\
\textbf{end while}

\end{algorithm}

\begin{table*}
	\centering 
	\caption{The Computational complexity of Algorithm \ref{ourmethod}}
	\scalebox{0.6}{
		\resizebox{\textwidth}{!}{%
			\begin{tabular}{|c|c|c|}
				\hline
				Step no. & Operation & Complexity \\ \hline
				1        & $\overbrace{{{\mathsf{\mathcal{A}}}^{(t+1)}}}^{{{n}_{1}}\times k\times {{n}_{3}}}=\underset{\mathsf{\mathcal{A}}}{\mathop{\arg \min }}\,\mu \|\overbrace{{{\mathsf{\mathcal{C}}}^{(t+1)}}}^{{{n}_{1}}\times {{n}_{2}}\times {{n}_{3}}}-\overbrace{\mathsf{\mathcal{A}}}^{{{n}_{1}}\times k\times {{n}_{3}}}*\overbrace{{{\mathsf{\mathcal{B}}}^{(t)}}}^{k\times {{n}_{2}}\times {{n}_{3}}}\|_{F}^{2}$ &$\mathcal{O}(n_1n_2n_3\log n_3 +kn_1n_2n_3)$            \\ \hline
				2        &$\overbrace{{{\mathsf{\mathcal{B}}}^{(t+1)}}}^{{k}\times {n}_{2}\times {{n}_{3}}}=\underset{\mathsf{\mathcal{B}}}{\mathop{\arg \min }}\,\mu \|\overbrace{{{\mathsf{\mathcal{C}}}^{(t+1)}}}^{{{n}_{1}}\times {{n}_{2}}\times {{n}_{3}}}-\overbrace{\mathsf{\mathcal{A}}^{(t+1)}}^{{{n}_{1}}\times k\times {{n}_{3}}}*\overbrace{{{\mathsf{\mathcal{B}}}}}^{k\times {{n}_{2}}\times {{n}_{3}}}\|_{F}^{2}$    &$\mathcal{O}(n_1n_2n_3\log n_3 +kn_1n_2n_3)$   \\ \hline
				3        &$[ \overbrace{{({\mathsf{\mathcal{Q}}}^{(t+1)})^*}}^{{{n}_{2}}\times k \times {{n}_{3}}},\overbrace{{({\mathsf{\mathcal{R}}}^{(t+1)})^*}}^{k\times k\times {{n}_{3}}} ]=\operatorname{QR}( \overbrace{{({\mathsf{\mathcal{B}}}^{(t+1)})^*}}^{{{n}_{2}}\times k \times {{n}_{3}}} )$ & $\mathcal{O}(n_2kn_3\log n_3+n_2k^2n_3)$            \\ \hline
				3        &$\overbrace{{{\mathsf{\mathcal{Z}}}^{(t+1)}}}^{{{n}_{1}}\times k\times {{n}_{3}}}=\overbrace{{{\mathsf{\mathcal{A}}}^{(t+1)}}}^{{{n}_{1}}\times k\times {{n}_{3}}}*\overbrace{{{\mathsf{\mathcal{R}}}^{(t+1)}}}^{k\times k\times {{n}_{3}}}$           &$\mathcal{O}((n_1+k)kn_3\log n_3+n_1k^2n_3)$            \\ \hline
				3        &${{\mathbb{S}}_{\frac{1}{\mu },g}}(\overbrace{{{\mathsf{\mathcal{Z}}}^{(t+1)}}}^{{{n}_{1}}\times k\times {{n}_{3}}})$ & $\mathcal{O}(n_1kn_3\log n_3+n_1k^2n_3)$  \\ \hline
				4        & $\overbrace{{{\mathsf{\mathcal{P}}}^{(t+1)}}}^{{{n}_{1}}\times {{n}_{2}}\times {{n}_{3}}}=\overbrace{{{\mathsf{\mathcal{A}}}^{(t+1)}}}^{{{n}_{1}}\times k\times {{n}_{3}}}*\overbrace{{{\mathsf{\mathcal{B}}}^{(t+1)}}}^{k\times {{n}_{2}}\times {{n}_{3}}}$          & $\mathcal{O}(n_1n_2n_3\log n_3+n_1kn_2n_3)$           \\ \hline
				total &   total cost at each iteration      & $\mathcal{O}(n_1n_2n_3\log n_3 +kn_1n_2n_3)$           \\ \hline
			\end{tabular}\label{ComCom}}}
\end{table*}

The lagrangian function of \eqref{TCDLR_auxi} is formulated as 
\begin{align}
&\mathcal{L}_{\mu}(\mathcal{X},\mathcal{E},\mathcal{A},\mathcal{B},\mathcal{Y})\notag \\
=&\|\mathcal{X}\|_{*,g}+\dfrac{\mu}{2}\|\mathcal{X}-\mathcal{A}\ast\mathcal{B}\|^2_F+ \langle\mathbf{P}_{\Omega}(\mathcal{M})-\mathcal{A}\ast\mathcal{B}-\mathcal{E}, \mathcal{Y}\rangle\notag \\
&+\dfrac{\mu}{2}\|\mathbf{P}_{\Omega}(\mathcal{M})-\mathcal{A}\ast\mathcal{B}-\mathcal{E}\|_F^2,
\end{align}
  where $\mathcal{Y}$ is Lagrange multiplier, and $ \mu $ is a positive scalar. Therefore, \eqref{TCDLR_auxi} is iteratively solved by ADMM as follows.\\
\textbf{Step 1} Calculate $\mathcal{A}^{(t+1)}$ by
\begin{align}
\mathcal{  A}^{(t+1)}=	&\mathop{\arg\min}_{ \mathcal{A}   }\mathcal{L}_{\mu}(\mathcal{X}^{(t)},\mathcal{E}^{(t)},\mathcal{A},\mathcal{B}^{(t)},\mathcal{Y}^{(t)})\notag \\
=	&\mathop{\arg\min}_{ \mathcal{A} } \|\mathcal{X}^{(t)}-\mathcal{A}\ast\mathcal{B}^{(t)}\|^2_F\notag \\
&~~~~~~+\|\mathbf{P}_{\Omega}(\mathcal{M})-\mathcal{A}\ast\mathcal{Q}^{(t)}-\mathcal{E}^{(t)}+\dfrac{1}{\mu^{(t)}}\mathcal{Y}^{(t)}\|_F^2\notag \\
=	&\mathop{\arg\min}_{ \mathcal{A} }\|\mathcal{C}^{(t+1)}-\mathcal{A}\ast\mathcal{B}^{(t)}\|_F^2 \notag \\
=&\mathcal{C}^{(t+1)}\ast(\mathcal{B}^{(t)})^*\ast(\mathcal{B}^{(t)}\ast(\mathcal{B}^{(t)})^*)^{\dag}  
\end{align} 
where $\mathcal{C}^{(t+1)}=(\mathbf{P}_{\Omega}(\mathcal{M})-\mathcal{E}^{(t)}+\mathcal{X}^{(t)}+\dfrac{\mathcal{Y}^{(t)}}{\mu^{(t)}})/2$. \\
\textbf{Step 2} Calculate $\mathcal{B}^{(t+1)}$ by \begin{align}\label{updateB}
	\mathcal{  B}^{(t+1)}=	&\mathop{\arg\min}_{ \mathcal{B}   }\mathcal{L}_{\mu}(\mathcal{X}^{(t)},\mathcal{E}^{(t)},\mathcal{A}^{(t+1)},\mathcal{B},\mathcal{Y}^{(t)})\notag\\ =&\mathop{\arg\min}_{\mathcal{B}}\|\mathcal{C}^{(t+1)}-\mathcal{A}^{(t+1)}\ast\mathcal{B}\|_F^2 \notag \\
	=&((\mathcal{A}^{(t+1)})^*\ast\mathcal{A}^{(t+1)})^{\dag} \ast(\mathcal{A}^{(t+1)})^*\ast\mathcal{C}^{(t+1)}
\end{align} 
\textbf{Step 3} Calculate $\mathcal{X}^{(t+1)}$ by
\begin{align}\label{upate_x}
	\mathcal{  X}^{(t+1)}=&\mathop{\arg\min}_{\mathcal{X}}\frac{1}{2}\|\mathcal{A}^{(t+1)}\ast\mathcal{B}^{(t+1)}-\mathcal{X}\|_F^2+\frac{1}{\mu^{(t)}}\|\mathcal{X}\|_{*,g}\notag\\
	= &  \mathbb{S}_{\frac{1}{\mu^{(t)}},g}(\mathcal{A}^{(t+1)}\ast\mathcal{B}^{(t+1)}),
\end{align}
and obtain $\mathcal{Z}^{(t+1)}$ and $\mathcal{Q}^{(t+1)}$  by
Algorithm \ref{fSolverABProx}. \\ 
\textbf{Step 4} Calculate $\mathcal{E}^{(t+1)}$ by
\begin{align}\label{update_e}
	\mathcal{E}^{(t+1)}=&\mathop{\arg\min}_{\mathcal{E}}\frac{\mu^{(t)}}{2}\|\mathbf{P}_{\Omega}(\mathcal{M})-\mathcal{P}^{(t+1)}+\dfrac{\mathcal{Y}^{(t)}}{\mu^{(t)}}-\mathcal{E}\|_F^2, \notag\\
 &~~~~~~s.t.~\mathcal{E}\in \textbf{I},
\end{align}
where $\mathcal{P}^{(t+1)}=\mathcal{Z}^{(t+1)}\ast\mathcal{Q}^{(t+1)}$.\\
\textbf{Step 5} Calculate $\mathcal{Y}^{(t)}$ by
\begin{equation}\label{update_y}
	\mathcal{Y}^{(t+1)}=\mu^{(t)}(\mathbf{P}_{\Omega}(\mathcal{M})-\mathcal{P}^{(t+1)}-\mathcal{E}^{(t+1)})+\mathcal{Y}^{(t)}.
\end{equation}
\textbf{Step 6} Calculate $\mu^{(t+1)}$ by
\begin{equation}\label{update_mu}
	\mu^{(t+1)}=\min(\bar{\mu},\rho\mu^{(t)}),
\end{equation}
where $\bar{\mu}$ is the upper bound of $\mu^{(t+1)}$, and $\rho>1$.


According to $\mathcal{A}^{(t)}\ast\mathcal{B}^{(t)}=\mathcal{Z}^{(t)}\ast\mathcal{Q}^{(t)}$, $\mathrm{range}(\mathcal{Z}^{(t)})$ and $\mathrm{range}(\mathcal{Q}^{(t)})$ are the column space and row space of $\mathcal{A}^{(t)}\ast\mathcal{B}^{(t)}$, respectively. Then, with $(\mathcal{B}^{(t+1)})^*=(\mathcal{Q}^{(t+1)})^*\ast(\mathcal{R}^{(t+1)})^*$, we have
\begin{align}
    &\min_{\mathcal{B}}\|\mathcal{C}^{(t+1)}-\mathcal{C}^{(t+1)}\ast(\mathcal{Q}^{(t)})^*\ast\mathcal{B} \|^2_F \notag \\ 
   \leq &\min_{\mathcal{B}} \|\mathcal{C}^{(t+1)}-\mathcal{C}^{(t+1)}\ast(\mathcal{B}^{(t)})^*\ast(\mathcal{B}^{(t)}\ast(\mathcal{B}^{(t)})^*)^{\dag}\ast\mathcal{B} \|^2_F. 
\end{align}  
 
Therefore, $ \mathcal{A}^{(t+1)}$ is updated by 
\begin{equation}\label{updateA}
    \mathop{\arg\min}_{\mathcal{A}}\|\mathcal{C}^{(t+1)}- \mathcal{A} \ast \mathcal{Q}^{(t)}\|_F^2=\mathcal{C}^{(t+1)}(\mathcal{Q}^{(t)})^*
\end{equation} 
    for  easy computation.   
 The whole procedure of the algorithm is presented in Algorithm \ref{ourmethod}.
 
\subsection{Complexity Analysis}
 According to Algorithm \ref{ourmethod}, since TCDLR only requires computing the t-SVD of $\mathcal{Z}$ with a smaller size, it avoids performing the t-SVD operation with a high time complexity of $\mathcal{O}(kn_1n_2n_3)$. 
The step-by-step computational complexity of the proposed algorithm is summarized in Table \ref{ComCom}. The total cost at each iteration (requires computing of FFT \cite{34}) is $\mathcal{O}(n_1n_2n_3\log n_3 +kn_1n_2n_3)$, as shown in Table \ref{ComCom}.

\subsection{Rank Estimation}

In the following, a method is proposed to estimate $k$ in \eqref{AB} for a lower bound $k_{\min}$ and an upper bound $k_{\max}$  of $k$. 

\subsubsection{Increase strategy}\label{in}

According to Theorem \ref{th1} (iv), $k<\mathrm{rank}(\widetilde{\mathcal{X}})$ holds and $k$ should be increased when $\|\hat{\mathcal{X}}\|_{*,(k,g)}=\infty$. In the following, the case of $\|\hat{\mathcal{X}}\|_{*,(k,g)}=\infty$ is considered.

Since $\|\hat{\mathcal{X}}\|_{*,(k,g)}=\infty$, $k<\mathrm{rank}(\mathcal{X})$ holds for $\forall \mathcal{X} \in \{\mathcal{X}|  \mathbf{P}_{\Omega}(\mathcal{M})=\mathbf{P}_{\Omega}(\mathcal{X})\}$, \eqref{AB} has no solution. A preliminary idea for the critical condition of increasing $k$ is $\|\mathcal{C}^{(t)}-\mathcal{A}^{(t)}\ast\mathcal{B}^{(t)}\|_F \nrightarrow 0$ when $t \rightarrow\infty$. However, it is difficult to determine whether $\|\mathcal{C}^{(t)}-\mathcal{A}^{(t)}\ast\mathcal{B}^{(t)}\|_F$ converges to $0$ and the large value of $\|\mathcal{C}^{(t)}-\mathcal{A}^{(t)}\ast\mathcal{B}^{(t)}\|_F$ should be allowed in the early iteration of the algorithm. According to \cite{21},  $\widetilde{\mathcal{A}}\ast\widetilde{\mathcal{B}}$ is the rank-$k$ estimation of $\mathcal{C}^{(t)}$   if $(\widetilde{\mathcal{A}},\widetilde{\mathcal{B}})$ is the the optimal solution to $\min_{\mathcal{A},\mathcal{B}}\|\mathcal{C}^{(t)}-\mathcal{A}\ast\mathcal{B}\|_F$.  Therefore, we  increases $k$ only if some important components of  $\mathcal{C}^{(t)}$ are lost in $\mathcal{A}^{(t)}\ast\mathcal{B}^{(t)}$.

Defining $\mathcal{D}=\mathcal{C}^{(t)}-\mathcal{A}^{(t)}\ast\mathcal{B}^{(t)} \in \mathbb{R}^{n_1\times n_2\times n_3}$, according to \cite{vershynin2010introduction}, if $ \bar D_{i} $ is Gaussian distributed for given $i$ (in this case, $ \bar D_{i} $ contains less meaningful information), $\sigma_{1}(\frac{\bar D_{i}-\hat \mu_i(\bar D_{i})}{\hat\delta_i(\bar D_{i})})\leq \sqrt{n_1}+\sqrt{n_2}+h (h\geq 1)$ holds with a high probability, where  $\hat\mu_i(\bar D_{i})$ is the mean value of all elements in $\bar D_{i}$ and $\hat\delta_i(\bar D_{i})=\sqrt{\frac{1}{n_1n_2-1}\sum^{n_1}_{w=1}\sum^{n_2}_{j=1}([\bar D_{i}]_{w,j}-\hat\mu_i)^2}$. Here, this paper  uses $\bar s_{\max}(\frac{\bar D_{i}-\hat \mu_i}{\hat\delta_i})=\|\textbf{p}^*\cdot \frac{\bar D_{i}-\hat \mu_i}{\hat\delta_i}\|_2$ to estimate  $\sigma_{1}(\frac{\bar D_{i}-\hat \mu_i(\bar D_{i})}{\hat\delta_i(\bar D_{i})})$, where $\textbf{p}^*$ is a $1\times n_1$ vector whose entries are independent standard normal random variables, $\hat\delta_i=\sqrt{\frac{1}{w-1}\sum^w_{j=1}(\textbf{s}_j-\hat\mu_i)^2}$, $\hat\mu_i=\frac{1}{w}\sum^w_{j=1}\textbf{s}_j$ and $\textbf{s}\in \mathbb{R}^{w}$ is a vector sampled from $\bar D_{i}$.


Thus, when $\bar s_{\max}(\frac{\bar D_{i}-\hat \mu_i}{\hat\delta_i})> \sqrt{n_1}+\sqrt{n_2}+h$  ($ h\geq 1$), $k^{(t)}_i$ is increased to  
\begin{equation}\label{increase}
k_i^{(t+1)}=\min(k_i^{(t)}+l, k_{\max}) (l>0).
\end{equation}

By performing the QR decomposition of $[\bar Q_i^{(t+1)}; P\cdot \bar D_{i}]^*$, we have 
$$[\bar Q_i^{(t+1)}; P\cdot \bar D_{i}]^*=\tilde Q_i^* \tilde R_i^*,$$ 
where $P$ is an $l\times n_1$ matrix whose entries are independent standard normal random variables. 
Thus, we augment
\begin{equation}\label{au_Q}
\bar Q_i^{(t+1)}\leftarrow \tilde Q_i\in \mathbb{C}^{ k_i^{(t+1)}\times n_2}
\end{equation}
and 
\begin{equation}\label{au_Z}
\bar Z_i^{(t+1)}\leftarrow [\bar Z_i^{(t+1)}, \textbf{0}]\tilde R_i\in \mathbb{C}^{n_1 \times k_i^{(t+1)}}. 
\end{equation}

\begin{algorithm}[h]
\caption{TCDLR with the proposed rank estimation (TCDLR-RE)}\label{ourmethod2new} 
\scriptsize
\KwIn{ The tensor data $ \mathcal{M} \in \mathbb{R}^{n_1 \times n_2 \times n_3}$, the observed set $ \Omega $,  $\rho=1.3$, $\bar{\mu}=10^{14}$, $ \varepsilon=10^{-9}$, $k_{\min}$, $k_{\max}$,   $l=\min(n_1,n_2)/50$, and $h=1$.} 
\KwOut{$\mathcal{X}^{(t+1)}$} 
\textbf{Initialize:} $t=0$,  $ \mathcal{E}^{(0)} $, $ \mathcal{X}^{(0)}$,   $ \mathcal{Y}^{(0)}$, $ \mu^{(0)}$,  $ k_i^{(0)} \in \mathbb{N}^+$  and $ \bar{Q}_i^{(0)} \in \mathbb{C}^{n_2 \times k_i^{(0)}}$  for $i=1,2,\cdots,\lfloor\frac{n_3+1}{2}\rfloor$. 

\textbf{While not converge do}\\
1. $\bar C^{(t+1)}=\mathrm{bdiag}(\mathrm{fft}((\mathbf{P}_{\Omega}(\mathcal{M})-\mathcal{E}^{(t)}+\mathcal{X}^{(t)}+\dfrac{\mathcal{Y}^{(t)}}{\mu})/2,[],3))$;\\
 ~~~~\textbf{for} $i=1,...,\lfloor\frac{n_3+1}{2}\rfloor$\textbf{do}\\
 ~~~~~~~~2.  
  $\bar A_i^{(t+1)}=\bar C^{(t+1)}_i(\bar Q_i^{(t)})^*$;\\
  ~~~~~~~~3.	$\bar B_i^{(t+1)}=((\bar A_i^{(t+1)})^*\bar A_i^{(t+1)})^{\dag}(\bar A_i^{(t+1)})^* \bar{C}_i^{(t)}$;\\
    ~~~~~~~~4. Update $  \bar Q_i^{(t+1)}$ and $ \bar Z_i^{(t+1)} $   as follows: \\ 
 ~~~~~~~~~~~~$[(\bar Q^{(t+1)}_i)^*, (\bar R^{(t+1)}_i)^*]=\mathrm{QR}((\bar B_i^{(t+1)})^*)$;\\
  ~~~~~~~~~~~~$\bar Z_i^{(t+1)}=\bar A_i^{(t+1)}\bar R_i^{(t+1)}$;\\
  ~~~~~~~~5. Increase  $ k_i^{(t+1)} $ by \eqref{increase}, and adjust $  \bar Q_i^{(t+1)}$ and $\bar Z_i^{(t+1)} $ by  \eqref{au_Q}\\ ~~~~~~~~~~~and \eqref{au_Z}, respectively;\\
~~~~~~~~6. $\bar X^{(t+1)}_i=\text{GSVT}(\bar Z^{(t+1)}_i)\bar Q^{(t+1)}_i$,  $\bar P^{(t+1)}_i=\bar Z^{(t+1)}_i\bar Q^{(t+1)}_i$;\\
~~~~~~~~7. Decrease  $ k_i^{(t+1)} $ by \eqref{decrease}, and adjust $  \bar Q_i^{(t+1)}$ and $\bar Z_i^{(t+1)} $by  \eqref{de_Q} \\ ~~~~~~~~~~~and \eqref{de_Z}, respectively;\\
  ~~~~\textbf{end for}\\
   ~~~~\textbf{for} $i=\lfloor\frac{n_3+1}{2}\rfloor+1,...,n_3$\textbf{do}\\ 
   ~~~~~~~~8. Update $ k_i^{(t+1)}$, $\bar X^{(t+1)}_i$ and  $\bar P^{(t+1)}_i$ as follows:\\
       ~~~~~~~~~~$ k_i^{(t+1)}=k_{n_3-i+2}^{(t+1)} $; $\bar X^{(t+1)}_i=\text{Conj}(\bar X^{(t+1)}_{n_3-i+2})$;\\
        ~~~~~~~~ $\bar P^{(t+1)}_i=\text{Conj}(\bar P^{(t+1)}_{n_3-i+2})$; 
       \\
     ~~~~\textbf{end for}\\

9. Calculate  $ \mathcal{ X}^{(t+1)}= \mathrm{ifft}(\mathcal{\bar X}^{(t+1)},[],3)$ and $ \mathcal{ P}^{(t+1)}= \mathrm{ifft}(\mathcal{\bar P}^{(t+1)},[],3)$;

10. Update $ \mathcal{ E}^{(t+1)} $, $ \mathcal{ Y}^{(t+1)} $ and $\mu^{(t+1)}$ by \eqref{update_e}, \eqref{update_y} and \eqref{update_mu}, respectively;

11. Check the convergence condition: $ \|\mathcal{P}^{(t+1)}-\mathcal{P}^{(t)}\|_{\infty}<\varepsilon$, $\|\mathcal{E}^{(t+1)}-\mathcal{E}^{(t)}\|_{\infty}<\varepsilon, \|\mathbf{P}_{\Omega}(\mathcal{M})-\mathcal{P}^{(t)}-\mathcal{E}^{(t)}\|_{\infty}<\varepsilon$ \\
12. $t=t+1$.\\ 
\textbf{end while}

\end{algorithm}

\subsubsection{Decrease strategy}


Note that in the computing process $\mathbb{S}_{\frac{1}{\mu},g}(\mathcal{Z}^{(t+1)})$, the singular values of $\bar Z_i^{(t+1)}$
will be obtained first. Therefore, it is assumed that $ \lambda_{1,i} \geq \lambda_{2,i} \geq \dots \geq \lambda_{k_i^{(t+1)},i}$ are singular values of $\bar Z_i^{(t+1)}$.  
Then, the quotient sequence $ \tilde{\lambda}_{j,i}=\lambda_{j,i}/\lambda_{j+1,i}(j=1,\dots,k_i^{(t+1)}-1) $ is computed. Suppose $s_i=\mathop{\arg\max}_{1\leq j < r_i}\tilde{\lambda}_{j,i}$ and  $\tau_i =\frac{(k_i^{(t+1)}-1)\tilde{\lambda}_{s_i,i}}{\sum_{j \neq s_i}  \tilde{\lambda}_{j,i}}$, 
if $ \tau_i\geq 10 $ indicates a large drop in the magnitude of singular values \cite{29}, $k_i^{(t+1)}$ should be updated as follows \cite{candes2011robust}:
\begin{equation}\label{decrease}
k_i^{(t+1)}=\max(\tilde{k}_i, k_{\min}),
\end{equation}
where $\tilde{k}_i$ satisfies $\sum_{j=1}^{\tilde{k}_i} \lambda_{j,i} / \sum_{j=1}^{r_{i}} \lambda_{j,i} \geq 95 \%$.  


Let $\bar Z^{(t+1)}_i=Q_Z\cdot R_Z$ and $ R_Z=\tilde{U}\cdot \tilde{S} \cdot \tilde{V}^*$ be the QR decomposition of $\bar Z^{(t+1)}_i$ and the SVD of $ R_Z$, respectively. This paper updates
\begin{equation}\label{de_Z}
\bar Z^{(t+1)}_i\leftarrow Q_Z\cdot [\tilde{U}]_{:,1:k_i^{(t+1)}}\cdot [\tilde{S}]_{1:k_i^{(t+1)},1:k_i^{(t+1)}},
\end{equation}
and 
\begin{equation}\label{de_Q}
\bar Q^{(t+1)}_i\leftarrow  [\tilde{V}]^*_{:,1:k_i^{(t+1)}}\cdot \bar Q^{(t+1)}_i.
\end{equation}

Different from \cite{29}, in the proposed rank estimation strategy, $ k^{(t+1)}_i $, $\bar Z^{(t+1)}_i$, and $\bar Q^{(t+1)}_i$ are adjusted for each slice by considering the difference of each slice in the tensor. Based on Algorithm \ref{ourmethod} and the proposed rank estimation method, Algorithm \ref{ourmethod2new} (TCDLR-RE) is given.

\begin{table*}[]
\caption{  Comparison of ${\tt relerr}$ and running time (seconds) on synthetic data when the sampling rate=30\% and $\bar{r}=0.1n$.}
\centering
\resizebox{\textwidth}{!}{%
\begin{tabular}{c|cc|cc|cc|cc|cc|c|c}
\hline
Method &
  \multicolumn{2}{c|}{t-TNN\cite{37}} &
  \multicolumn{2}{c|}{TCTF\cite{29}} &
  \multicolumn{2}{c|}{PSTNN\cite{jiang2020multi}} &
  \multicolumn{2}{c|}{TC-RE\cite{shi2021robust}} &
  \multicolumn{2}{c|}{TCDLR} &
   \multicolumn{2}{c}{TCDLR-RE} 
   \\ \hline
Data &
  \multicolumn{1}{c|}{${\tt relerr}$} &
  time(s) &
  \multicolumn{1}{c|}{${\tt relerr}$} &
  time(s) &
  \multicolumn{1}{c|}{${\tt relerr}$} &
  time(s) &
  \multicolumn{1}{c|}{${\tt relerr}$} &
  time(s) &
  \multicolumn{1}{c|}{${\tt relerr}$} &
  time(s) &
  ${\tt relerr}$ &
  time(s) \\ \hline
$n=1000$   &
  \multicolumn{1}{c|}{$5.24E^{-2}$} &
  140.9 &
  \multicolumn{1}{c|}{$5.29E^{-2}$} &
  \textbf{30.4} &
  \multicolumn{1}{c|}{$1.19E^{-1}$} &
  64.5 &
  \multicolumn{1}{c|}{$5.06E^{-2}$} &
  1338.2 &
  \multicolumn{1}{c|}{$\mathbf{4.75E^{-3}}$} &
  62.9 &
  $\mathbf{4.75E^ { -3 }}$ &
  $\mathbf{42.4}$ \\ \hline
$n=2000$   &
  \multicolumn{1}{c|}{$5.15E^{-2}$} &
  1591.7 &
  \multicolumn{1}{c|}{$6.85E^{-2}$} &
  \textbf{165.6} &
  \multicolumn{1}{c|}{$2.37E^{-1}$} &
  742.9 &
  \multicolumn{1}{c|}{$7.23E^{-2}$} &
  19701.8 &
  \multicolumn{1}{c|}{$\mathbf{5.28E^{-3}}$} &
  432.0 &
  $\mathbf{5.40E^ { -3 }}$ &
    $\mathbf{228.9}$  \\ \hline
$n=3000$   &
  \multicolumn{1}{c|}{$5.10E^{-2}$} &
  7837.8 &
  \multicolumn{1}{c|}{$6.81E^{-2}$} &
  \textbf{481.0} &
  \multicolumn{1}{c|}{$3.24E^{-1}$} &
  2384.2 &
  \multicolumn{1}{c|}{$9.67E^{-2}$} &
  63472.4 &
  \multicolumn{1}{c|}{$\mathbf{6.25E^{-3}}$} &
  1536.0 &
  $\mathbf{6.52E^ { -3 }}$ &
   $\mathbf{644.6}$  \\ \hline
$n=4000$   &
  \multicolumn{1}{c|}{$5.19E^{-2}$} &
  17870.6 &
  \multicolumn{1}{c|}{$6.81E^{-2}$} &
  \textbf{1033.2} &
  \multicolumn{1}{c|}{$3.86E^{-1}$} &
  5543.4 &
  \multicolumn{1}{c|}{$8.70E^{-2}$} &
  131239.6 &
  \multicolumn{1}{c|}{$\mathbf{7.76E^{-3}}$} &
  3784.1 &
  $\mathbf{8.30E^ { -3 }}$ &
  $\mathbf{1706.8}$ \\ \hline
\end{tabular}\label{table:synthetic_result}
}
\end{table*}
\begin{figure*}
	\centering
	\subfigure[]{
	\includegraphics[width=1.6in]{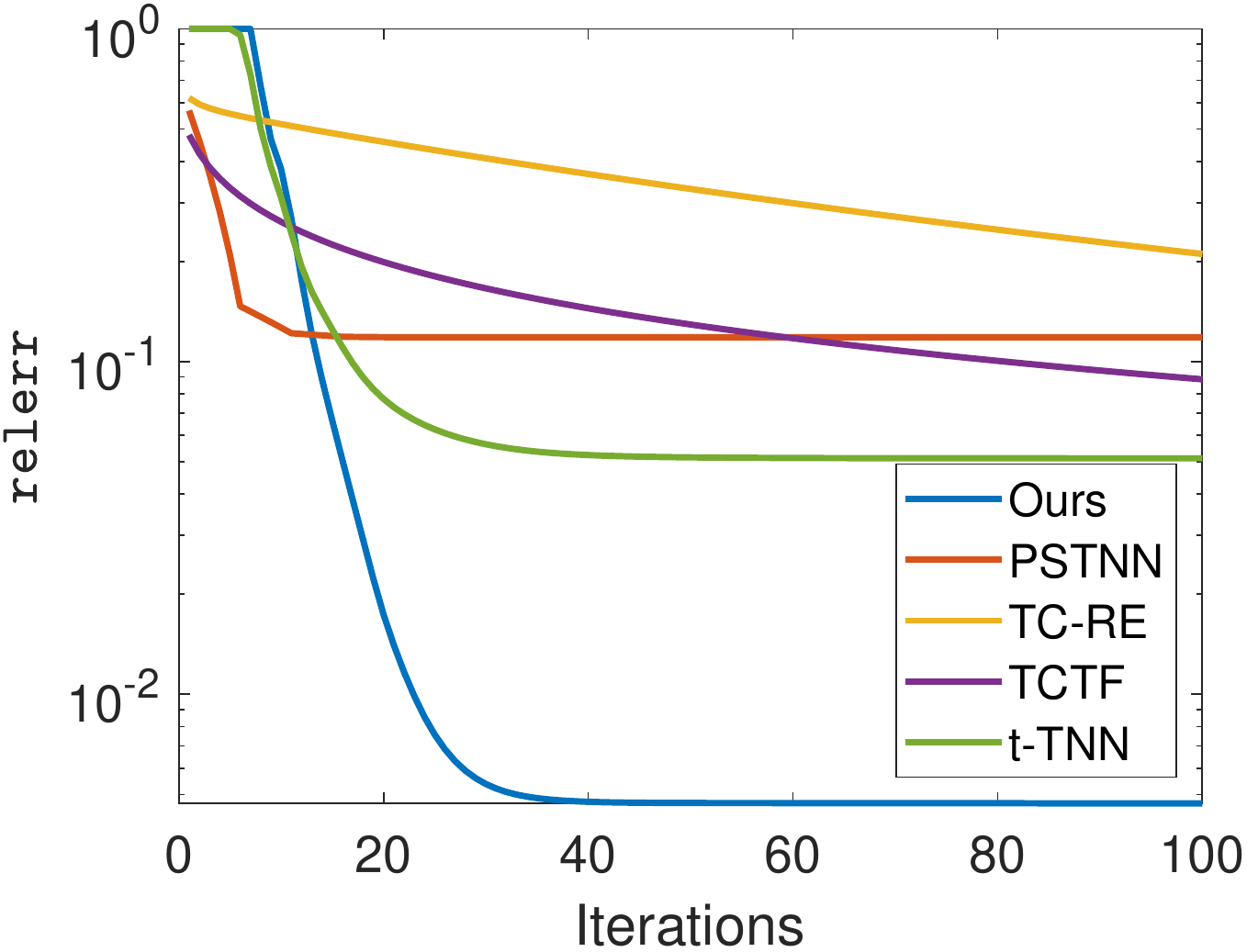}} 
	\subfigure[]{
\includegraphics[width=1.59in]{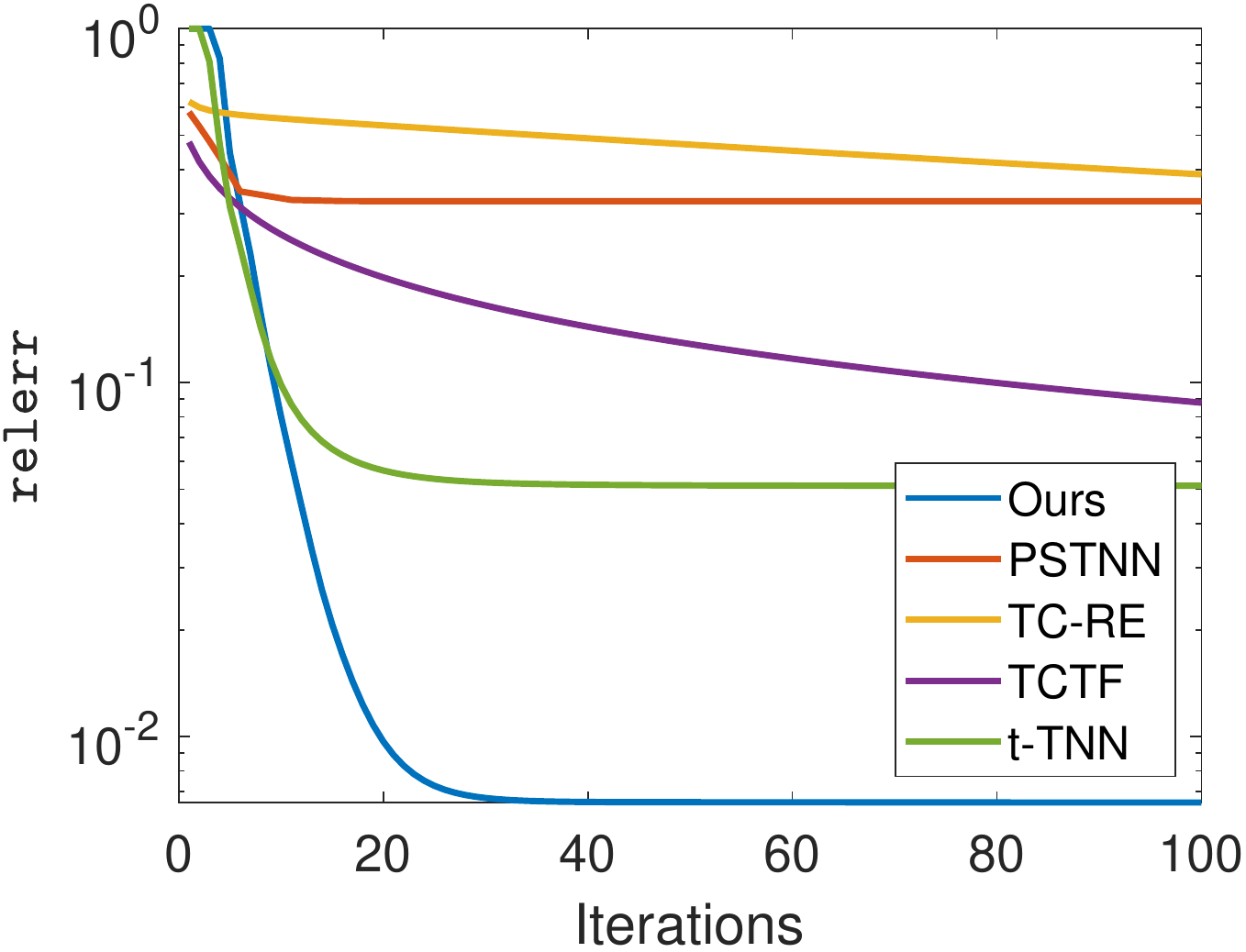}}
\subfigure[]{\includegraphics[width=1.6in]{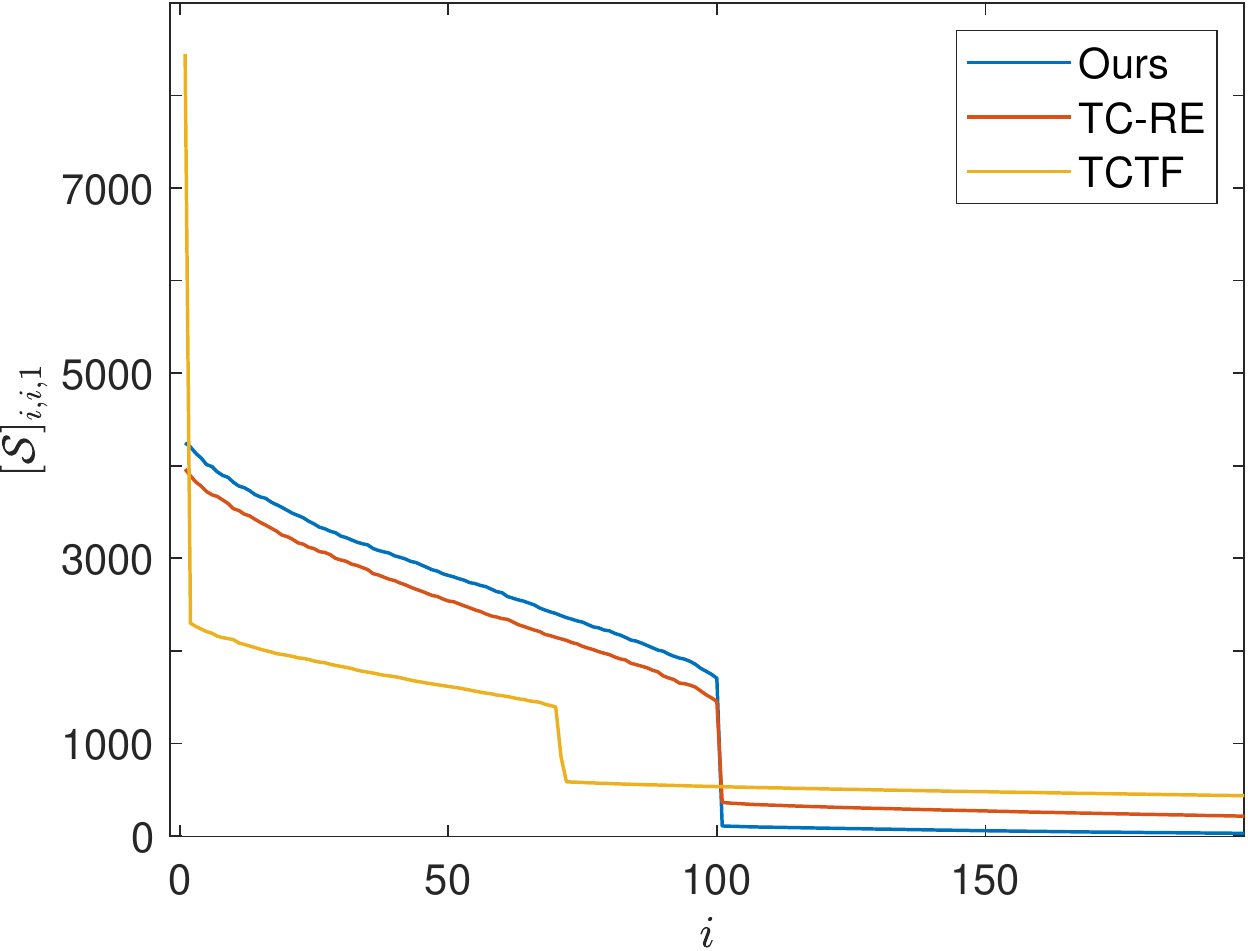}}
	\caption{The relative error with iterations  for 
 (a) $n=1000$ and (b) $n=3000$, and (c) the plot of the singular values (i.e., $[\mathcal{S}]_{i,i,1}$) of the recovered low-rank tensor by different methods for $n=1000$, under the tubal rank $\bar r=0.1\times n$ and a sampling rate of $30\%$. }\label{loss}
\end{figure*}

\begin{figure}
	\centering 
	\subfigure[t-TNN \cite{37}]{
		\includegraphics[width=1in]{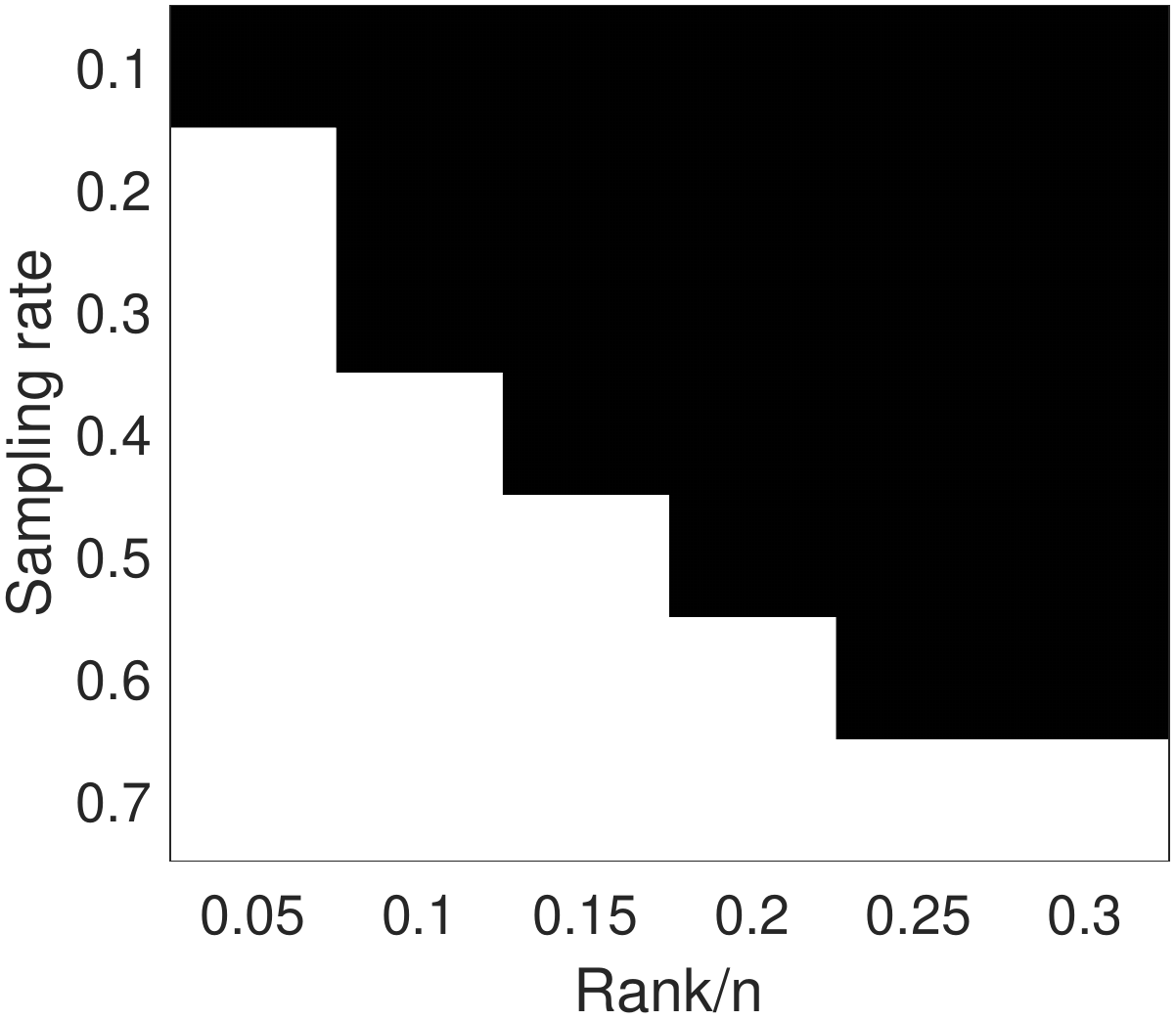}}
    	\subfigure[TC-RE \cite{shi2021robust}]{					
		\includegraphics[width=1in]{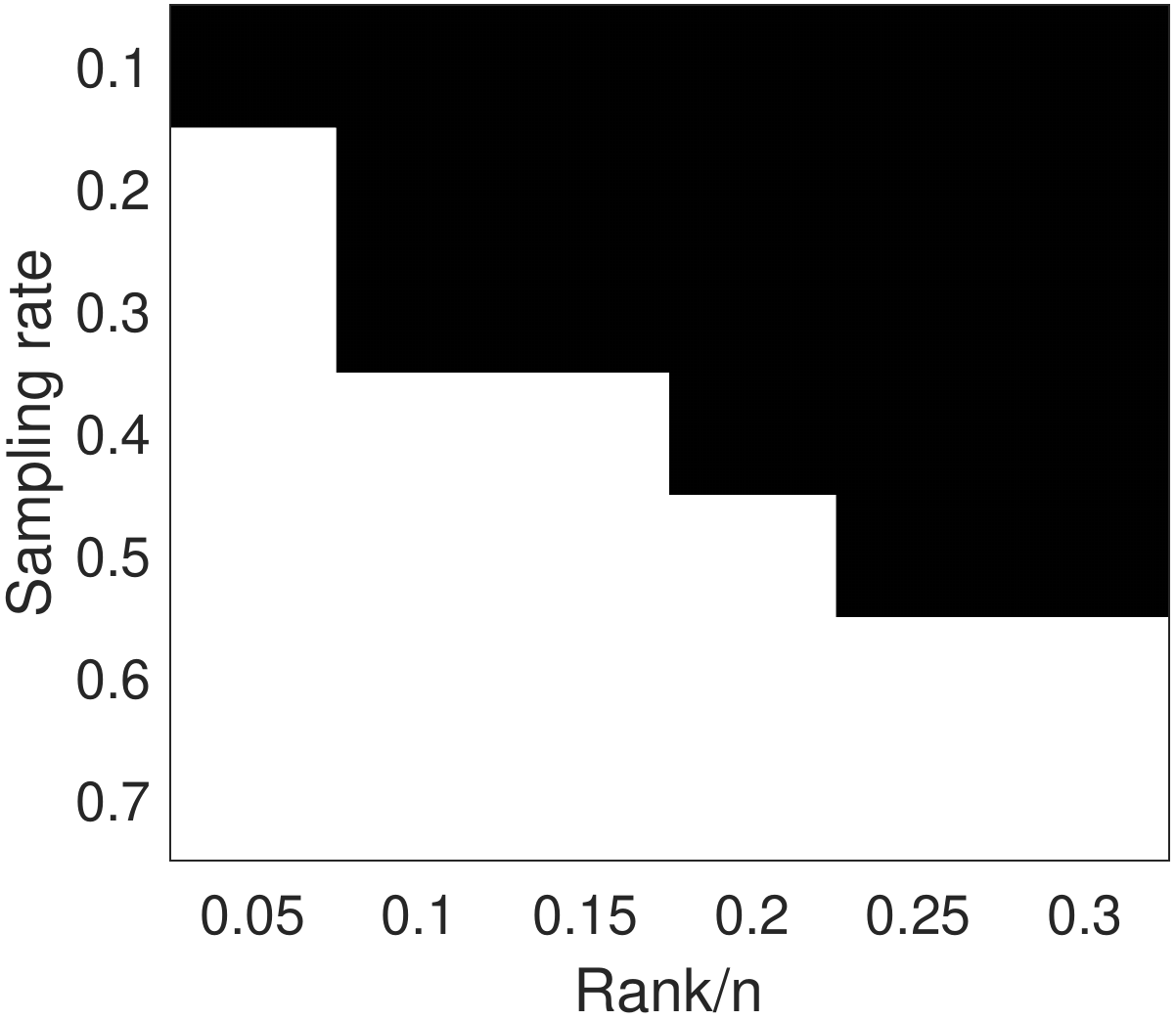}}
  \subfigure[TCDLR-RE]{					
		\includegraphics[width=1in]{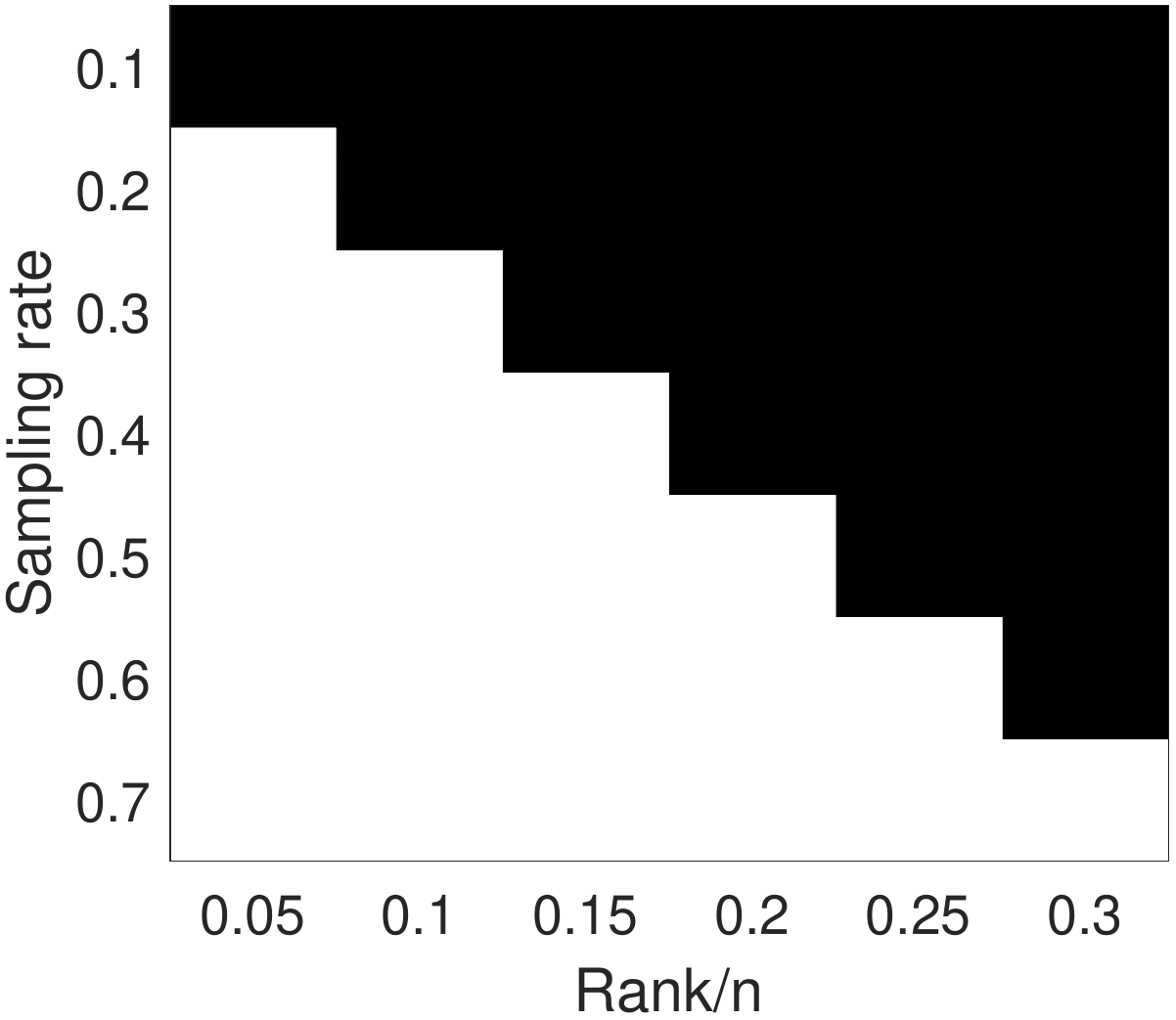}}
	
	\caption{ Comparison of the recovery capacity in $ 1000\times1000\times3 $ tensor with varying tubal ranks and sampling rates. The white regions denote successful recovery with $ {\tt relerr} \leq 10^{-2}$; the black regions denote failed recovery with $ {\tt relerr}>10^{-2}$.   } \label{aaa}
	
\end{figure}

\begin{figure*}
\centering
\includegraphics[width=6in]{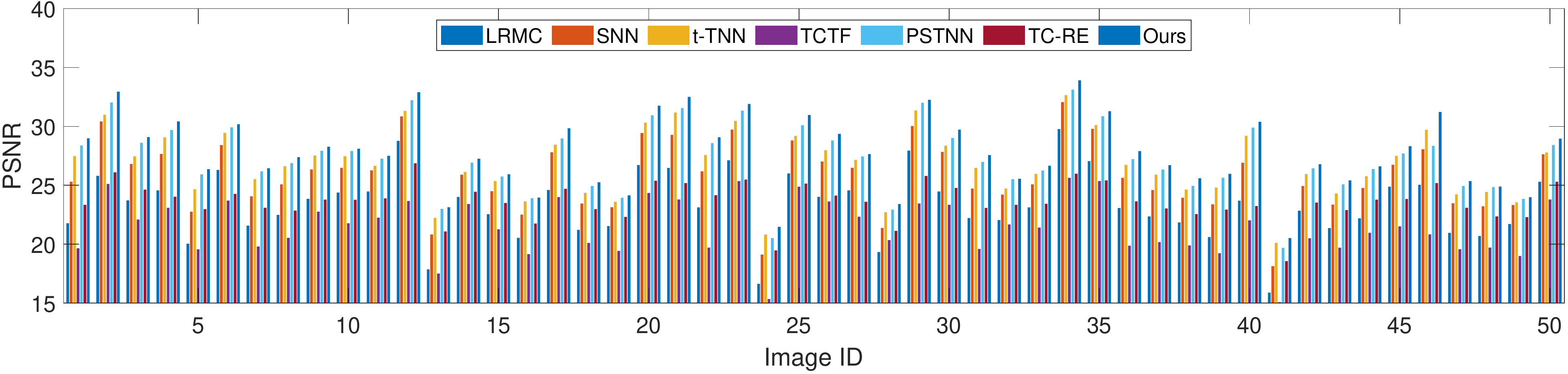} 
\caption{Comparison of the PSNR values on the randomly selected 50 images of the Berkeley Segmentation Dataset.}\label{segment_result}
\end{figure*}

\begin{figure*}

	\subfigure[Original]{
		\begin{minipage}[b]{0.12\linewidth}
			\includegraphics[width=1\textwidth]{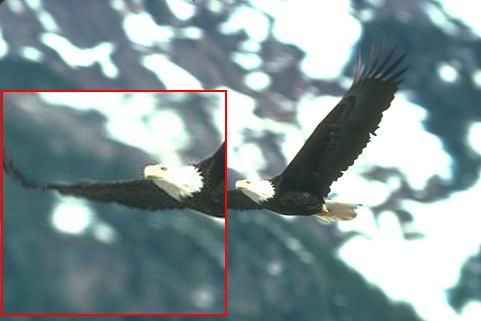}\vskip 2pt
			\includegraphics[width=1\textwidth]{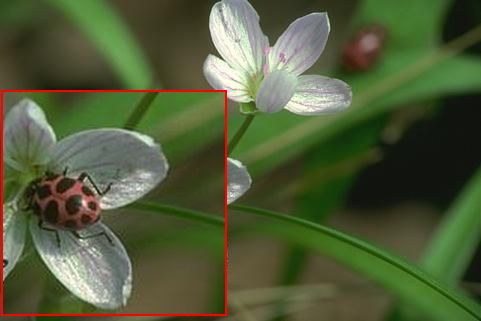}\vskip 2pt
			\includegraphics[width=1\textwidth]{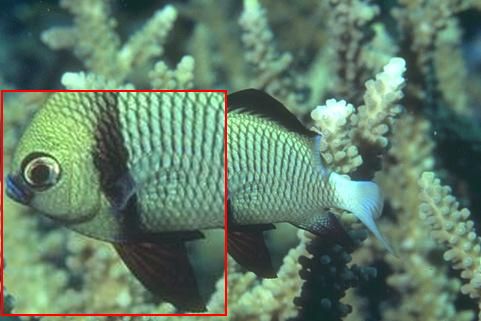}		\end{minipage}
		\hspace{-12pt}
	}
 \subfigure[LRMC \cite{36}]{
		\begin{minipage}[b]{0.12\linewidth}
			\includegraphics[width=1\textwidth]{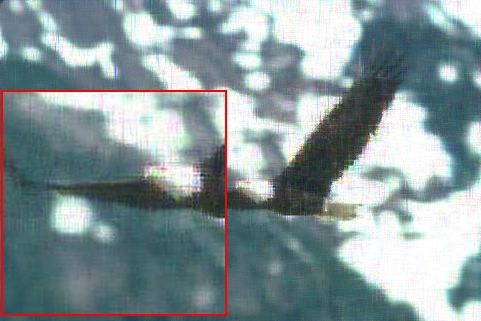}\vskip 2pt
			\includegraphics[width=1\textwidth]{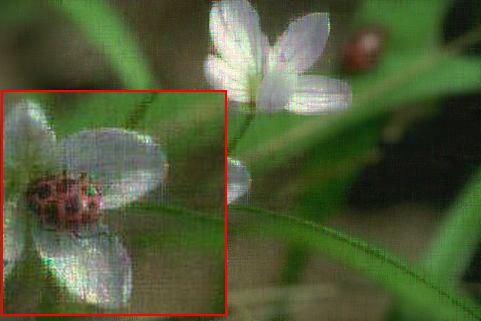}\vskip 2pt
			\includegraphics[width=1\textwidth]{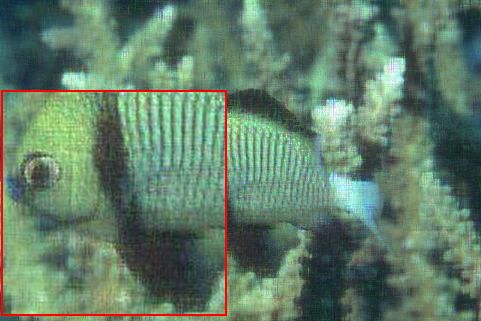}		\end{minipage}
		\hspace{-12pt}
	} 
	\subfigure[SNN \cite{12}]{
		\begin{minipage}[b]{0.12\linewidth}
			\includegraphics[width=1\textwidth]{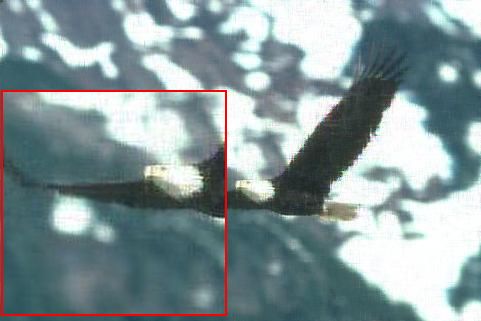}\vskip 2pt
			\includegraphics[width=1\textwidth]{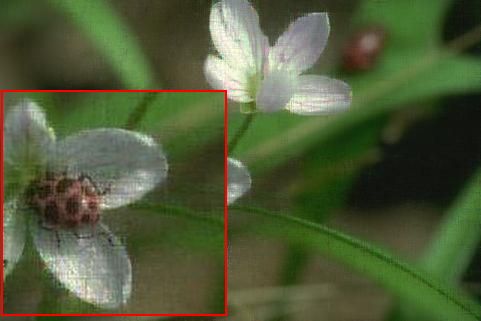}\vskip 2pt
			\includegraphics[width=1\textwidth]{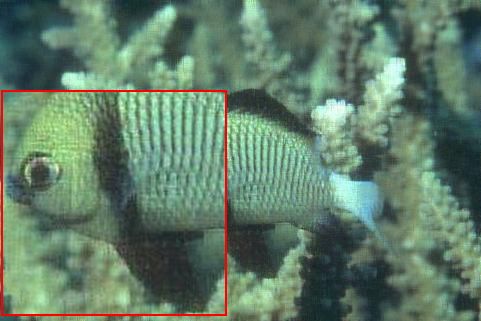}
		\end{minipage}
		\hspace{-12pt}
	} 
   	\subfigure[t-TNN \cite{37}]{
		\begin{minipage}[b]{0.12\linewidth}
			\includegraphics[width=1\textwidth]{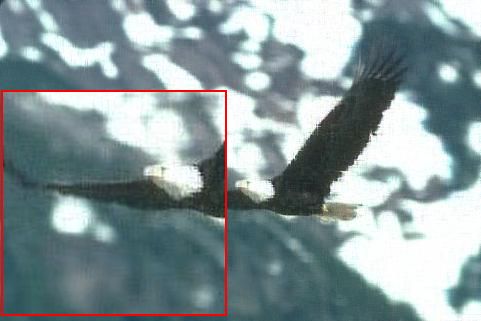}\vskip 2pt
			\includegraphics[width=1\textwidth]{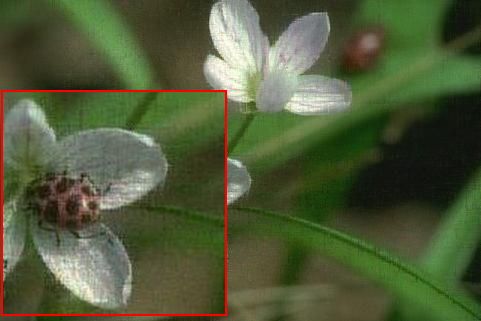}\vskip 2pt
			\includegraphics[width=1\textwidth]{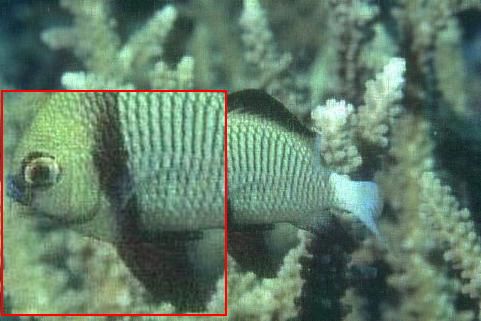}
		\end{minipage}
		\hspace{-12pt}
	}
	\subfigure[TCTF \cite{29}]{
		\begin{minipage}[b]{0.12\linewidth}
			\includegraphics[width=1\textwidth]{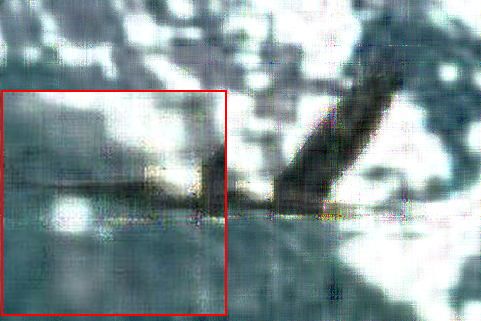}\vskip 2pt
			\includegraphics[width=1\textwidth]{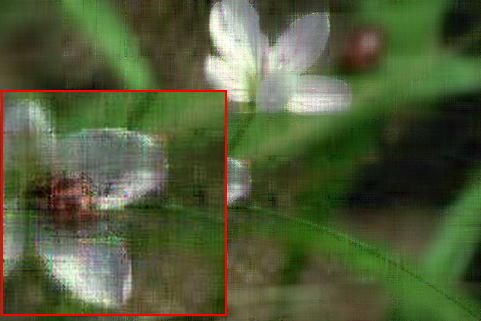}\vskip 2pt
			\includegraphics[width=1\textwidth]{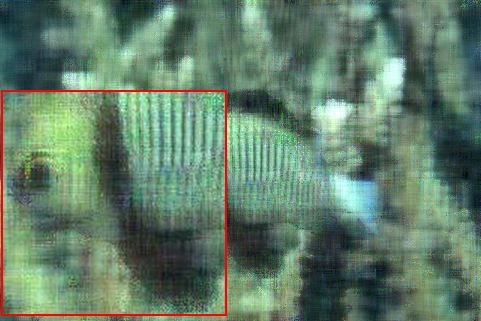}		\end{minipage}
		\hspace{-12pt}
	}
 \subfigure[PSTNN\cite{jiang2020multi}]{
		\begin{minipage}[b]{0.12\linewidth}
			\includegraphics[width=1\textwidth]{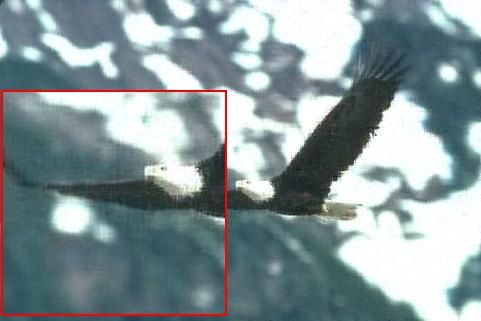}\vskip 2pt
			\includegraphics[width=1\textwidth]{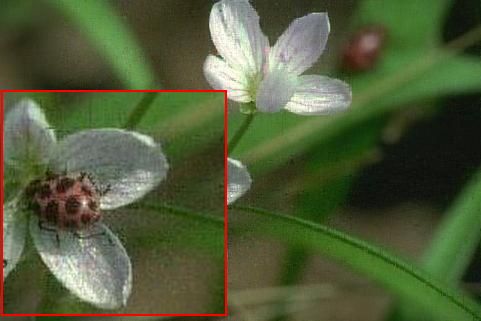}\vskip 2pt
			\includegraphics[width=1\textwidth]{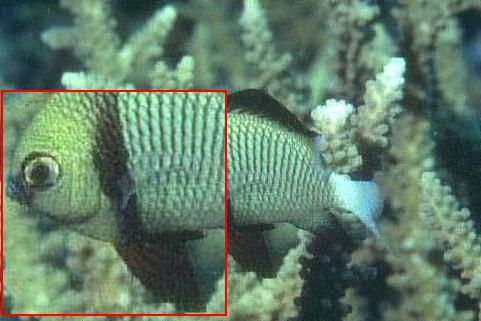}
		\end{minipage}
		\hspace{-12pt}
  }
 \subfigure[TC-RE\cite{shi2021robust}]{
		\begin{minipage}[b]{0.12\linewidth}
			\includegraphics[width=1\textwidth]{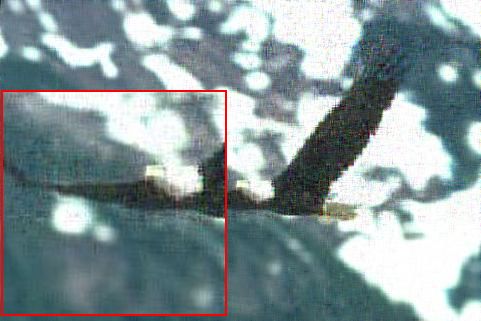}\vskip 2pt
			\includegraphics[width=1\textwidth]{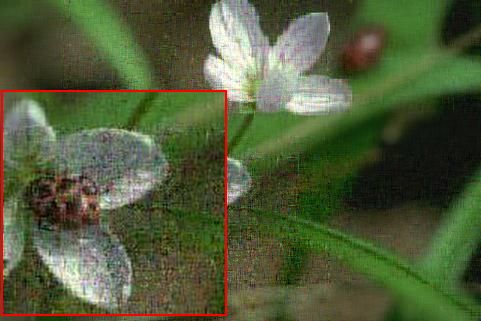}\vskip 2pt
			\includegraphics[width=1\textwidth]{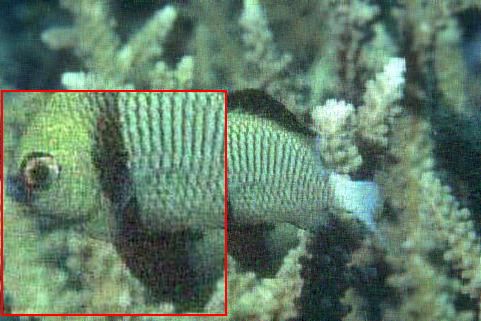}
		\end{minipage}
		\hspace{-12pt}
	}
 \subfigure[TCDLR-RE]{
		\begin{minipage}[b]{0.12\linewidth}
			\includegraphics[width=1\textwidth]{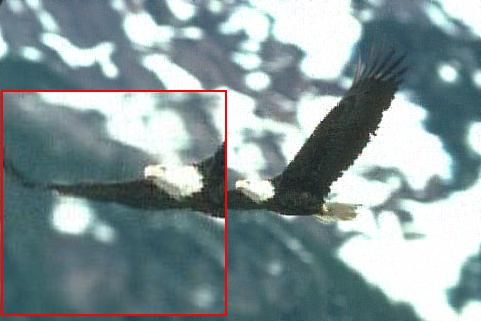}\vskip 2pt
			\includegraphics[width=1\textwidth]{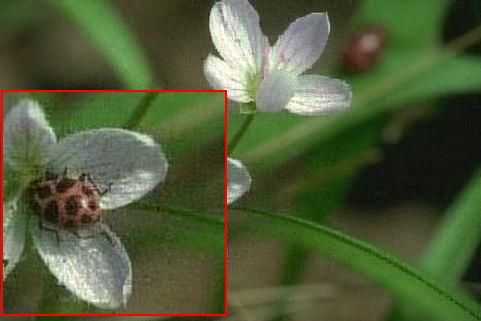}\vskip 2pt
			\includegraphics[width=1\textwidth]{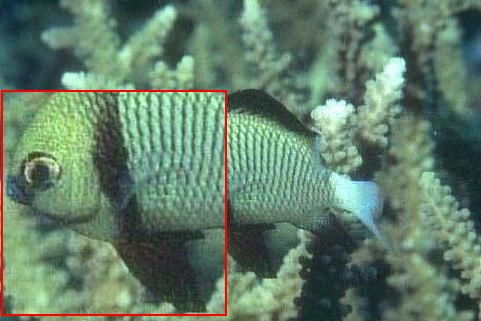}
		\end{minipage}
		\hspace{-12pt}
	}

	\caption{ Completion of the visual results on the Berkeley Segmentation Dataset with a sampling rate of 30\%:  (a) The original image and the results by different methods including (b) LRMC,
 (c) SNN, (d) t-TNN, (e) TCTF, (f) PSTNN, (g) TC-RE, and (h) TCDLR-RE, respectively. }\label{comparedimages}
\end{figure*}
 \begin{table*}[]
\centering
\caption{Comparison of the MPSNR and total time (seconds) on the 50 randomly selected images with a sample rate of 30\% on the Berkeley Segmentation Dataset.}
	\scalebox{0.7}{
\resizebox{\textwidth}{!}{%
\begin{tabular}{cccccccc}
\hline
           & LRMC\cite{36}   & SNN\cite{12}     & t-TNN\cite{37}  & TCTF\cite{29}     & PSTNN\cite{jiang2020multi}   &TC-RE\cite{shi2021robust}   &  TCDLR-RE        \\ \hline
MPSNR      & 23.26  & 25.76   & 26.78  & 21.41     & 27.32   & 23.64   & \textbf{27.83}  \\
Average time & 11.69 & 29.17 & 12.46 & 4.07  & 24.63 & 62.60 & \textbf{3.58} \\ \hline
\end{tabular}%
}}\label{smallimg}
\end{table*}

\subsection{Differences to Previous Works}
Here, this paper investigates the differences between TCDLR-RE and the earlier t-product-based tensor completion methods including t-SVD-based methods (including TNN-based methods \cite{22,37}, IRNN(T) \cite{8913528,wang2021generalized}, GTSVT-based methods \cite{Zhang2022Tensor,sun2020nonlocal}, the Laplace function-based method \cite{xu2019laplace}, and PSTNN \cite{jiang2020multi}), the tensor Schatten-p norm-based method \cite{kong2018t}, TCTF \cite{29}, and TC-RE \cite{shi2021robust}.

\subsubsection{TCDLR-RE vs. the standard t-SVD-based methods} 
There are two differences between TCDLR-RE and the t-SVD-based methods: (1) The computing of t-SVD for tensors of size $n_1\times n_2 \times n_3$ is required at each iteration in the standard t-SVD-based methods.
TCDLR-RE turns to compute the t-SVD for a smaller tensor by using the proposed tensor norm and a simple trick.  (2) In TCDLR-RE, the tensor tubal rank  information provided by the proposed rank estimation method is introduced to improve the performance of tensor recovery.

\subsubsection{TCDLR-RE vs. the three non-convex methods (including the tensor Schatten-p norm-based method, the Laplace function-based method, and PSTNN) }
There are two biggest differences between TCDLR-RE and the three methods: (1) Different from the three methods, TCDLR-RE is proposed to solve a more general problem  \eqref{TCDLR}, where $g$ can be any of the non-convex surrogate functions listed in Table \ref{tab1}. 
This paper has proven that the optimal solutions to \eqref{TClg} and \eqref{TCDLR} are equivalent when $k \geq \mathrm{rank}(\mathring{\mathcal{X}})$, where $\mathring{\mathcal{X}}$ is an optimal solution to \eqref{TClg}.
(2) Different from the tensor Schatten-p norm-based method, the tensor tubal rank information provided by the proposed rank estimation method is introduced into TCDLR-RE. 

\subsubsection{TCDLR-RE vs. TCTF and TC-RE} 
(1) Different from TCTF and TC-RE, TCDLR-RE uses a new tensor norm. Benefiting from the proposed norm, TCDLR-RE can adapt to a series of surrogate functions (possibly non-convex) to achieve better tensor recovery performance and stronger robustness to inaccurate rank estimation for $k$.  
(2)  The rank estimation methods used in TCDLR-RE, TCTF, and TC-RE are different. Different from TCTF, our method adopts an additionally developed rank-increasing scheme to avoid underestimating the tensor rank. Also, for the rank-decreasing scheme, the singular values of $\bar Z_i^{(t)}$ instead of $\bar A_i^{(t)}$ are used. Compared with TC-RE, TCDLR-RE uses the rank-increasing and decreasing scheme, while TC-RE estimates the rank by solving \eqref{ShiRe}.

\section{Experiments \label{ER}}
To verify the effectiveness and efficiency of TCDLR-RE for tensor completion, it was compared with six state-of-the-art methods, including  Low-Rank Matrix Completion 
(LRMC)\footnote{https://github.com/canyilu/LibADMM-toolbox\label{git}}\cite{36}, SNN\textsuperscript{\ref {git}}\cite{12},   t-TNN\textsuperscript{\ref {git}}\cite{37}, TCTF\footnote{https://panzhous.github.io/assets/code/TCTF\_code.rar}\cite{29}, PSTNN\footnote{https://github.com/zhaoxile/Multi-dimensional-imaging-data-recovery-via-minimizing-the-partial-sum-of-tubal-nuclear-norm} \cite{jiang2020multi},  and TC-RE \cite{shi2021robust} on both synthetic and real-world data. For TCDLR-RE, $ \ell_p $ was selected as the non-convex penalty function in $\|\mathcal{X}\|_{*, (k,g)}$, and the parameters were set as $k_{\max}=0.5\times\min(n_1,n_2)$, $k_{\min}=25$,  and $p=0.8$. For TCTF,  the parameters suggested in \cite{29} did not lead to good performance on a large dataset, so they were empirically tuned, as will be discussed later.
 The parameter settings of SNN, PSTNN, and TC-RE are consistent with the suggestions by the authors, and LRMC and t-TNN  are free parameters. For fairness, these methods were run respectively 10 times in these experiments, and the average results
were reported for each method. All the experiments were conducted on a personal computer running Windows 10 operating system and MATLAB (R2020b) (Intel Core i7-8700 3.20-GHz CPU and 16 GB memory).

\subsection{Synthetic Experiments}

 All tensor product-based methods including t-TNN, TCTF, PSTNN, TC-RE, and TCDLR-RE were tested with synthetic data.
The tensors of size $n\times n\times 3$ with varying $ n=\{1000, 2000, 3000, 4000\}$ were considered.
The low-rank tensor data  $ \mathcal{M} \in \mathbb{R}^{n\times n\times 3} $ were generated with a tensor tubal rank $\bar r$ by  $ \mathcal{M}=\mathcal{M}_1* \mathcal{M}_2 $, where the entries of $ \mathcal{M}_1\in \mathbb{R}^{n\times \bar{r}\times 3} $ and $ \mathcal{M}_2\in \mathbb{R}^{\bar{r}\times n \times 3} $ were independently sampled from an  $\mathcal{N}(0,1)$. Then, $  3cn_1n_2  $ elements of $ \mathcal{M} $ were sampled uniformly to construct $ \mathbf{P}_{\Omega}(\mathcal{M}) $, where $c$ is the sampling rate. This paper takes the relative error (${\tt relerr}$) $$ {\tt relerr}= \parallel \mathcal{\hat X}-\mathcal{M}\parallel^2_F /\parallel \mathcal{M} \parallel^2_F $$ and the running time to evaluate the effectiveness and efficiency of different algorithms, where $ \mathcal{\hat X} $ is the recovered tensor of $\mathbf{P}_{\Omega}(\mathcal{M}) $. Following \cite{29}, the initialized rank $ k^{(0)}=[1.5\bar{r}, 1.5 \bar{r}, 1.5 \bar{r}] $ was set for TCTF and TCDLR-RE. All experimental results are presented in Table \ref{table:synthetic_result} and Figs. \ref{loss}-\ref{aaa}.


 In Table \ref{table:synthetic_result}, our methods (TCDLR and TCDLR-RE) are compared with other four tensor-product-based methods in terms of the running time and ${\tt relerr}$, where $k$ in TCDLR is set to $0.5n$ according to the low tubal rank prior. The best two results for each case are shown in bold. 
It can be seen from Table \ref{table:synthetic_result} and Fig. \ref{loss} that:  
\begin{itemize}
    \item[(1)]  In all cases, TCDLR-RE and TCDLR achieve the best performance on ${\tt relerr}$ attributed to the proposed dual low-rank constraint and rank estimation. The superiority of TCDLR on ${\tt relerr}$ demonstrates its robustness to $k$. Besides, as shown in Fig. \ref{loss} (c), for TCTF, the singular values $[\mathcal{S}]_{i,i,1}$ ($i=1,2,\cdots,n$) decrease significantly at $i=2$ and $i=72$, and for TC-RE and TCDLR-RE, the singular values decrease significantly at  $i=\bar r$, which indicates that the proposed TCDLR-RE and TC-RE estimated the tensor tubal rank more accurately than TCTF. In addition, owing to introducing the proposed dual low-rank constraint, the singular values $[\mathcal{S}]_{i,i,1}$ ($i>\bar r$) of TCDLR-RE are much close to zeros than those of TC-RE.   

 \item[(2)]   Table \ref{table:synthetic_result} shows that TCDLR-RE and TCTF achieved the best performance in terms of running time, and this is because they have the same low computational complexity ($\mathcal{O}(kn_1n_2n_3)$) for each iteration. Attributed to the proposed rank estimation strategy, TCDLR-RE performed better in running time than TCDLR. The running time of TCDLR is about two times that of TCDLR-RE. Besides,   Fig. \ref{loss} (a)-(b) show that less running time is obtained is not only because TCDLR-RE has low computational complexity for each iteration, but also because it converges to the true solution $\mathcal{M}$ with fewer iterations than other methods.

\end{itemize} 
 
 In Fig.~\ref{aaa}, TCDLR-RE is compared with t-TNN and TC-RE which achieved better performance on ${\tt relerr}$ than other compared methods.  
Fig.~\ref{aaa} presents the results of ${\tt relerr}$ with varying $c$ and $\bar r$ for fixed $n=1000$. It was determined that a trial is successful if $ {\tt relerr}\leq 10^{-2} $, \emph{i.e.,} the cases corresponding  to the white regions were regarded as successful recovery to $ \mathcal{M} $. It can be seen from Fig \ref{aaa} that the region of correct recovery in Fig \ref{aaa} (c)   is broader than that in Fig. \ref{aaa} (a)-(b). These results demonstrate the effectiveness and efficiency of TCDLR-RE.  


\begin{table*}[]
\centering
	\caption{Comparison of the PSNR and running time (seconds) on 10 randomly chosen images from the DOTA-v2.0 Dataset with a sampling rate of 30\%.}
	\scalebox{1}{
\resizebox{\textwidth}{!}{%
\begin{tabular}{c|cc|cc|cc|cc|cc|cc|cc}
\hline
\multirow{2}{*}{images} &
  \multicolumn{2}{c|}{LRMC\cite{36}} &
  \multicolumn{2}{c|}{SNN\cite{12}} &
  \multicolumn{2}{c|}{t-TNN\cite{37}} &
  \multicolumn{2}{c|}{TCTF\cite{29}} & 
  \multicolumn{2}{c|}{PSTNN\cite{jiang2020multi}} &
  \multicolumn{2}{c|}{TC-RE\cite{shi2021robust}} &
  \multicolumn{2}{c}{TCDLR-RE} \\ \cline{2-15} 
 &
  \multicolumn{1}{c|}{PSNR} &
  time(s) &
  \multicolumn{1}{c|}{PSNR} &
  time(s) &
  \multicolumn{1}{c|}{PSNR} &
  time(s) &
  \multicolumn{1}{c|}{PSNR} &
  time(s) & 
  \multicolumn{1}{c|}{PSNR} &
  time(s) &
  \multicolumn{1}{c|}{PSNR} &
  time(s) &
  \multicolumn{1}{c|}{PSNR} &
  time(s) \\ \hline
1 &
  \multicolumn{1}{c|}{24.93} &
  137.21 &
  \multicolumn{1}{c|}{27.59} &
  209.22 &
  \multicolumn{1}{c|}{28.98} &
  163.61 &
  \multicolumn{1}{c|}{25.91} &
  28.09  & 
  \multicolumn{1}{c|}{29.34} &
  347.02 &
  \multicolumn{1}{c|}{27.71} &
  1113.40 &
  \multicolumn{1}{c|}{\textbf{29.94}} &
  \textbf{24.70} \\
2 &
  \multicolumn{1}{c|}{26.94} &
  128.35 &
  \multicolumn{1}{c|}{29.43} &
  198.85 &
  \multicolumn{1}{c|}{30.68} &
  161.81 &
  \multicolumn{1}{c|}{27.48} &
  28.08  & 
  \multicolumn{1}{c|}{30.92} &
  320.25 &
  \multicolumn{1}{c|}{29.22} &
  1015.84 &
  \multicolumn{1}{c|}{\textbf{31.75}} &
  \textbf{24.93} \\
3 &
  \multicolumn{1}{c|}{24.54} &
  138.01 &
  \multicolumn{1}{c|}{26.98} &
  202.09 &
  \multicolumn{1}{c|}{28.17} &
  167.40 &
  \multicolumn{1}{c|}{25.14} &
  28.19  & 
  \multicolumn{1}{c|}{28.32} &
  314.94 &
  \multicolumn{1}{c|}{27.05} &
  1111.42 &
  \multicolumn{1}{c|}{\textbf{28.79}} &
  \textbf{24.47} \\
4 &
  \multicolumn{1}{c|}{29.19} &
  129.75 &
  \multicolumn{1}{c|}{32.56} &
  201.79 &
  \multicolumn{1}{c|}{34.15} &
  149.11 &
  \multicolumn{1}{c|}{29.18} &
  27.75  & 
  \multicolumn{1}{c|}{34.83} &
  314.63 &
  \multicolumn{1}{c|}{31.04} &
  918.00 &
  \multicolumn{1}{c|}{\textbf{36.08}} &
  \textbf{24.50} \\
5 &
  \multicolumn{1}{c|}{24.53} &
  139.80 &
  \multicolumn{1}{c|}{27.37} &
  211.15 &
  \multicolumn{1}{c|}{29.12} &
  184.40 &
  \multicolumn{1}{c|}{24.58} &
  27.70  & 
  \multicolumn{1}{c|}{29.56} &
  314.76 &
  \multicolumn{1}{c|}{27.82} &
  1106.52 &
  \multicolumn{1}{c|}{\textbf{30.17}} &
  \textbf{24.36} \\
6 &
  \multicolumn{1}{c|}{27.60} &
  132.55 &
  \multicolumn{1}{c|}{30.11} &
  189.97 &
  \multicolumn{1}{c|}{31.34} &
  165.90 &
  \multicolumn{1}{c|}{26.86} &
  27.85  & 
  \multicolumn{1}{c|}{31.44} &
  314.75 &
  \multicolumn{1}{c|}{29.66} &
  1024.86 &
  \multicolumn{1}{c|}{\textbf{32.55}} &
  \textbf{24.58} \\
7 &
  \multicolumn{1}{c|}{26.00} &
  136.96 &
  \multicolumn{1}{c|}{28.23} &
  189.99 &
  \multicolumn{1}{c|}{29.20} &
  168.21 &
  \multicolumn{1}{c|}{26.47} &
  27.45  & 
  \multicolumn{1}{c|}{29.41} &
  314.46 &
  \multicolumn{1}{c|}{27.98} &
  1114.94 &
  \multicolumn{1}{c|}{\textbf{29.85}} &
  \textbf{24.71} \\
8 &
  \multicolumn{1}{c|}{27.59} &
  133.63 &
  \multicolumn{1}{c|}{29.99} &
  189.82 &
  \multicolumn{1}{c|}{30.98} &
  172.42 &
  \multicolumn{1}{c|}{28.04} &
  27.43  & 
  \multicolumn{1}{c|}{31.15} &
  314.90 &
  \multicolumn{1}{c|}{29.59} &
  1016.91 &
  \multicolumn{1}{c|}{\textbf{31.83}} &
  \textbf{24.60} \\
9 &
  \multicolumn{1}{c|}{24.25} &
  128.18 &
  \multicolumn{1}{c|}{26.88} &
  193.03 &
  \multicolumn{1}{c|}{28.82} &
  165.67 &
  \multicolumn{1}{c|}{25.23} &
  27.61  & 
  \multicolumn{1}{c|}{29.30} &
  314.40 &
  \multicolumn{1}{c|}{27.53} &
  1111.54 &
  \multicolumn{1}{c|}{\textbf{30.10}} &
  \textbf{24.36} \\
10 &
  \multicolumn{1}{c|}{25.33} &
  135.28 &
  \multicolumn{1}{c|}{28.30} &
  189.08 &
  \multicolumn{1}{c|}{29.46} &
  156.05 &
  \multicolumn{1}{c|}{25.57} &
  27.42  & 
  \multicolumn{1}{c|}{29.70} &
  314.06 &
  \multicolumn{1}{c|}{28.38} &
  1073.92 &
  \multicolumn{1}{c|}{\textbf{30.59}} &
  \textbf{24.36} \\ \hline
Average &
  \multicolumn{1}{c|}{26.09} &
  133.97 &
  \multicolumn{1}{c|}{28.74} &
  197.50 &
  \multicolumn{1}{c|}{30.09} &
  165.46 &
  \multicolumn{1}{c|}{26.44} &
  27.76  & 
  \multicolumn{1}{c|}{30.40} &
  318.42 &
  \multicolumn{1}{c|}{28.60} &
  1060.74 &
  \multicolumn{1}{c|}{\textbf{31.16}} &
  \textbf{24.56} \\ \hline
\end{tabular}%
}}\label{DOTA_table}
\end{table*}
\begin{figure*}
\subfigure[Original]{
		\begin{minipage}[b]{0.12\linewidth}
			\includegraphics[width=1\textwidth]{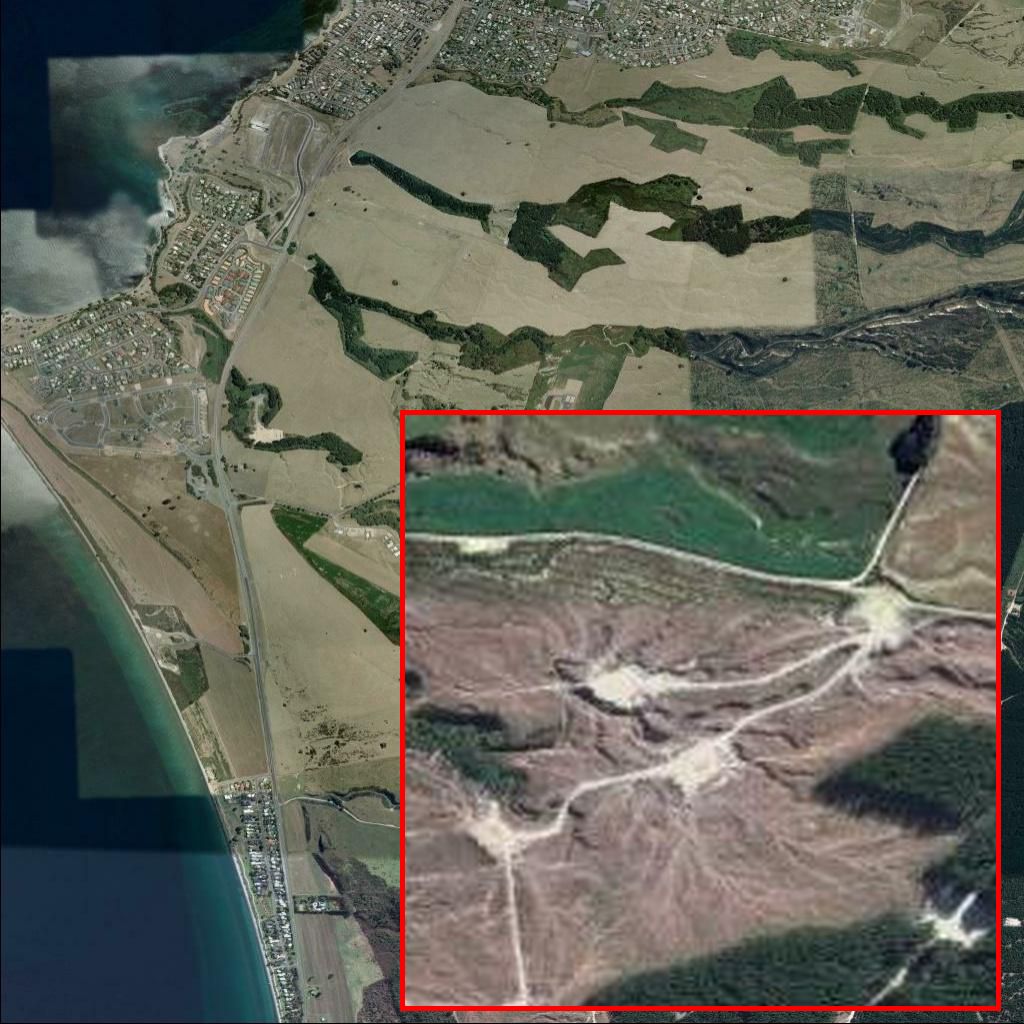}\vskip 2pt
			\includegraphics[width=1\textwidth]{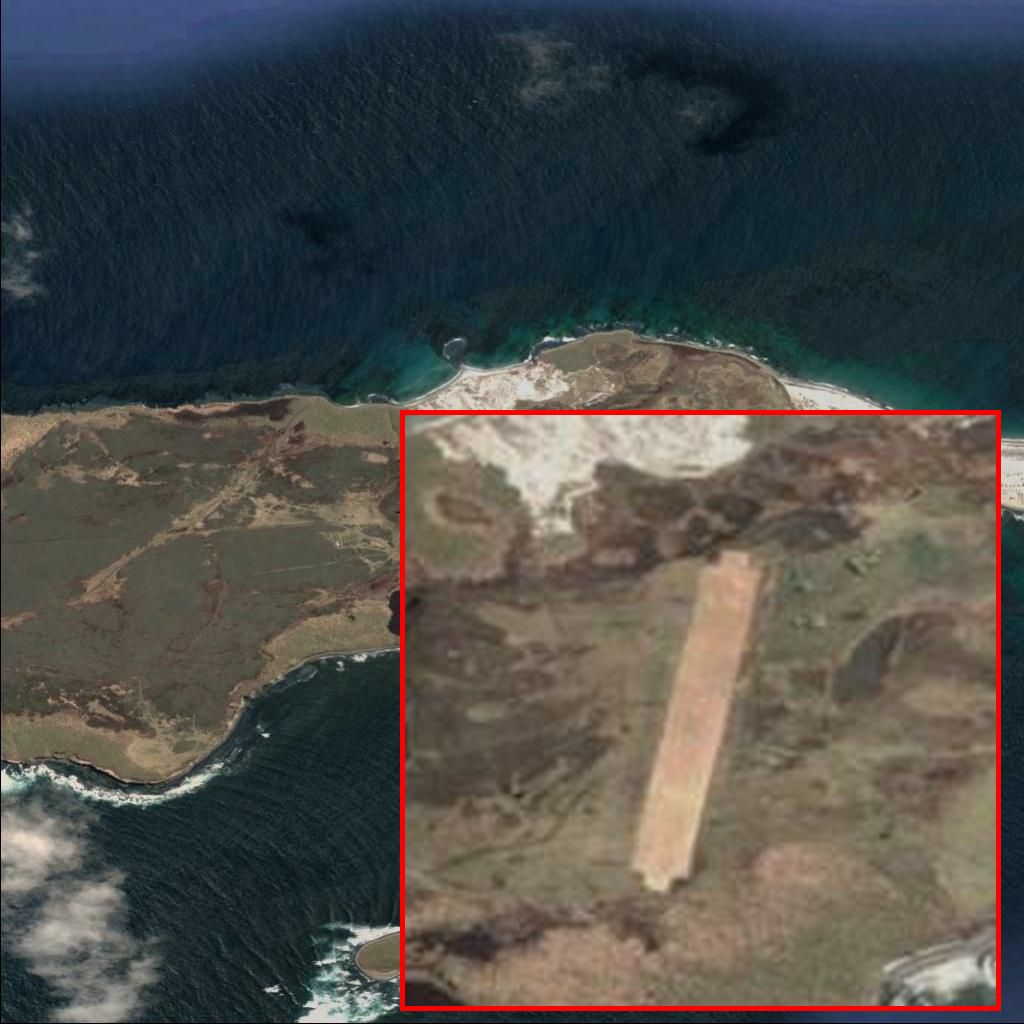}
					\end{minipage}
		\hspace{-12pt}
	}
\subfigure[LRMC\cite{36}]{
		\begin{minipage}[b]{0.12\linewidth}
			\includegraphics[width=1\textwidth]{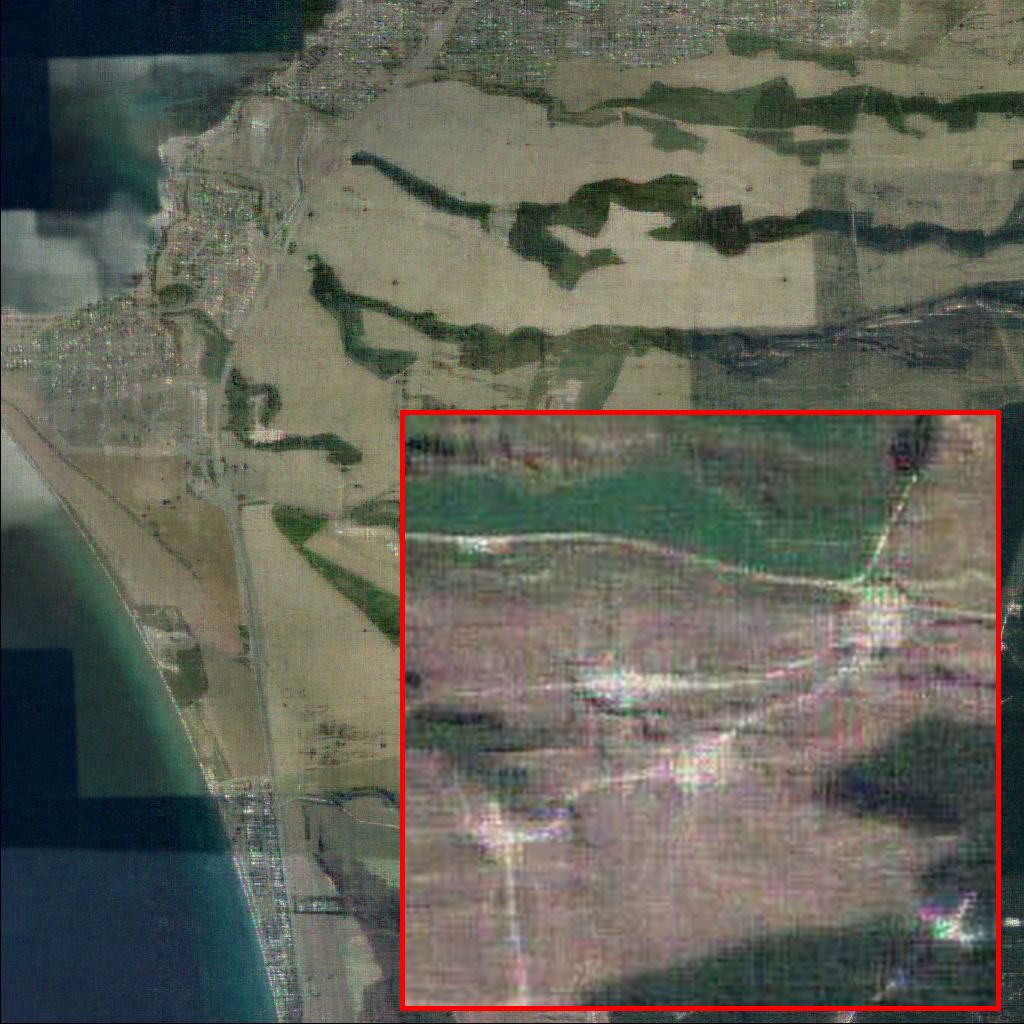}\vskip 2pt
			\includegraphics[width=1\textwidth]{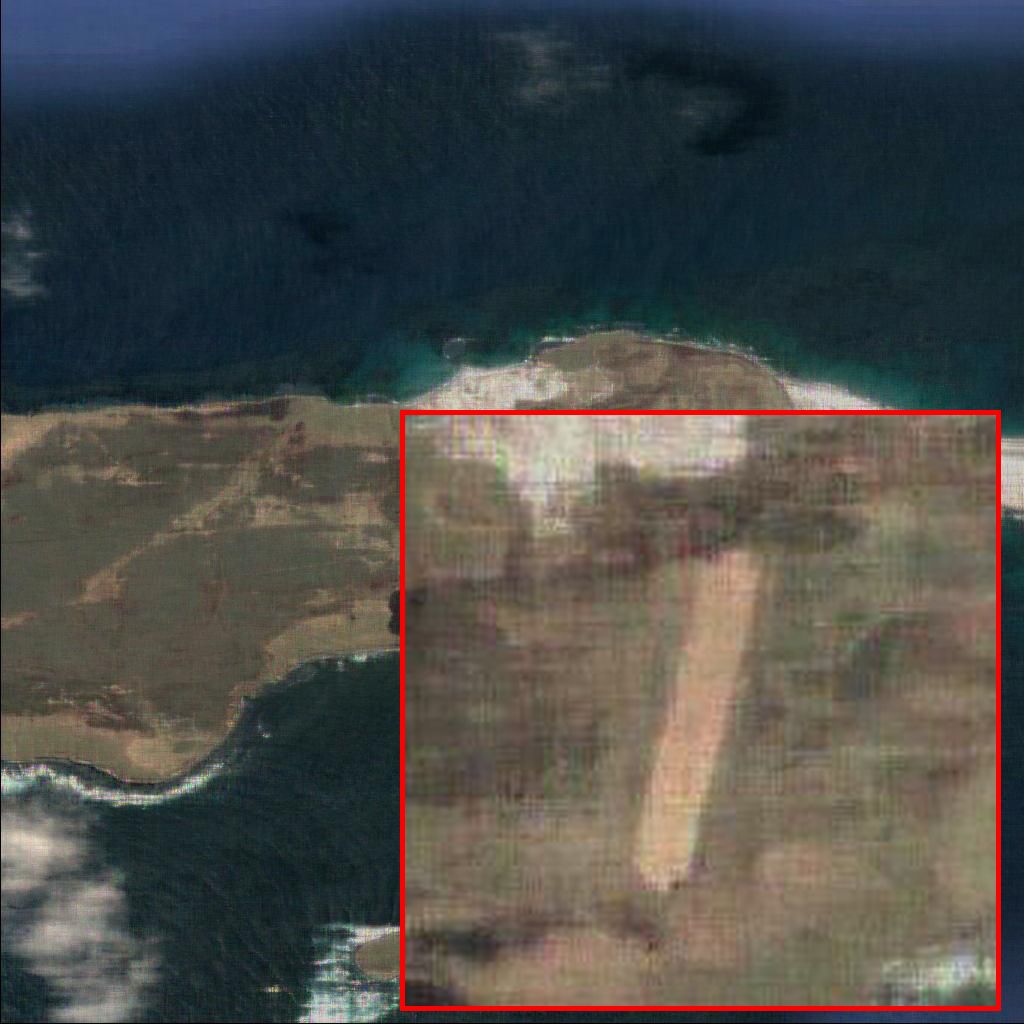}
			\end{minipage}
		\hspace{-12pt}
	} 
        \subfigure[SNN \cite{12}]{
		\begin{minipage}[b]{0.12\linewidth}
			\includegraphics[width=1\textwidth]{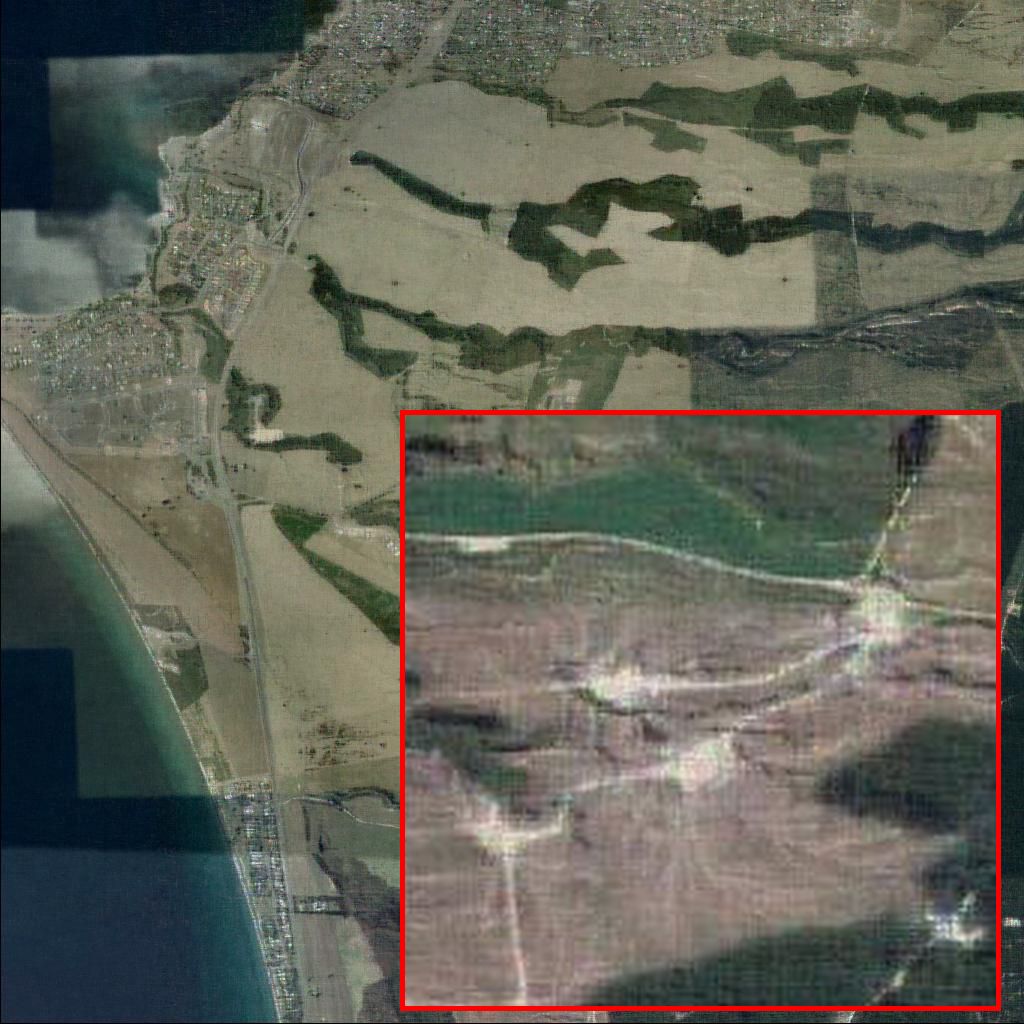}\vskip 2pt
			\includegraphics[width=1\textwidth]{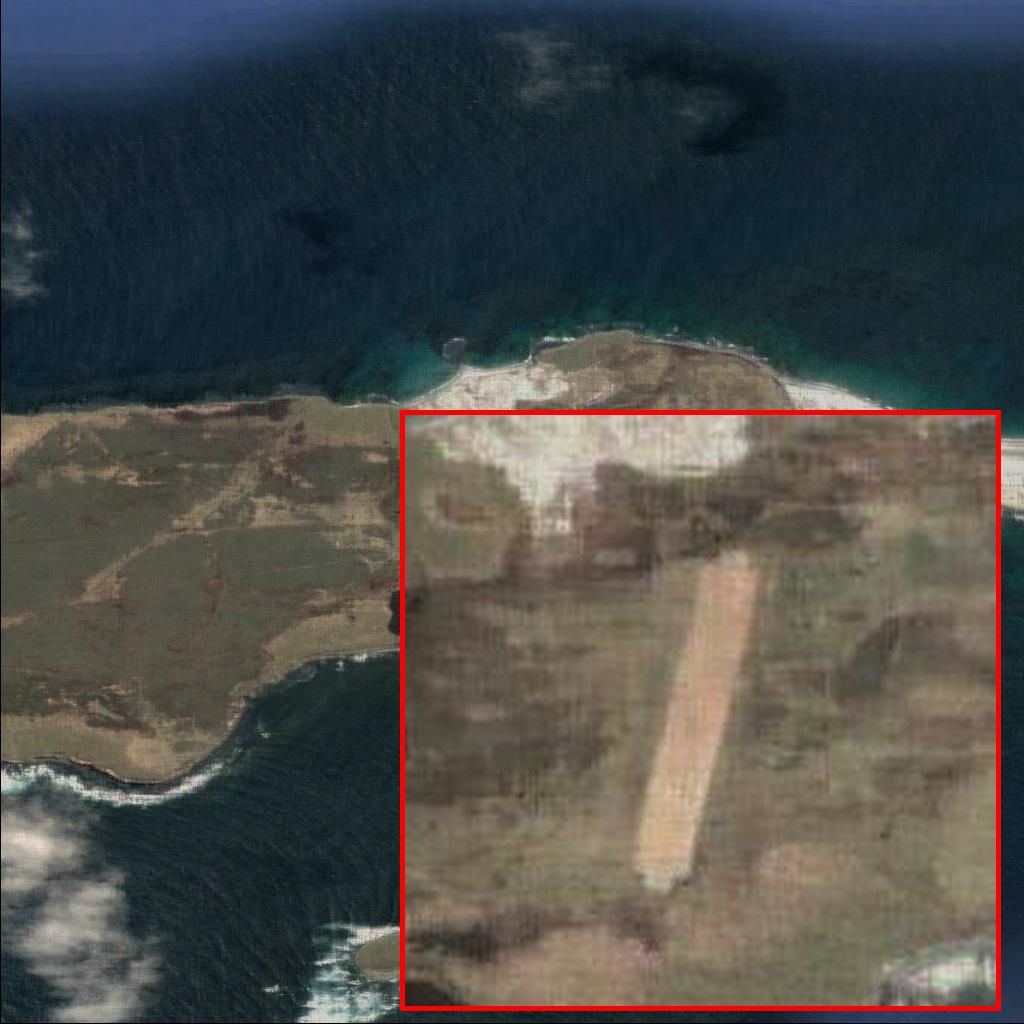}
					\end{minipage}
		\hspace{-12pt}
	} 
 \subfigure[t-TNN \cite{37}]{
		\begin{minipage}[b]{0.12\linewidth}
			\includegraphics[width=1\textwidth]{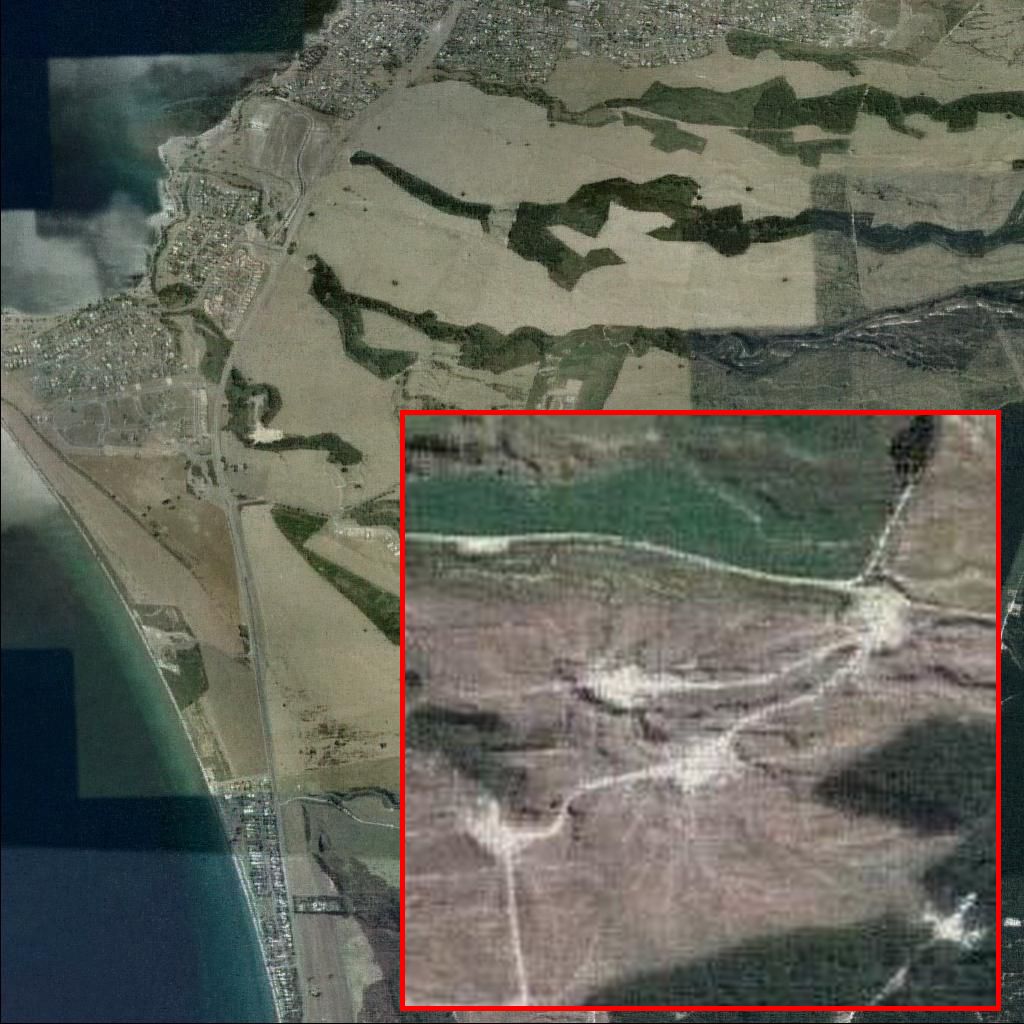}\vskip 2pt
			\includegraphics[width=1\textwidth]{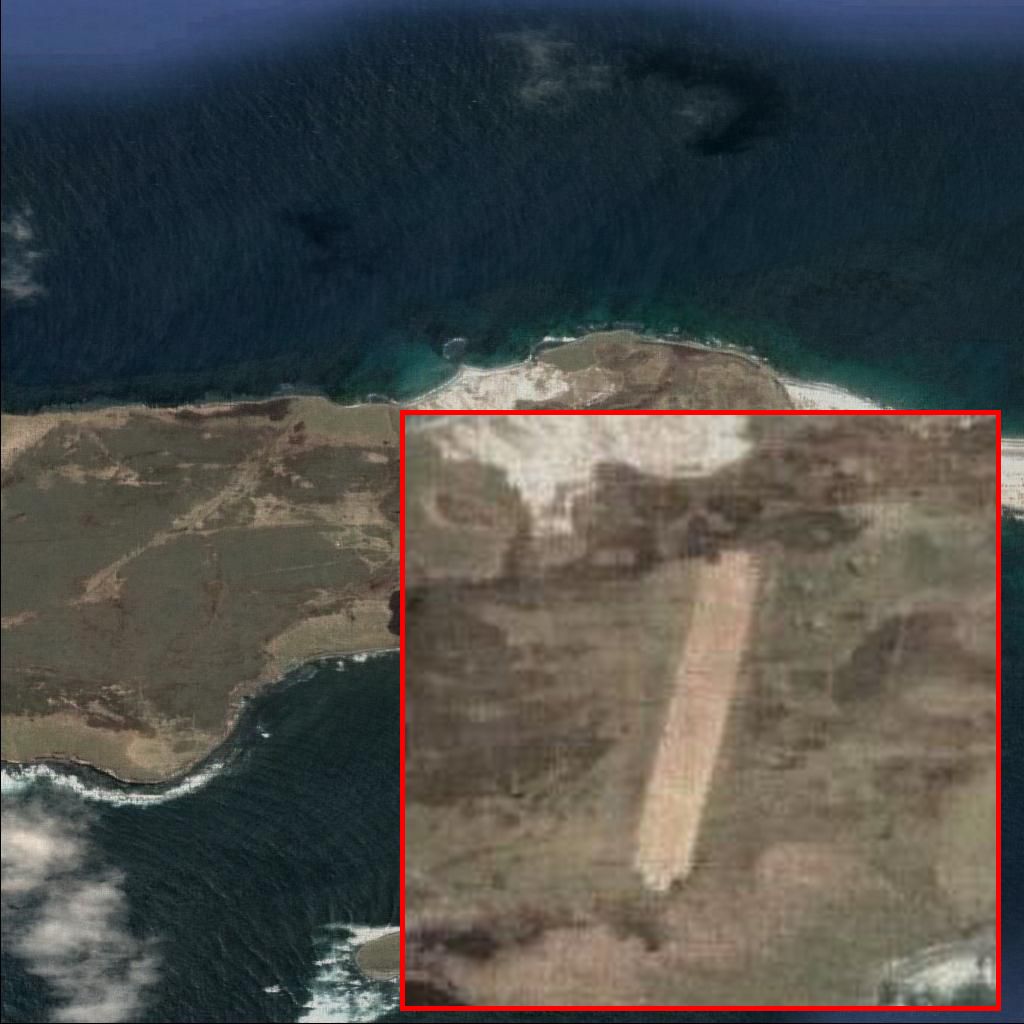}
		\end{minipage}
		\hspace{-12pt}
	}
        \subfigure[TCTF \cite{29}]{
		\begin{minipage}[b]{0.12\linewidth}
			\includegraphics[width=1\textwidth]{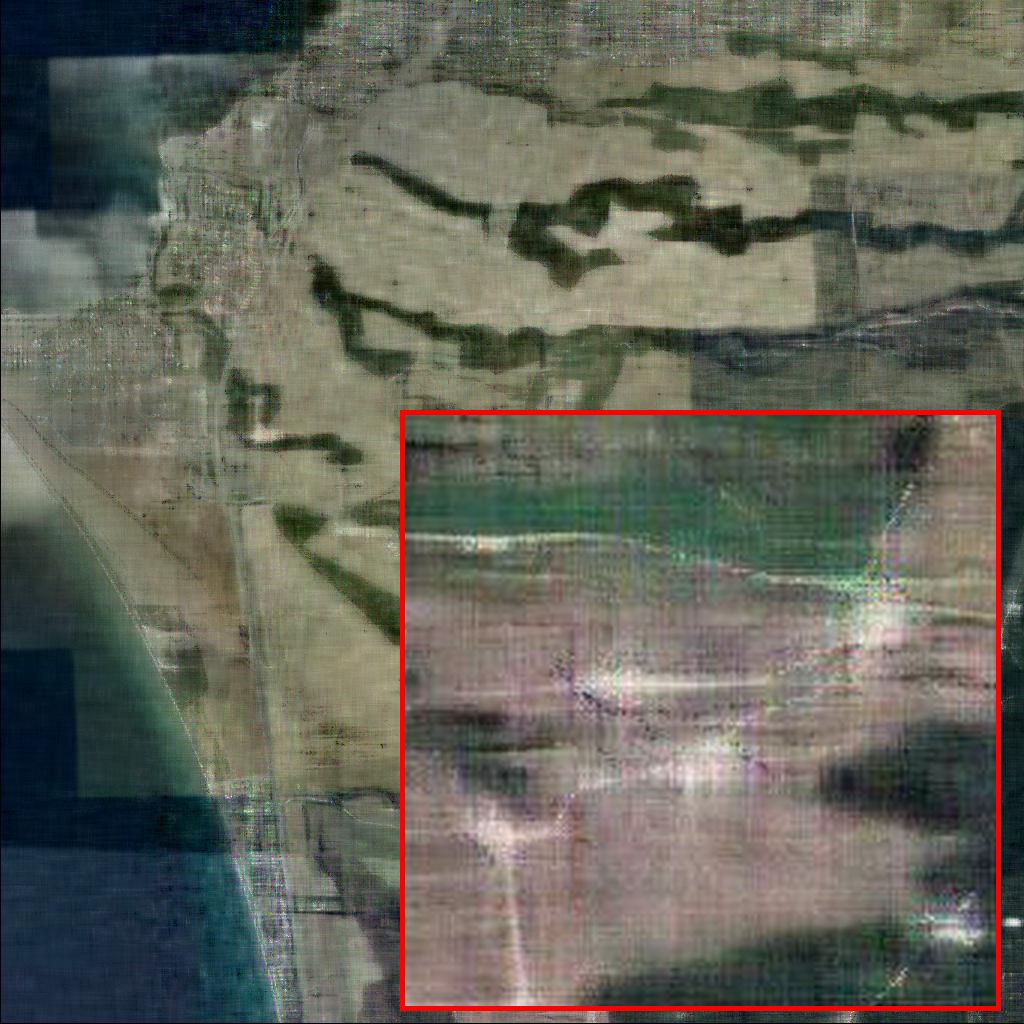}\vskip 2pt
			\includegraphics[width=1\textwidth]{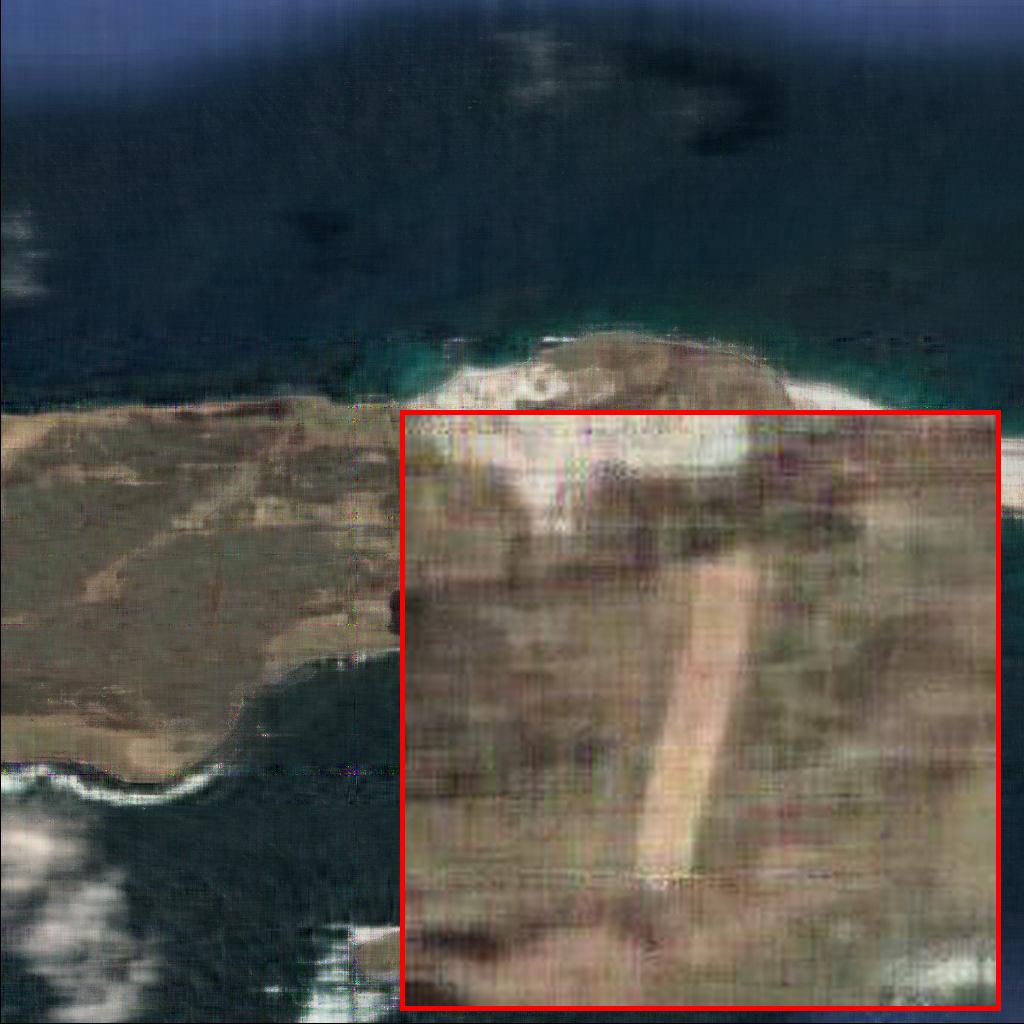}
					\end{minipage}
		\hspace{-12pt}
	} \subfigure[PSTNN\cite{jiang2020multi}]{
		\begin{minipage}[b]{0.12\linewidth}
			\includegraphics[width=1\textwidth]{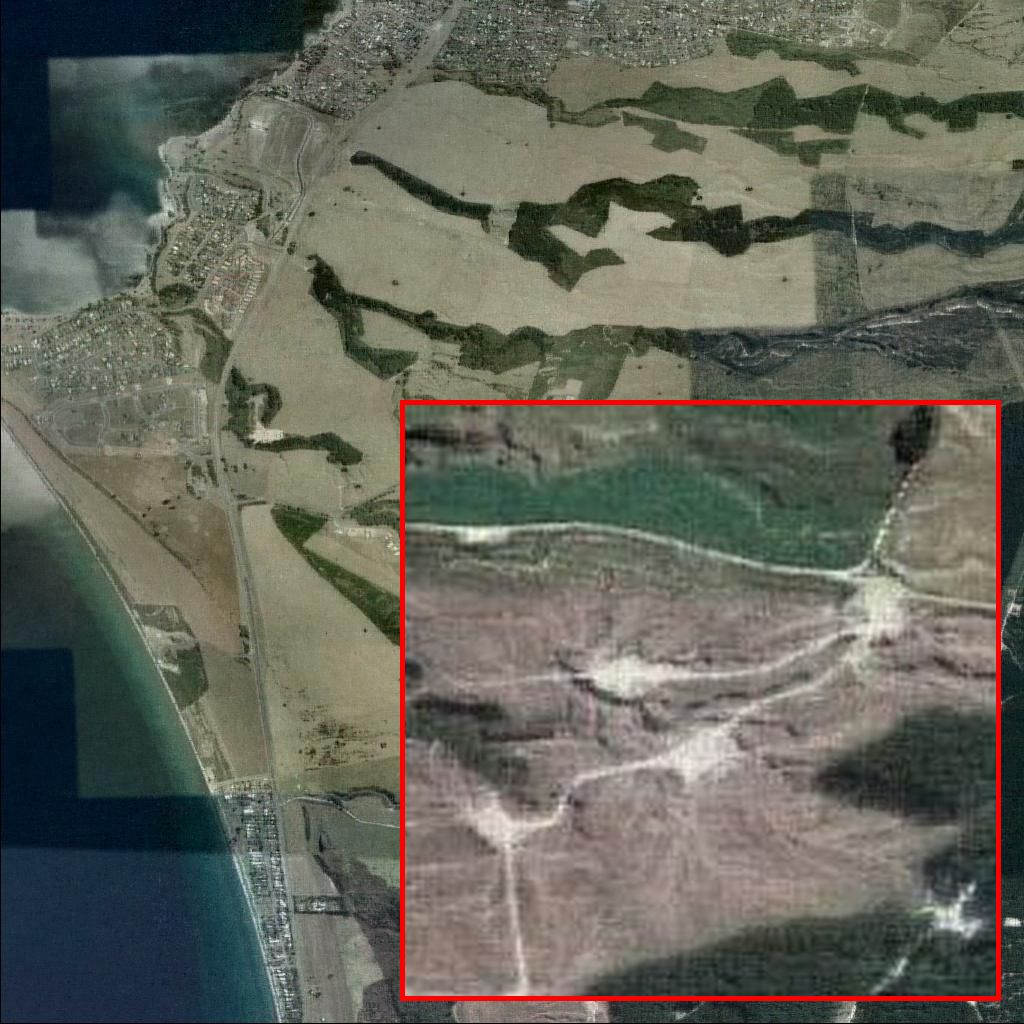}\vskip 2pt
			\includegraphics[width=1\textwidth]{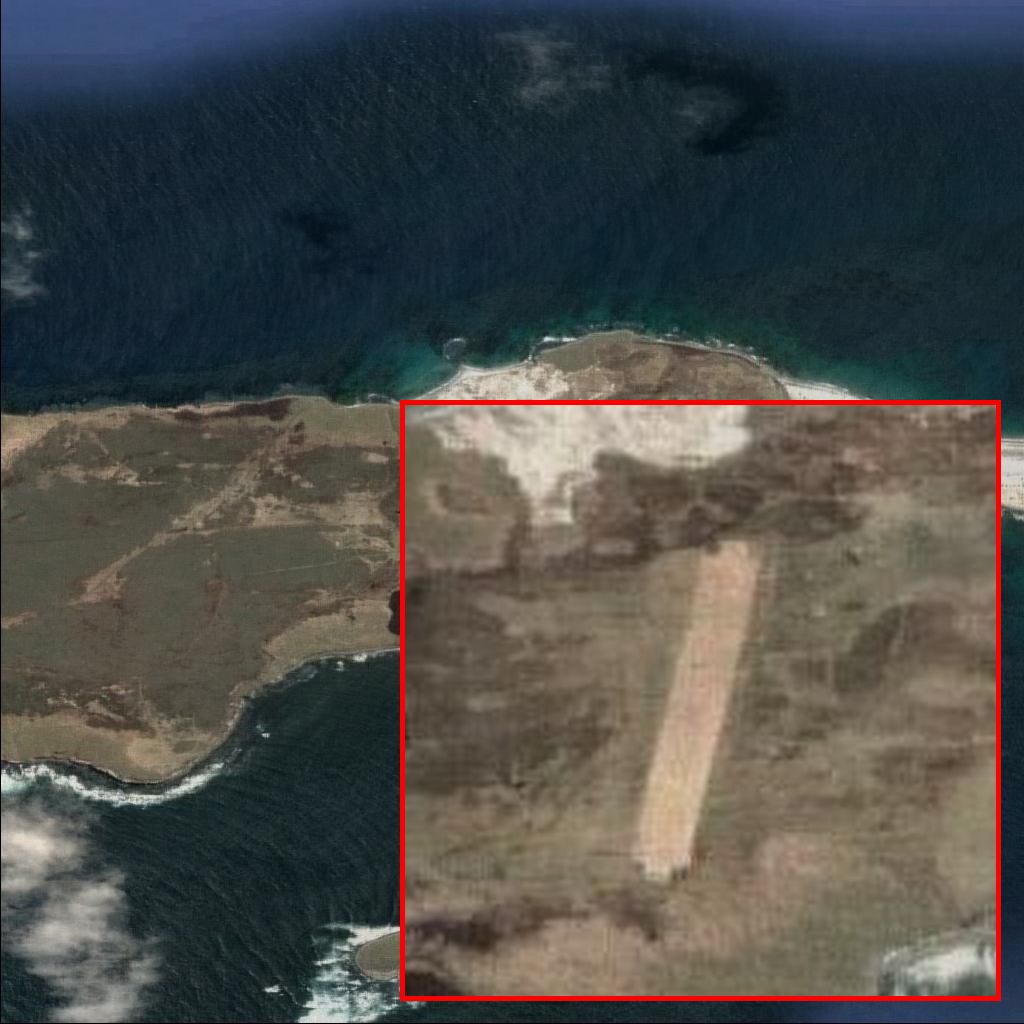}
					\end{minipage}
		\hspace{-12pt}
	}
	\subfigure[TC-RE\cite{shi2021robust}]{
		
		\begin{minipage}[b]{0.12\linewidth}
			\includegraphics[width=1\textwidth]{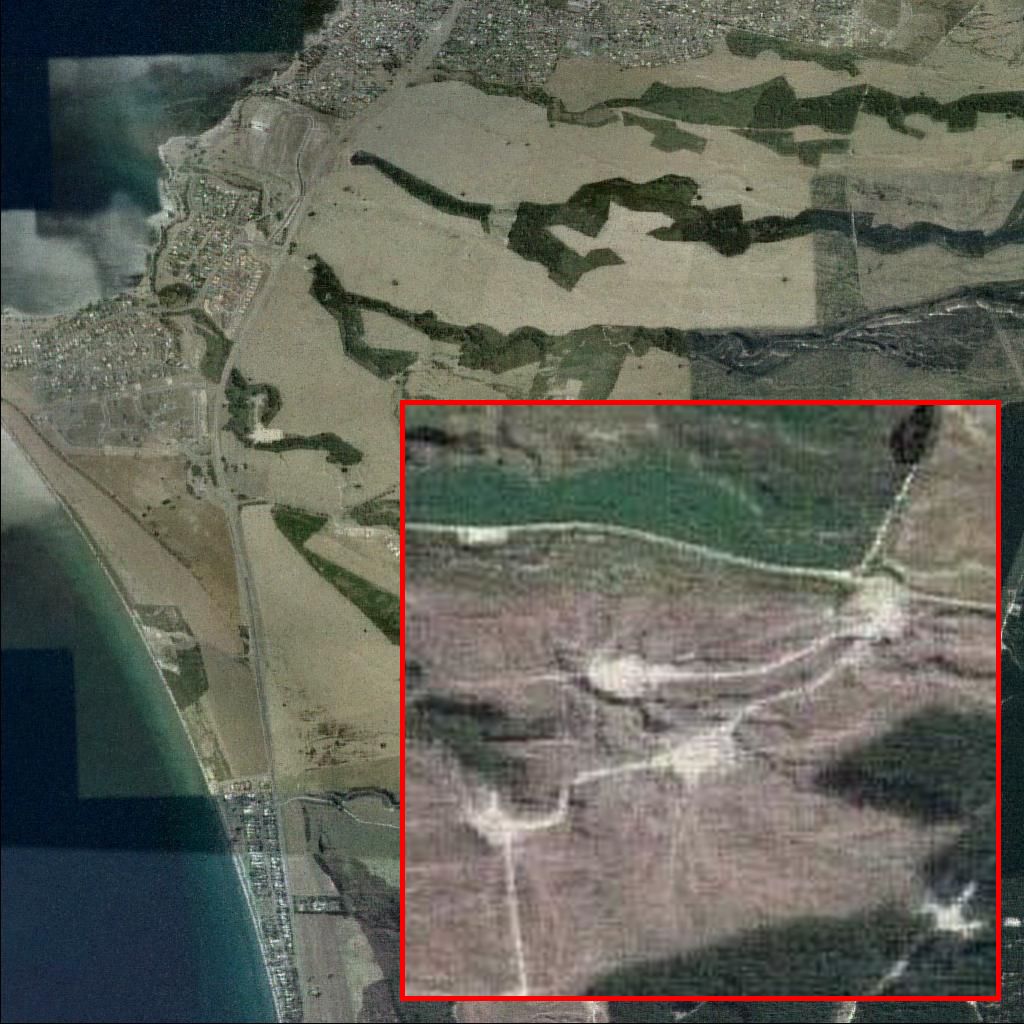}\vskip 2pt
			\includegraphics[width=1\textwidth]{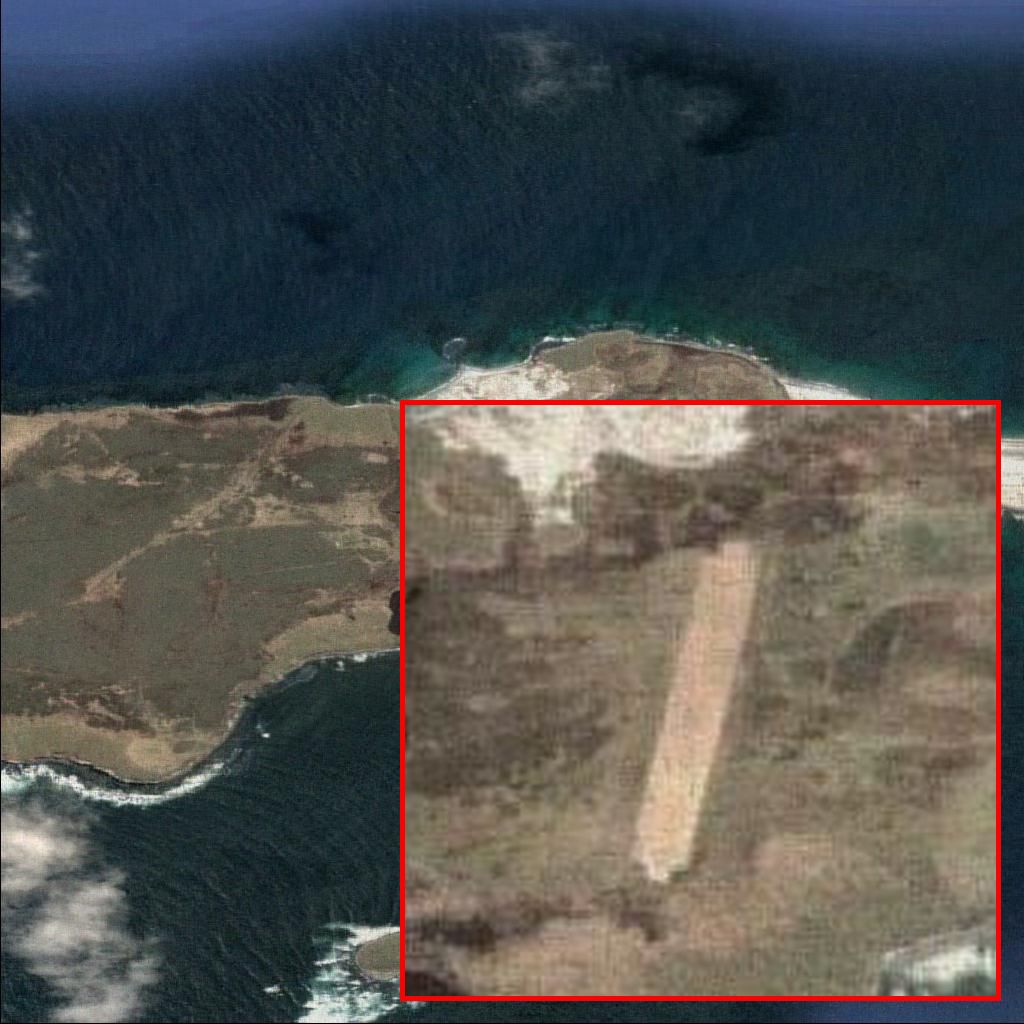}
		\end{minipage}
		\hspace{-12pt}
	}
\subfigure[TCDLR-RE]{
		\begin{minipage}[b]{0.12\linewidth}
			\includegraphics[width=1\textwidth]{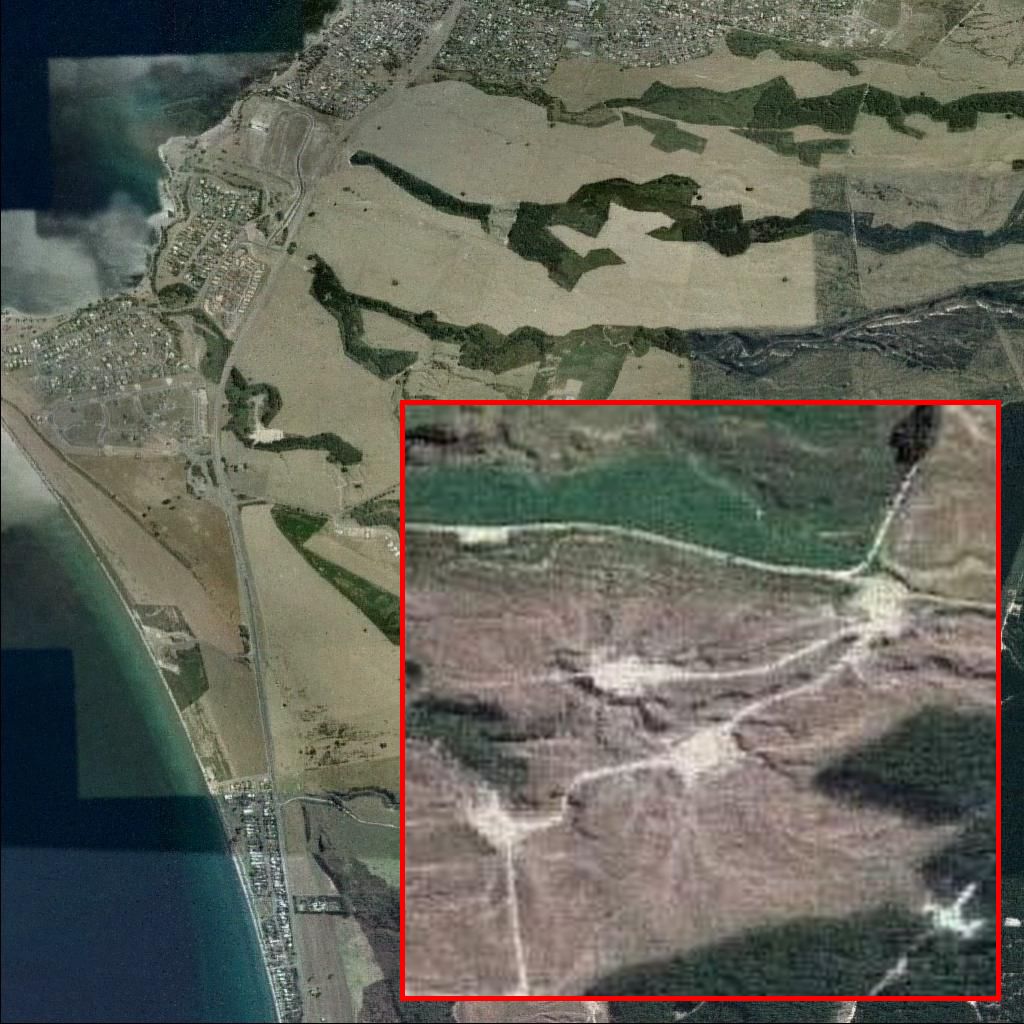}\vskip 2pt
			\includegraphics[width=1\textwidth]{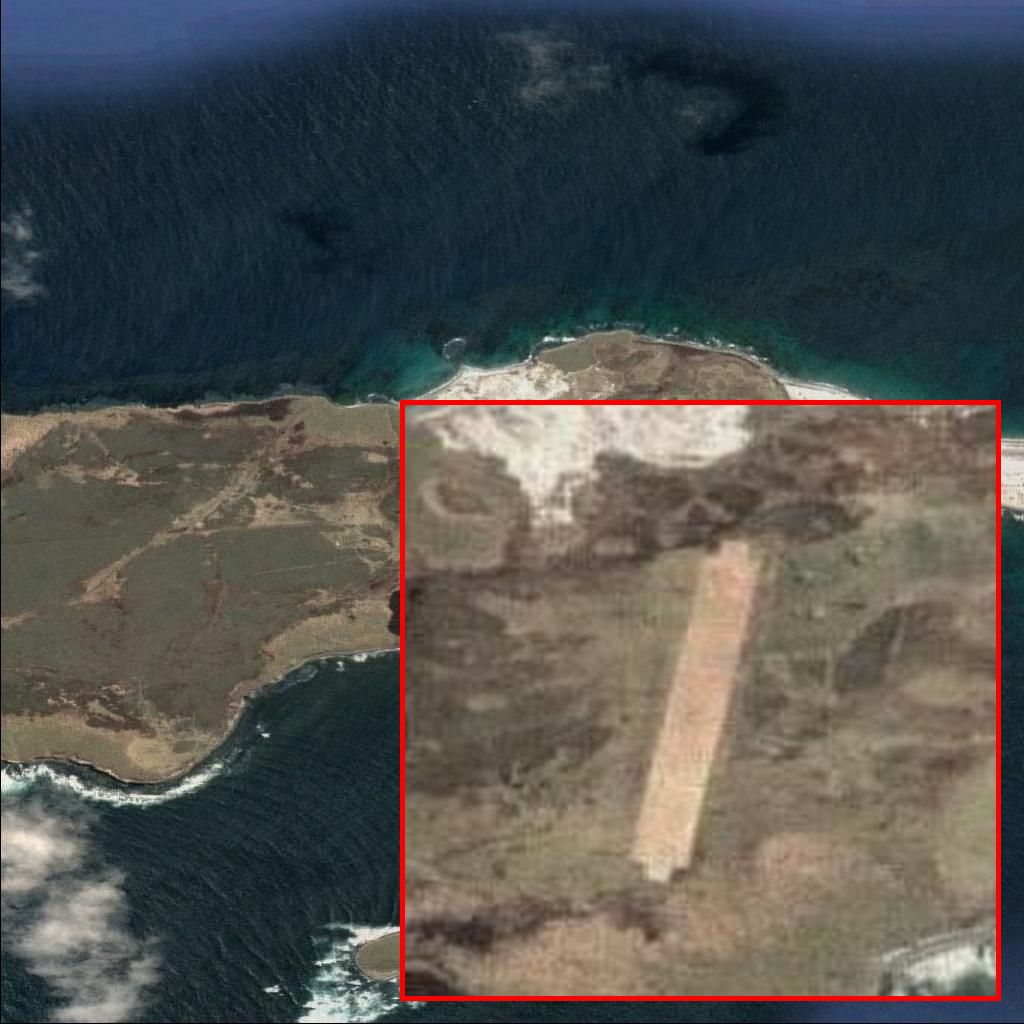}
		\end{minipage}
		\hspace{-12pt}
	}
 
	\caption{Completion  of visual results on the DOTA-v2.0 Dataset with a sampling rate of 30\%: (a) The original image and  the results obtained by different methods including (b) LRMC (c) SNN, (d) t-TNN, (e) TCTF, (f) PSTNN, (g) TC-RE, and (h) TCDLR-RE, respectively. }\label{DOTAimages}
	
\end{figure*}
\subsection{Real-world Applications}

In this subsection,  all seven tensor completion algorithms were tested in terms of image and video inpainting.   $cn_1n_2n_3$ elements were sampled uniformly from the data tensor $\mathcal{M}\in \mathbb{R}^{n_1\times n_2 \times n_3}$ to generate the observation matrices or the observation tensor   $\mathbf{P}_{\Omega}(\mathcal{M})$.

 Some implementation details are provided: (1) In these experiments, the sampling rate was set to $c=0.3$. The PSNR (the peak signal-to-noise ratio), MPSNR (mean PSNR), and running time were adopted to evaluate the effectiveness and efficiency of the method. The best results
for each case are shown in bold. Assuming that   $\mathcal{\hat X}$  is the recovered tensor from   $\mathbf{P}_{\Omega}(\mathcal{M})$, the PSNR value of  $\mathcal{\hat X}$ is formulated as 
$$
\operatorname{PSNR}=10 \log _{10}\left(\frac{n_{1} n_{2} n_{3}\|\mathcal{M}\|_{\infty}^{2}}{\|\hat{\mathcal{X}}-\mathcal{M}\|_{F}^{2}}\right),
$$  and the mean PSNR is defined as the average PSNR results of  $m$  selected images.   (2)  For the matrix recovery-based method (LRMC), matrix completion was performed on each frontal slice of the observed tensor, and the results were combined to obtain the recovered tensor.  
 (3) For TCTF, the initialized rank was set to $ k^{(0)}=[30,30,30] $ for the dataset with a small size (\ie, $n_1,~n_2<700$) as suggested in \cite{29}. Since $ k^{(0)}=[30,30,30] $ did not perform well for a large dataset, this paper empirically set the initialized rank as $ k^{(0)}=[70,70,70] $ in this case.
(4) The initialized rank in TCDLR-RE was set as $k^{(0)}=[0.05,0.05,\cdots,0.05]\times \min(n_1,n_2)$.

\subsubsection{Image Inpainting}
In this part, all algorithms were tested on two color image databases of different sizes: the Berkeley Segmentation Dataset \cite{32} and the DOTA Dataset\footnote{https://captain-whu.github.io/DOTA/index.html}\cite{Xia_2018_CVPR,Ding_2019_CVPR}.

\begin{table*}[]
\caption{Comparison of the MPSNR and average time (seconds) on the CAVE dataset with a sampling rate of  30\%.}
\centering
\scalebox{0.8}{
\resizebox{\textwidth}{!}{%
\begin{tabular}{cccccccc}
\hline
             & LRMC\cite{36}   & SNN\cite{12}     & t-TNN\cite{37}  & TCTF\cite{29}    & PSTNN\cite{jiang2020multi}   &TC-RE\cite{shi2021robust}   & TCDLR-RE \\ \hline
MPSNR        & 34.04  & 40.36  & 44.00   & 31.86     & 44.57  & 40.69   & \textbf{45.23}    \\
Average Time & 283.81 & 409.48 & 1256.16 & 102.28    & 345.62 & 1393.67 & \textbf{71.99}    \\ \hline
\end{tabular}%
}}\label{cave_table}
\end{table*}

 The MPSNR and time consumption of the seven competing inpainting algorithms on the Berkeley Segmentation Dataset are reported in Table \ref{smallimg}. Fig. \ref{segment_result} compares the PSNR values of different algorithms on 50 randomly selected images. From Table \ref{smallimg} and Fig. \ref{segment_result}, it can be seen that TCDLR-RE achieves the best performance in tensor recovery and took the least running time. Meanwhile, the visual quality of the seven algorithms is reported in Fig. \ref{comparedimages}, from which it can be seen that the visual quality of TCDLR-RE is more convincing. Specifically, the enlarged area in Fig. \ref{comparedimages} indicates that our TCDLR-RE well restores the eye of the eagle, the spots of the ladybug, and the scales and fins of the fish. Compared with LRMC and TCTF, there is less visible noise in the recovered images by TCDLR-RE. 

Fig. \ref{DOTAimages} and Table \ref{DOTA_table} present the experimental results of all algorithms on the DOTA-v2.0 dataset.  As shown in Table \ref{DOTA_table}, the PSNR and running time of the seven tensor completion algorithms on ten images indicate that: (1) the average PSNR value of TCDLR-RE is over 0.5 dB larger than those of the comparison methods; (2) TCDLR-RE runs much faster than other methods. Especially, the average running times of SNN, t-TNN, and PSTNN are 
about six times that of TCDLR-RE. Compared with TC-RE, TCDLR-RE even runs 40 times faster. In addition, the visual recovery results given in Fig.\ref{DOTAimages} indicate that the TCDLR-RE can retain more detail  within the image data than other methods. Besides, there is  more spot noise caused by the image inpainting algorithm on the recovered images by LRMC and TCTF.

\begin{figure*}
\centering
\includegraphics[width=6in]{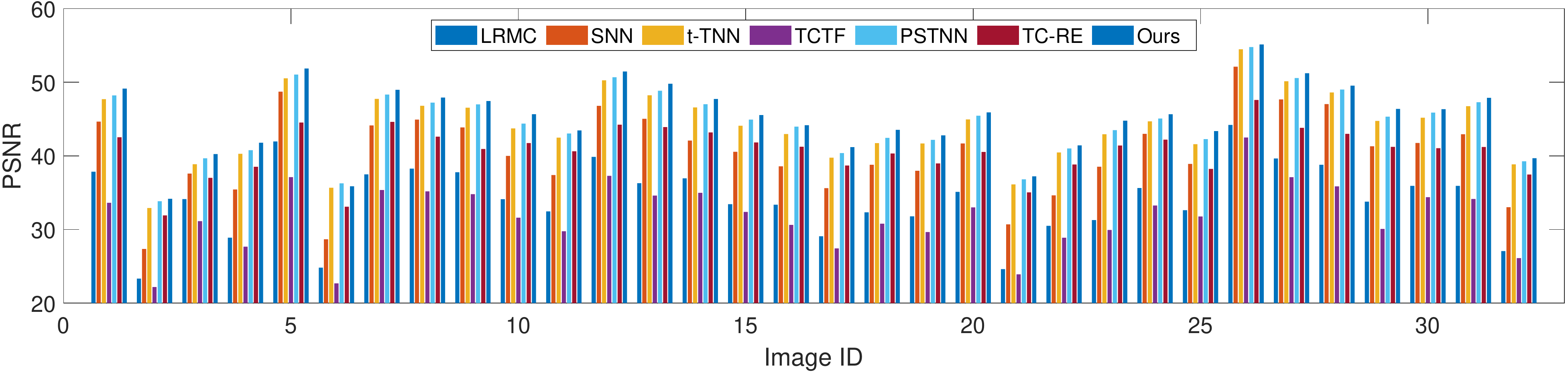}
\caption{Comparison of   the PSNR values  of all methods on 32 HSIs from the CAVE database.}\label{cave_result}
\end{figure*}
 
\begin{table*}[]
\centering
\caption{Comparison of the PSNR and running time (seconds) on first 8 HSIs from the BiHID with a sampling rate of 30\%.}
\resizebox{\textwidth}{!}{%
\begin{tabular}{c|cc|cc|cc|cc|cc|cc|cc}
\hline
\multirow{2}{*}{images} &
  \multicolumn{2}{c|}{LRMC\cite{36}} &
  \multicolumn{2}{c|}{SNN\cite{12}} &
  \multicolumn{2}{c|}{t-TNN\cite{37}} &
  \multicolumn{2}{c|}{TCTF\cite{29}} & 
  \multicolumn{2}{c|}{PSTNN\cite{jiang2020multi}} &
  \multicolumn{2}{c|}{TC-RE\cite{shi2021robust}} &
  \multicolumn{2}{c}{TCDLR-RE} \\ \cline{2-15} 
 &
  \multicolumn{1}{c|}{PSNR} &
  time(s) &
  \multicolumn{1}{c|}{PSNR} &
  time(s) &
  \multicolumn{1}{c|}{PSNR} &
  time(s) &
  \multicolumn{1}{c|}{PSNR} &
  time(s) & 
  \multicolumn{1}{c|}{psnr} &
  time &
  \multicolumn{1}{c|}{psnr} &
  time &
  \multicolumn{1}{c|}{PSNR} &
  time(s) \\ \hline
1 &
  \multicolumn{1}{c|}{39.13} &
  2774.19 &
  \multicolumn{1}{c|}{46.75} &
  3850.68 &
  \multicolumn{1}{c|}{49.98} &
  5370.50 &
  \multicolumn{1}{c|}{36.19} &
  695.07 & 
  \multicolumn{1}{c|}{50.44} &
  5083.09 &
  \multicolumn{1}{c|}{46.90} &
  37663.53 &
  \multicolumn{1}{c|}{\textbf{51.41}} &
  \textbf{589.36} \\
2 &
  \multicolumn{1}{c|}{36.52} &
  2822.36 &
  \multicolumn{1}{c|}{43.56} &
  3707.55 &
  \multicolumn{1}{c|}{47.45} &
  5420.99 &
  \multicolumn{1}{c|}{33.75} &
  766.11 & 
  \multicolumn{1}{c|}{47.92} &
  5080.61 &
  \multicolumn{1}{c|}{40.55} &
  38794.25 &
  \multicolumn{1}{c|}{\textbf{48.75}} &
  \textbf{585.96} \\
3 &
  \multicolumn{1}{c|}{34.62} &
  3211.77 &
  \multicolumn{1}{c|}{43.15} &
  3839.95 &
  \multicolumn{1}{c|}{47.52} &
  5937.54 &
  \multicolumn{1}{c|}{32.68} &
  690.66 & 
  \multicolumn{1}{c|}{48.03} &
  5170.47 &
  \multicolumn{1}{c|}{40.88} &
  40202.43 &
  \multicolumn{1}{c|}{\textbf{49.58}} &
  \textbf{583.63} \\
4 &
  \multicolumn{1}{c|}{40.22} &
  3061.57 &
  \multicolumn{1}{c|}{49.21} &
  4044.59 &
  \multicolumn{1}{c|}{51.84} &
  5478.93 &
  \multicolumn{1}{c|}{38.80} &
  697.81 & 
  \multicolumn{1}{c|}{52.36} &
  5120.01 &
  \multicolumn{1}{c|}{47.43} &
  35283.78 &
  \multicolumn{1}{c|}{\textbf{53.76}} &
  \textbf{585.62} \\
5 &
  \multicolumn{1}{c|}{35.13} &
  3110.88 &
  \multicolumn{1}{c|}{43.27} &
  3910.30 &
  \multicolumn{1}{c|}{47.90} &
  5867.13 &
  \multicolumn{1}{c|}{32.72} &
  700.20 & 
  \multicolumn{1}{c|}{48.35} &
  5128.86 &
  \multicolumn{1}{c|}{44.71} &
  36735.30 &
  \multicolumn{1}{c|}{\textbf{49.85}} &
  \textbf{582.80} \\
6 &
  \multicolumn{1}{c|}{42.32} &
  3032.16 &
  \multicolumn{1}{c|}{51.01} &
  4157.88 &
  \multicolumn{1}{c|}{53.18} &
  5585.10 &
  \multicolumn{1}{c|}{36.87} &
  701.17 & 
  \multicolumn{1}{c|}{53.72} &
  5127.92 &
  \multicolumn{1}{c|}{47.44} &
  32960.93 &
  \multicolumn{1}{c|}{\textbf{55.26}} &
  \textbf{581.58} \\
7 &
  \multicolumn{1}{c|}{36.63} &
  3098.21 &
  \multicolumn{1}{c|}{45.03} &
  4042.21 &
  \multicolumn{1}{c|}{49.66} &
  5829.48 &
  \multicolumn{1}{c|}{33.90} &
  701.07 & 
  \multicolumn{1}{c|}{50.17} &
  5167.41 &
  \multicolumn{1}{c|}{46.49} &
  35307.90 &
  \multicolumn{1}{c|}{\textbf{51.93}} &
  \textbf{572.78} \\
8 &
  \multicolumn{1}{c|}{38.42} &
  2736.59 &
  \multicolumn{1}{c|}{46.02} &
  3807.17 &
  \multicolumn{1}{c|}{47.82} &
  5087.35 &
  \multicolumn{1}{c|}{34.76} &
  681.42 & 
  \multicolumn{1}{c|}{48.28} &
  5176.80 &
  \multicolumn{1}{c|}{44.33} &
  49373.34 &
  \multicolumn{1}{c|}{\textbf{49.29}} &
  \textbf{585.03} \\ \hline
Average &
  \multicolumn{1}{c|}{37.87} &
  2980.97 &
  \multicolumn{1}{c|}{46.00} &
  3920.04 &
  \multicolumn{1}{c|}{49.42} &
  5572.13 &
  \multicolumn{1}{c|}{34.96} &
  704.19 & 
  \multicolumn{1}{c|}{49.91} &
  5131.90 &
  \multicolumn{1}{c|}{44.84} &
  38290.18 &
  \multicolumn{1}{c|}{\textbf{51.23}} &
  \textbf{583.35} \\ \hline
\end{tabular}}\label{bgu_table}
\end{table*}

\subsubsection{HSI Inpainting}
 Here, the performance of all methods was evaluated on two different hyperspectral image (HSIs) databases:  the CAVE database\footnote{http://www.cs.columbia.edu/CAVE/databases/multispectral/} \cite{Xie_2016_CVPR} and the BGU iCVL Hyperspectral Image Dataset (BiHID) \footnote{http://icvl.cs.bgu.ac.il/hyperspectral/} \cite{35}.

The experimental results on the CAVE database and the BiHID are presented in Fig. \ref{cave_result}, Table \ref{cave_table} and Table \ref{bgu_table}, respectively. From these results, the following observations are obtained.  First, TCDLR-RE achieves the best PSNR and MPRNR on both datasets. Second, on both datasets, the proposed method (TCDLR-RE) runs much faster than the comparison methods. Especially,  
for the BiHID, the average running time of SNN, t-TNN, and PSTNN is about six times that of TCDLR-RE, and TCDLR-RE runs even 60 times faster than TC-RE. 
All these results demonstrate the effectiveness and high efficiency of TCDLR-RE for HSIs.


\subsubsection{Video Inpainting}
In this part, the seven methods were tested on the first five videos in the GOT-10k video database\footnote{http://got-10k.aitestunion.com/ }\cite{39}, including  ``Dolphin''($1920 \times 1080 \times 100$), ``City'' ($ 1920\times1080\times80 $), `Dock'($ 1920\times1080\times80 $), ``Ship'' ($ 1280\times720\times71 $), ``Handrail'' ($ 1920\times1080\times68$), ``Penguin'' ($ 1920\times1080\times100 $), ``Leg'' ($1280\times 720\times 100$), ``Chicken'' ($1920\times1080\times100$), ``Bird'' ($1920\times1080\times97$), and ``Swan'' ($1280\times 720\times 100$).  The first 30 frames of each video sequence were taken and 
converted to the gray format. 
In this way, a tensor $\mathcal{M} \in \mathbb{R}^{n_1\times n_2 \times n_3}$ was constructed for a gray video sequence with a frame size of $n_1\times n_2$, where $n_3$ is the number of frames in the video sequence.   

\begin{figure*}
	\subfigure[Original]{
		\begin{minipage}[b]{0.12\linewidth}
			\includegraphics[width=1\textwidth]{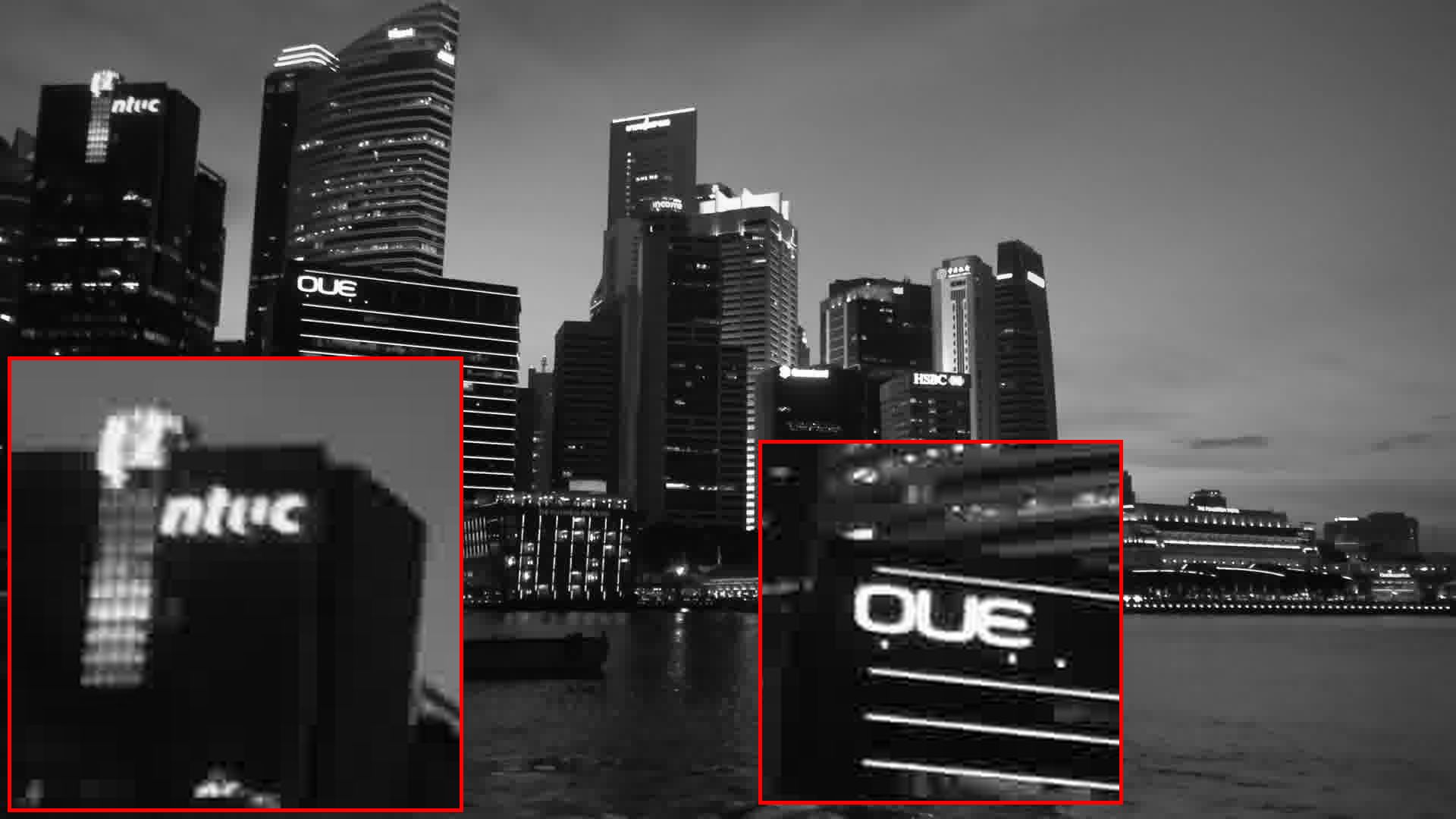}\vskip 2pt
			\includegraphics[width=1\textwidth]{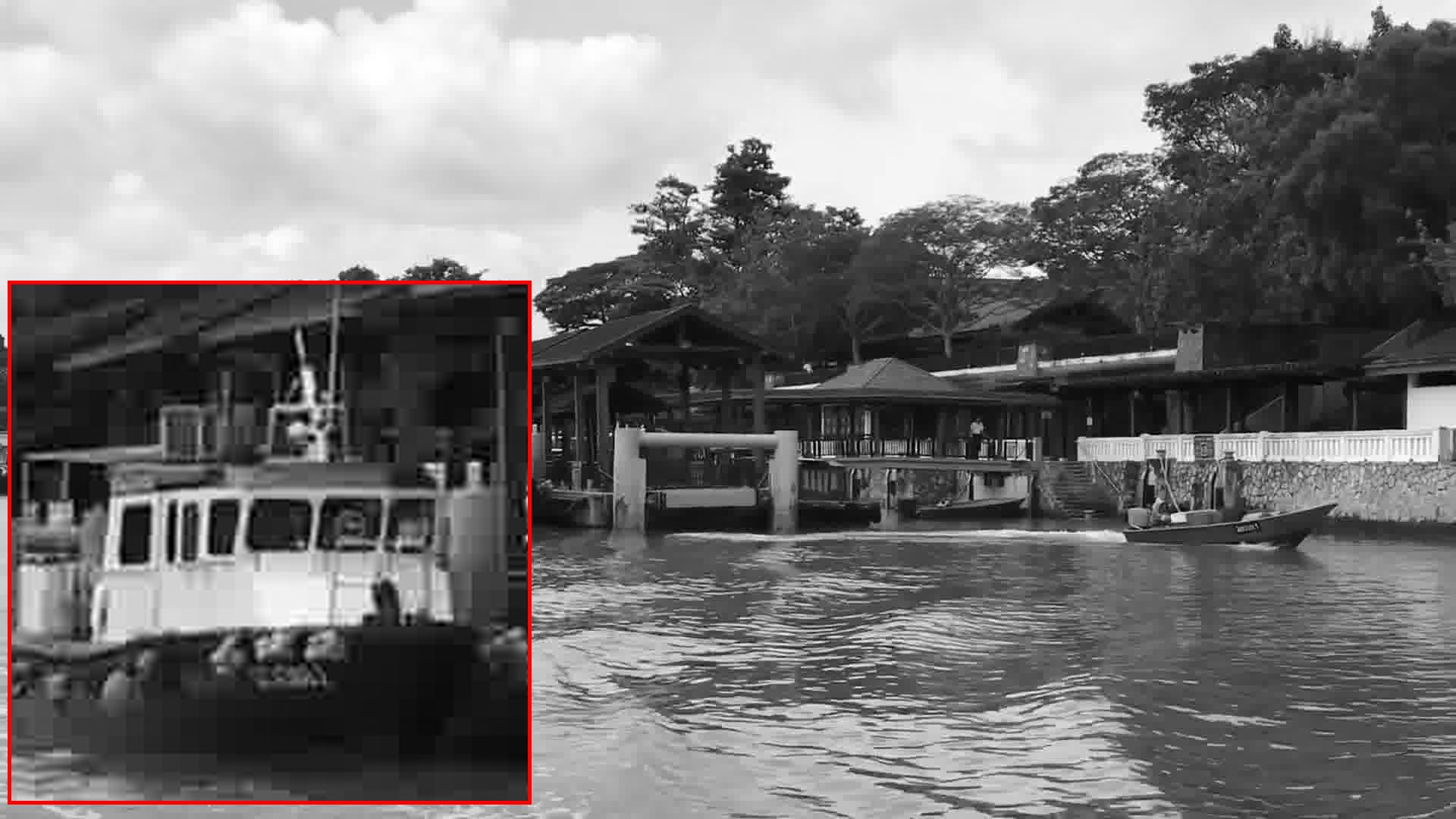}\vskip 2pt
			\includegraphics[width=1\textwidth]{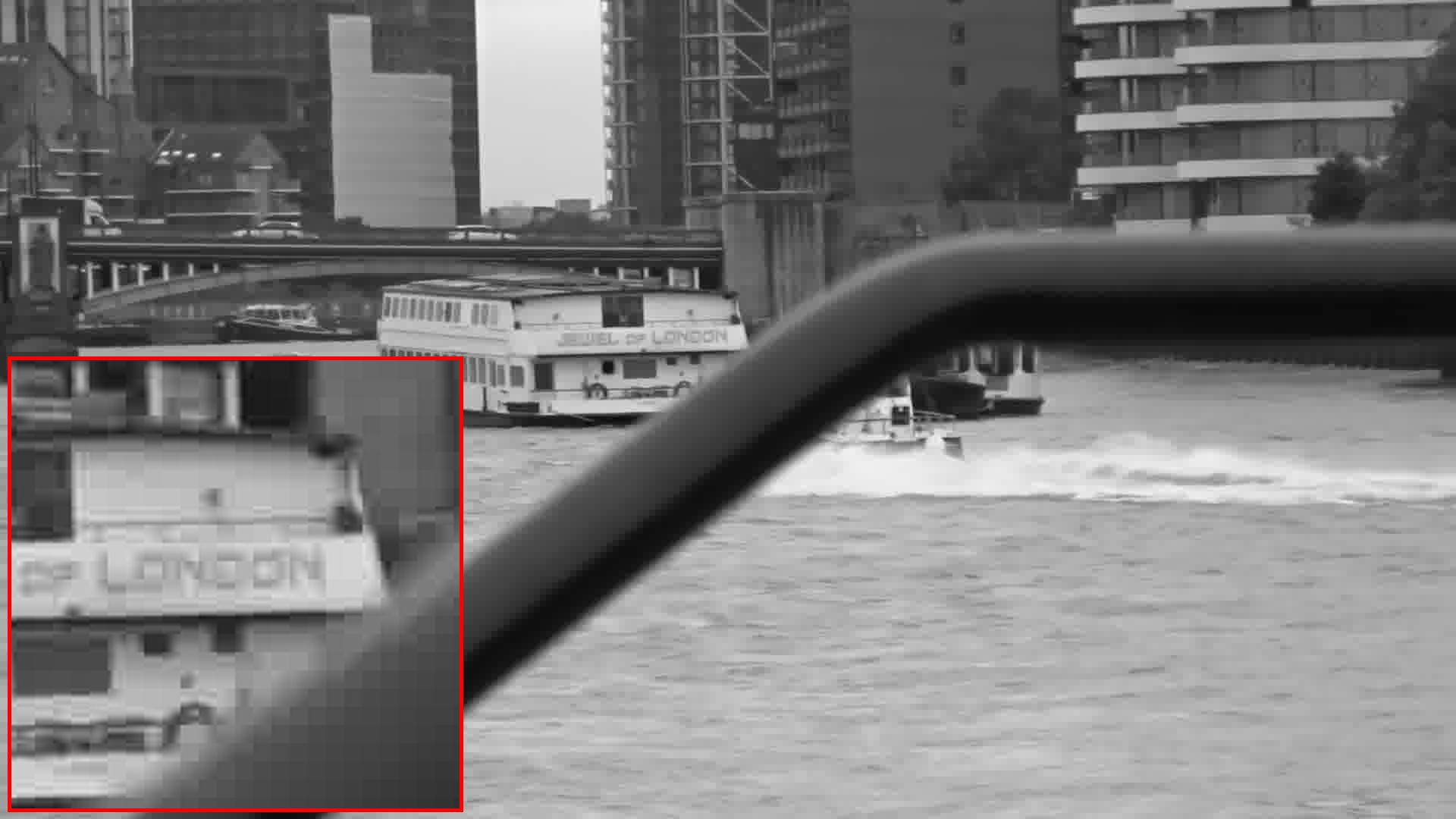}
		\end{minipage}
	\hspace{-12pt}
	}
   \subfigure[LRMC \cite{36}]{
		\begin{minipage}[b]{0.12\linewidth}
			\includegraphics[width=1\textwidth]{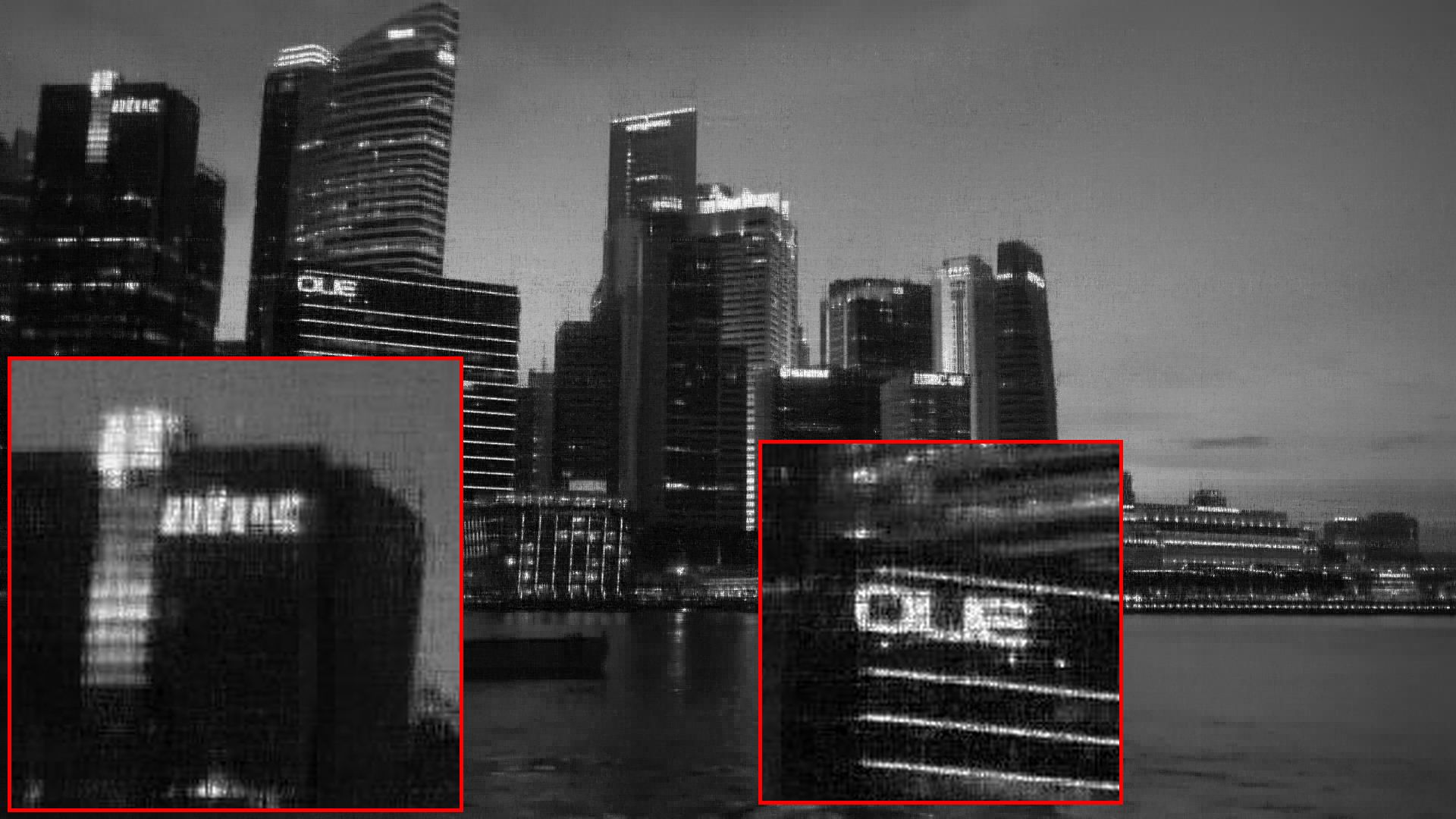}\vskip 2pt
			\includegraphics[width=1\textwidth]{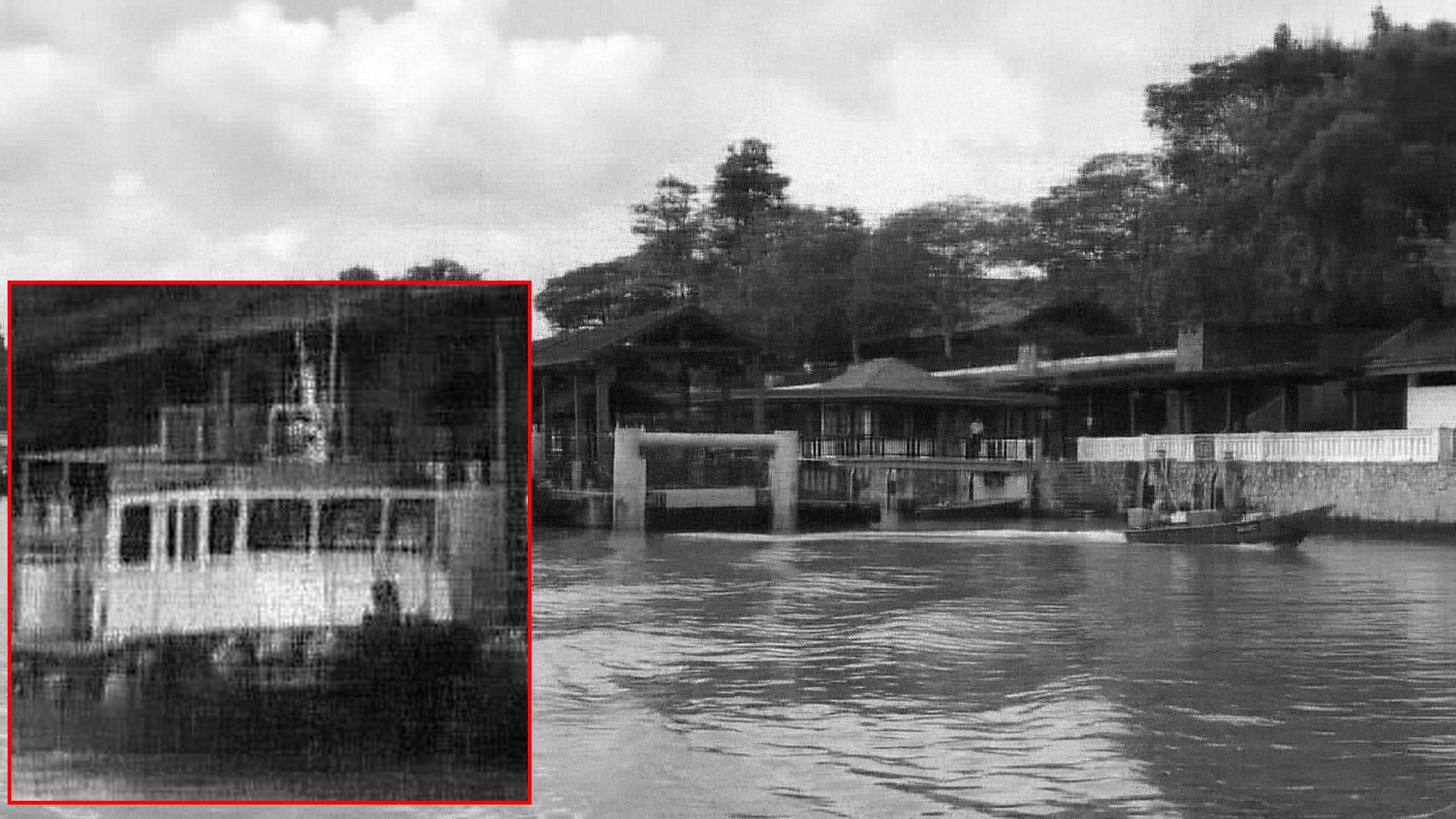}\vskip 2pt
			\includegraphics[width=1\textwidth]{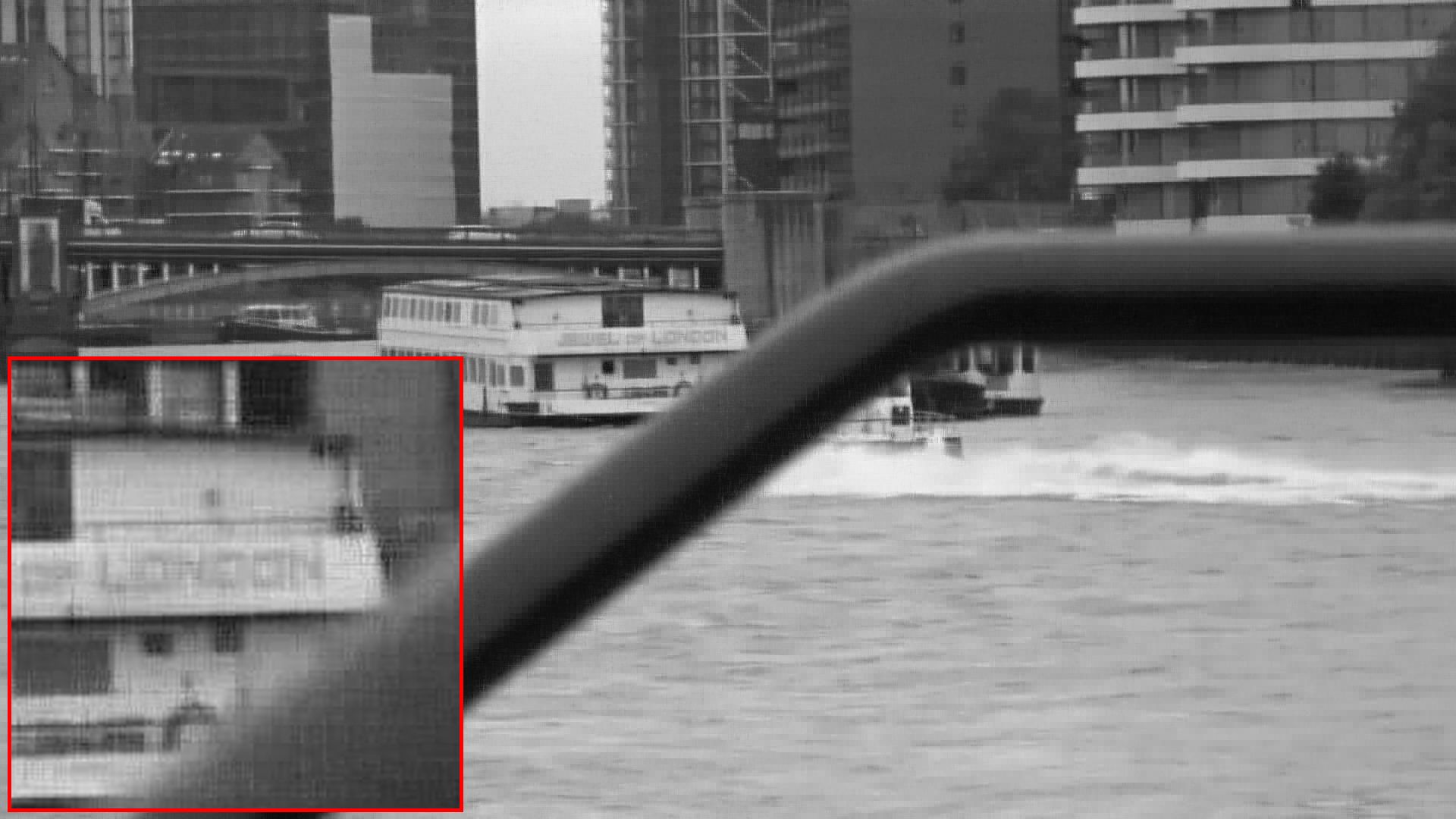}
		\end{minipage}
	\hspace{-12pt}
	}
        \subfigure[SNN \cite{12}]{
	\begin{minipage}[b]{0.12\linewidth}
		\includegraphics[width=1\textwidth]{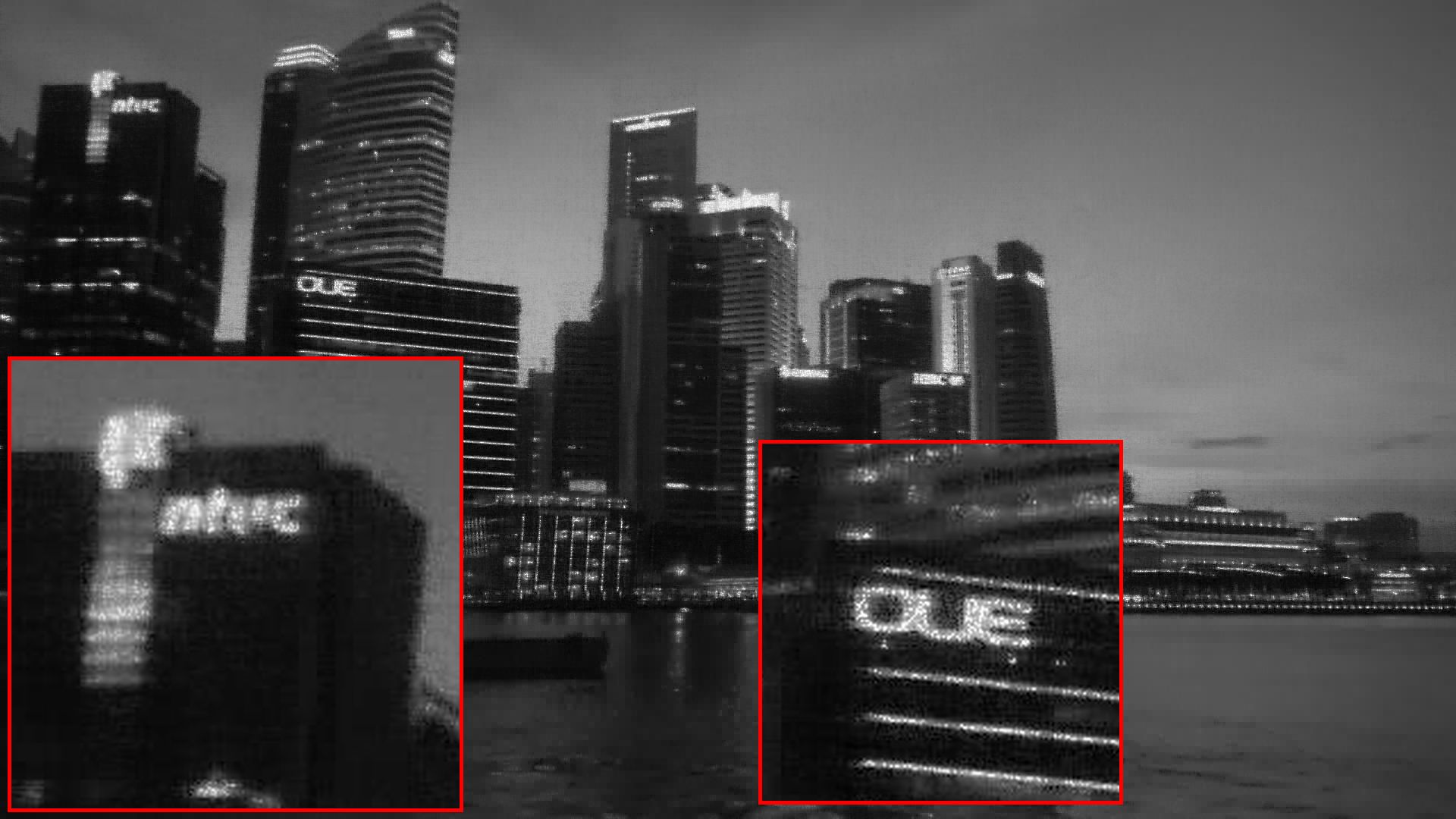}\vskip 2pt
		\includegraphics[width=1\textwidth]{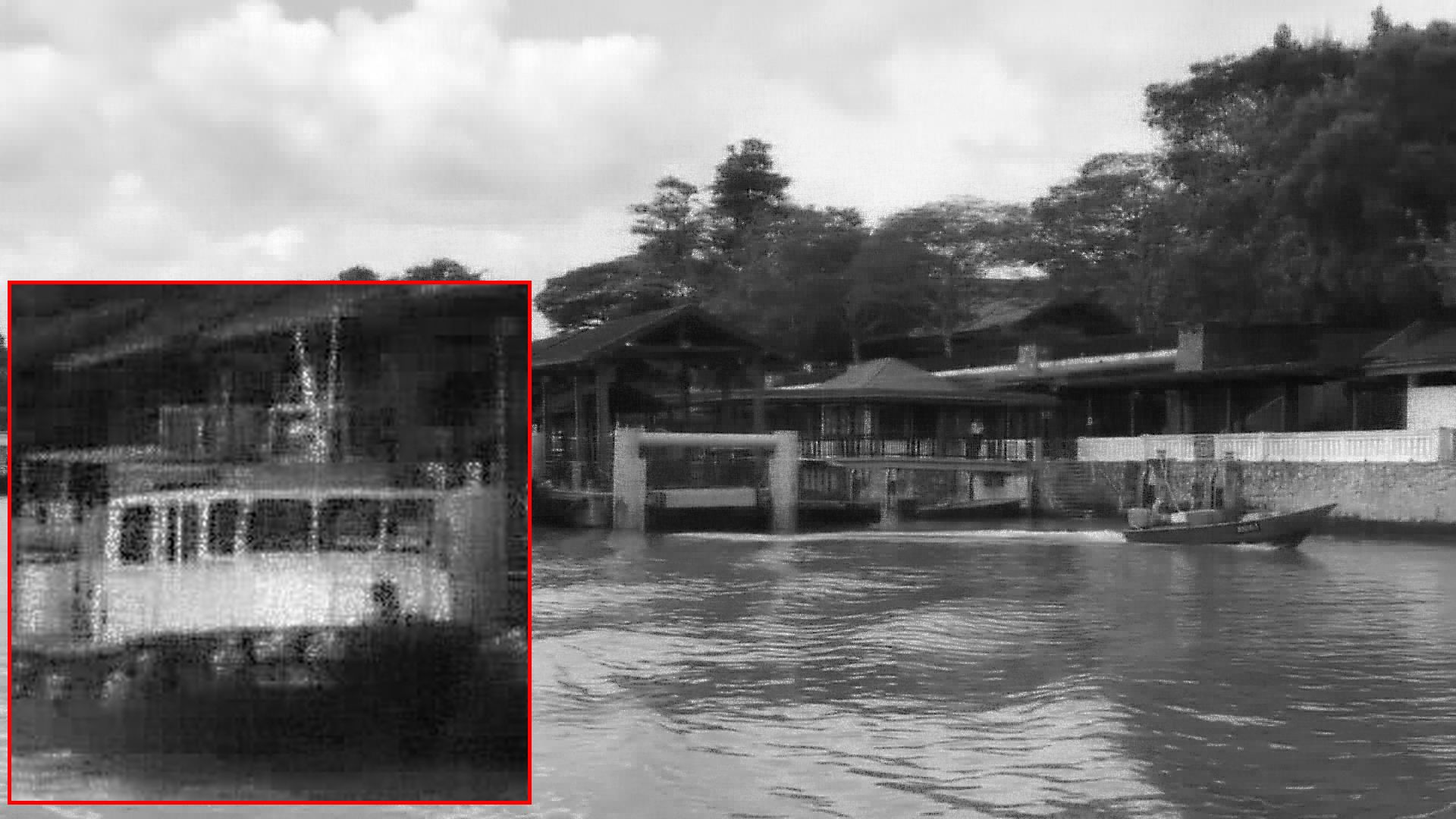}\vskip 2pt
		\includegraphics[width=1\textwidth]{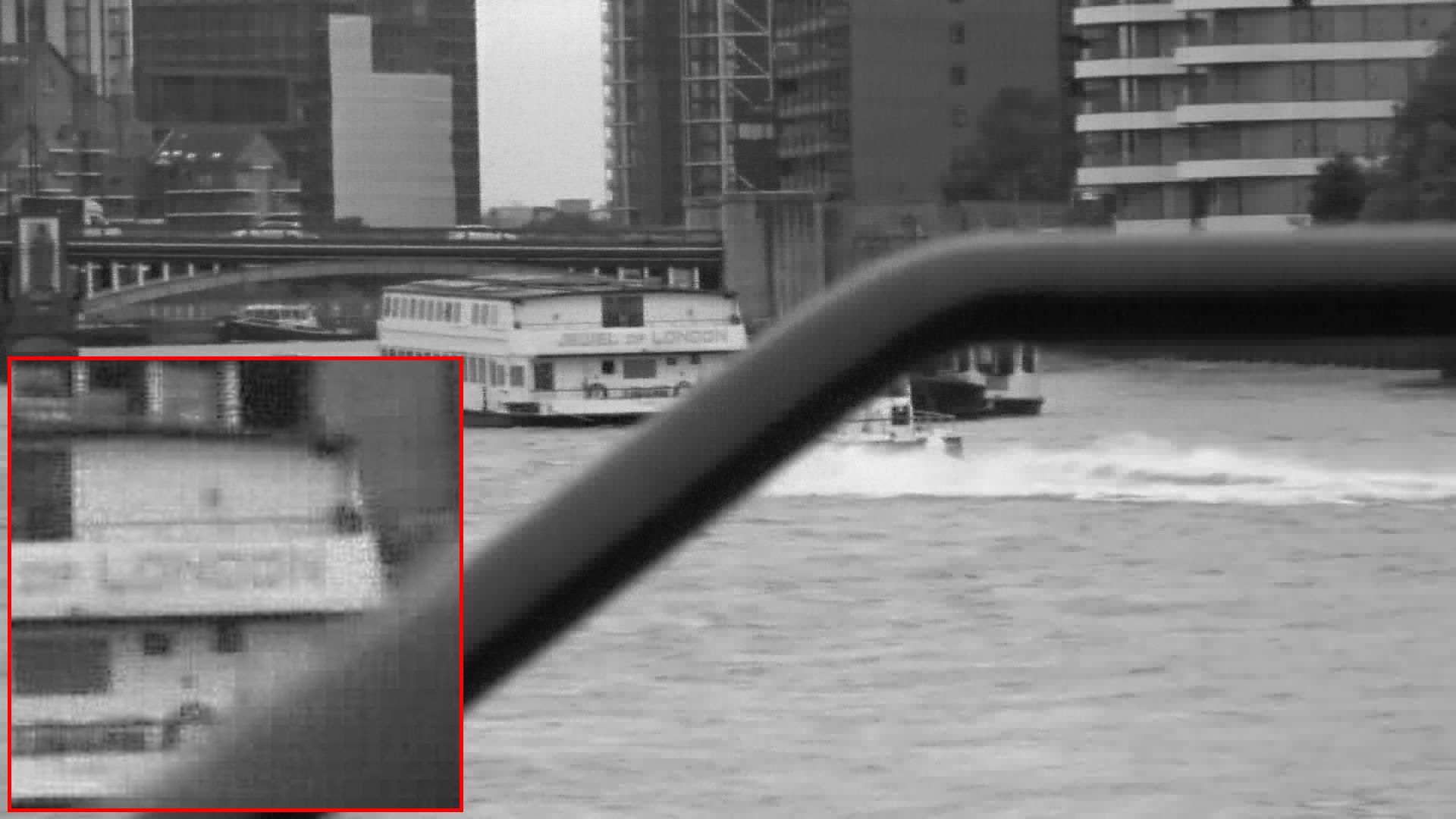}
	\end{minipage}
	\hspace{-12pt}
}
        \subfigure[t-TNN \cite{37}]{
	\begin{minipage}[b]{0.12\linewidth}
		\includegraphics[width=1\textwidth]{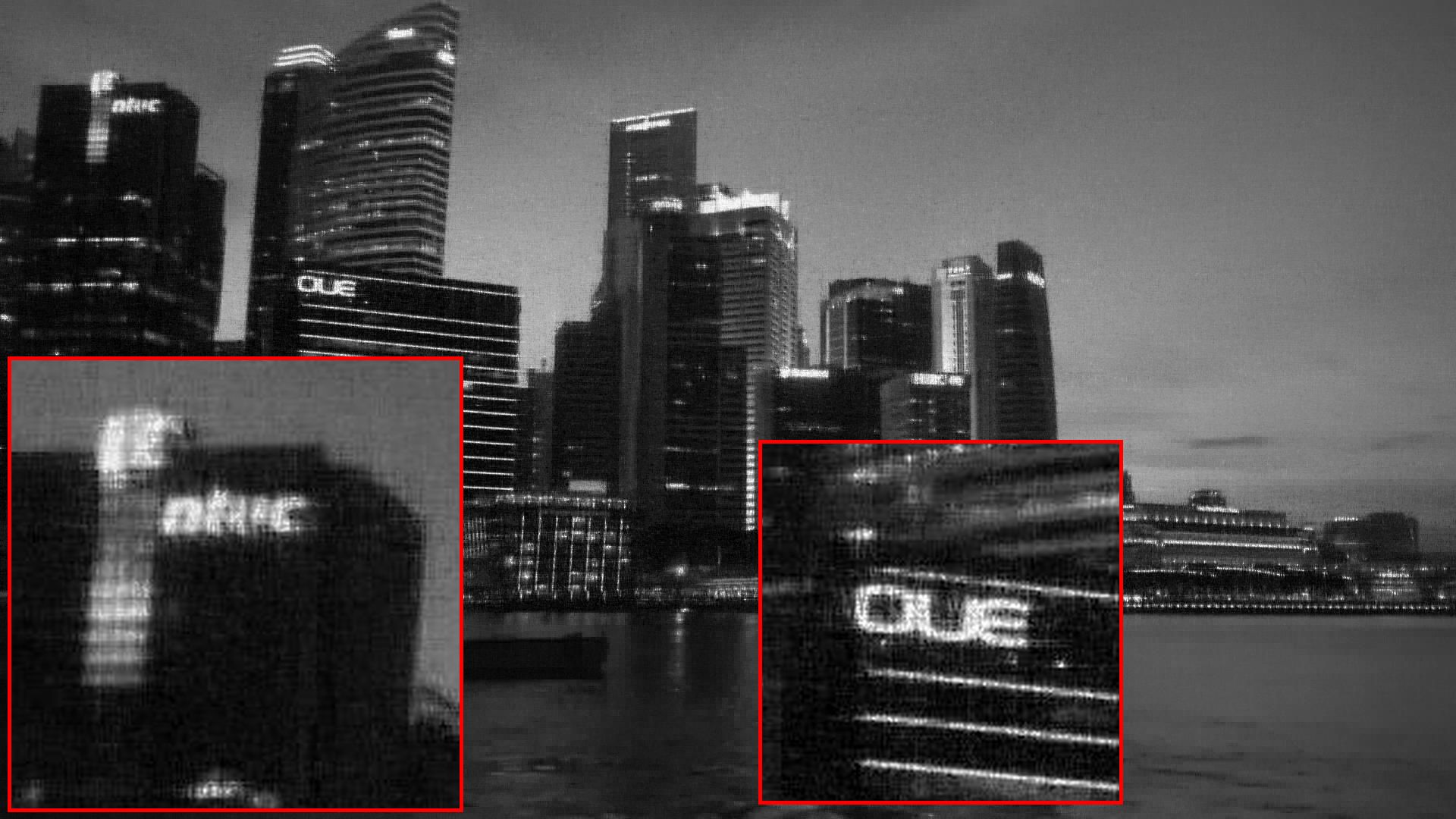}\vskip 2pt
		\includegraphics[width=1\textwidth]{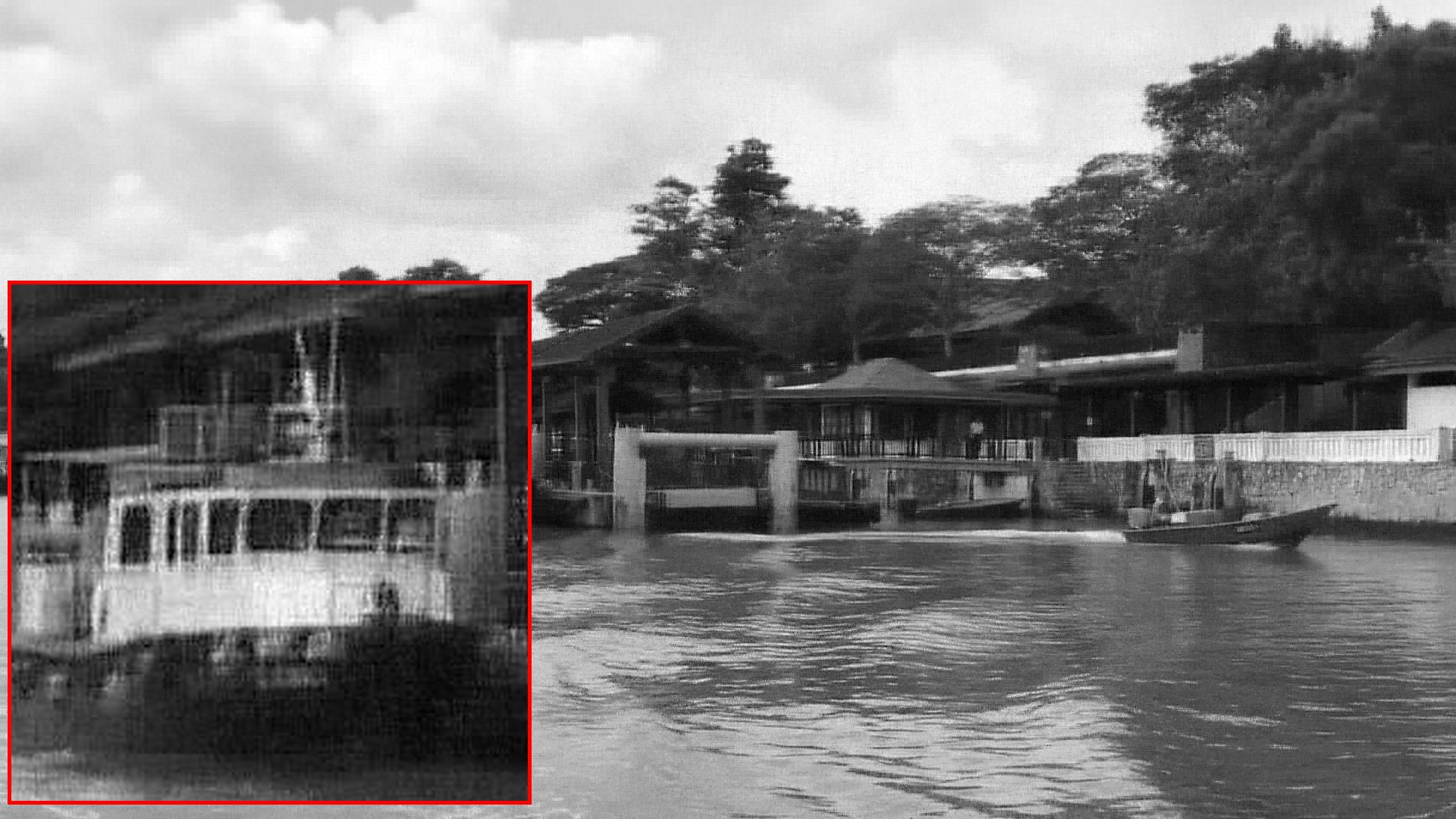}\vskip 2pt
		\includegraphics[width=1\textwidth]{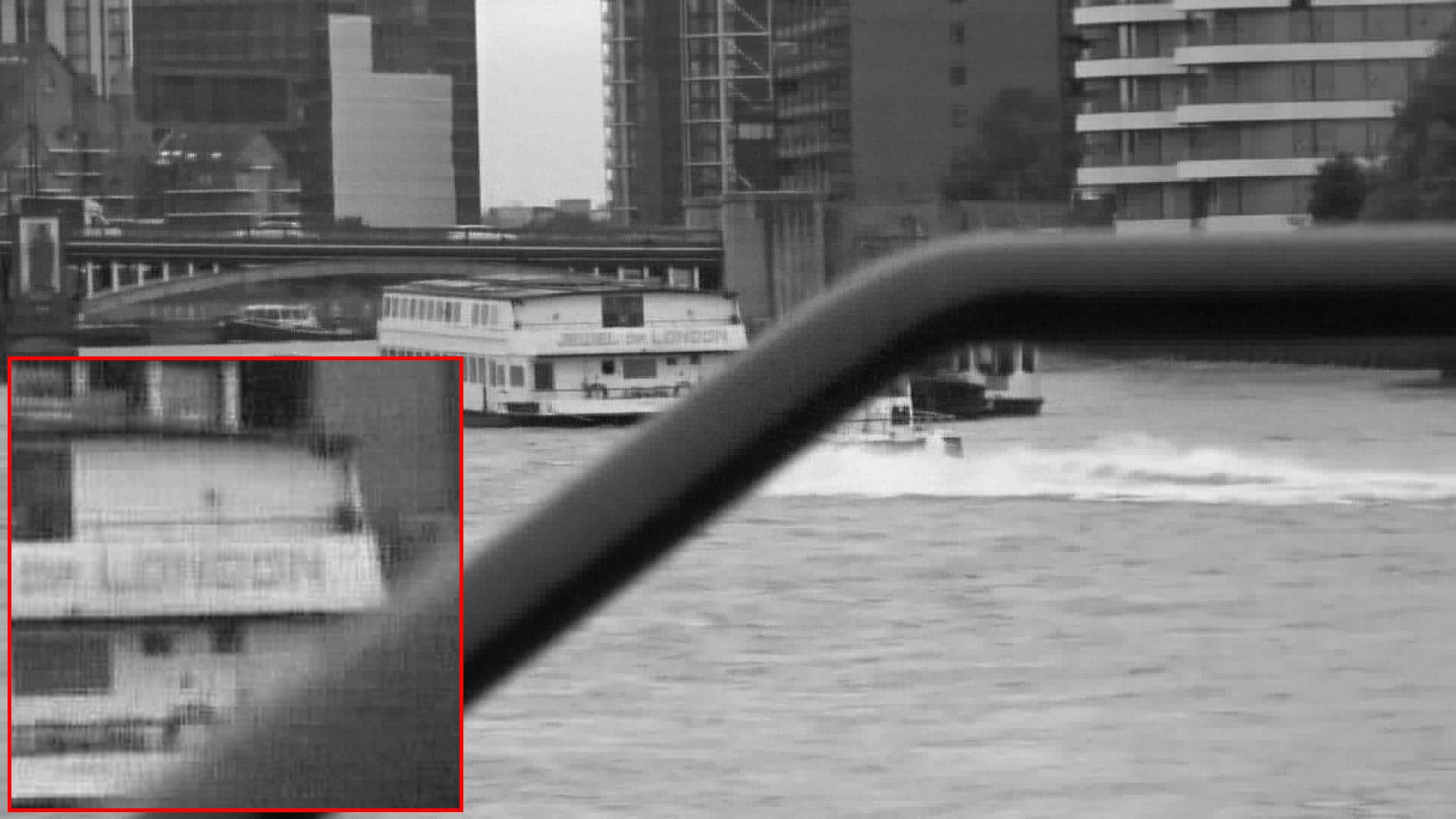}
	\end{minipage}
	\hspace{-12pt}
}
        \subfigure[TCTF \cite{29}]{
	\begin{minipage}[b]{0.12\linewidth}
		\includegraphics[width=1\textwidth]{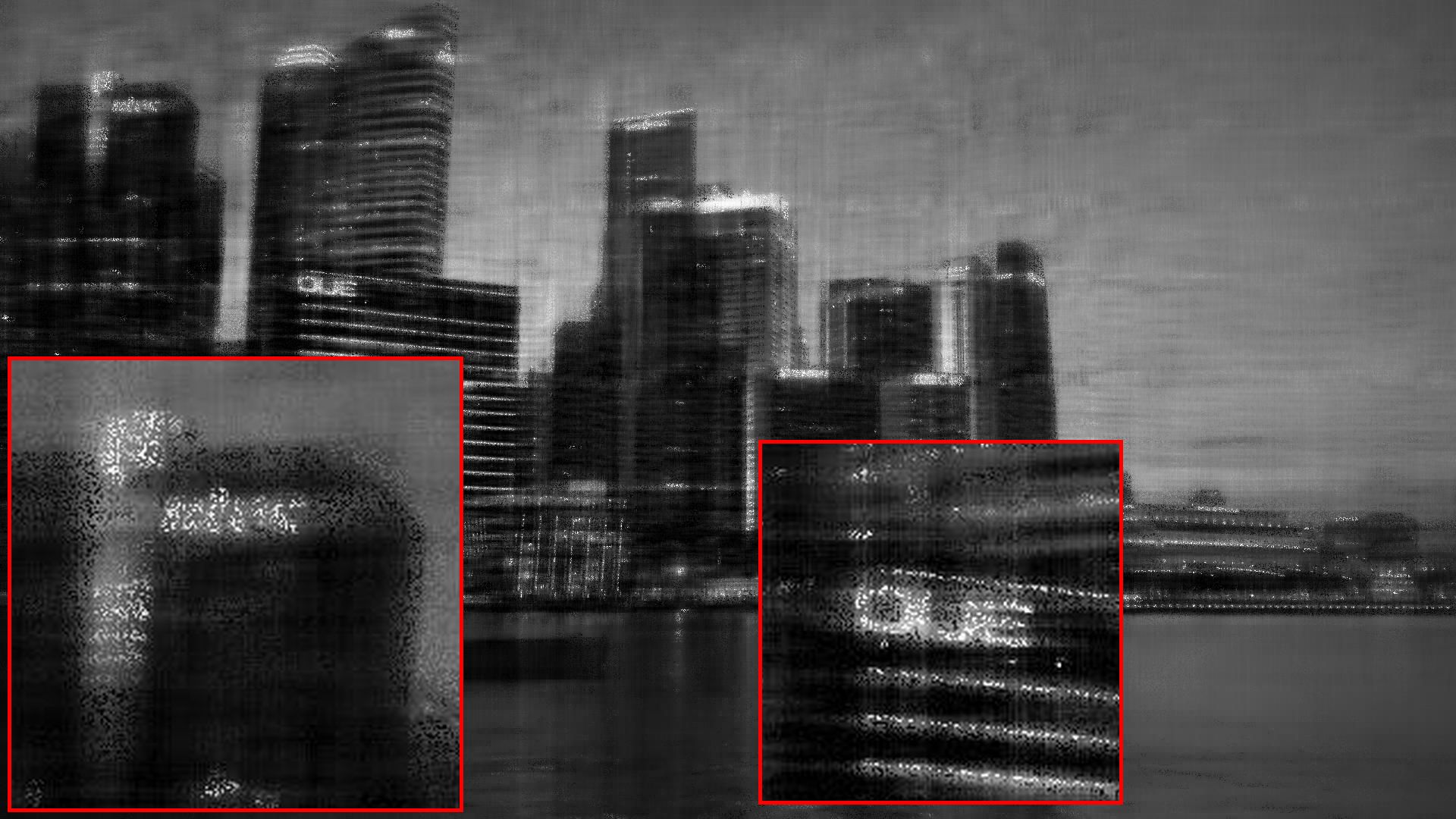}\vskip 2pt
		\includegraphics[width=1\textwidth]{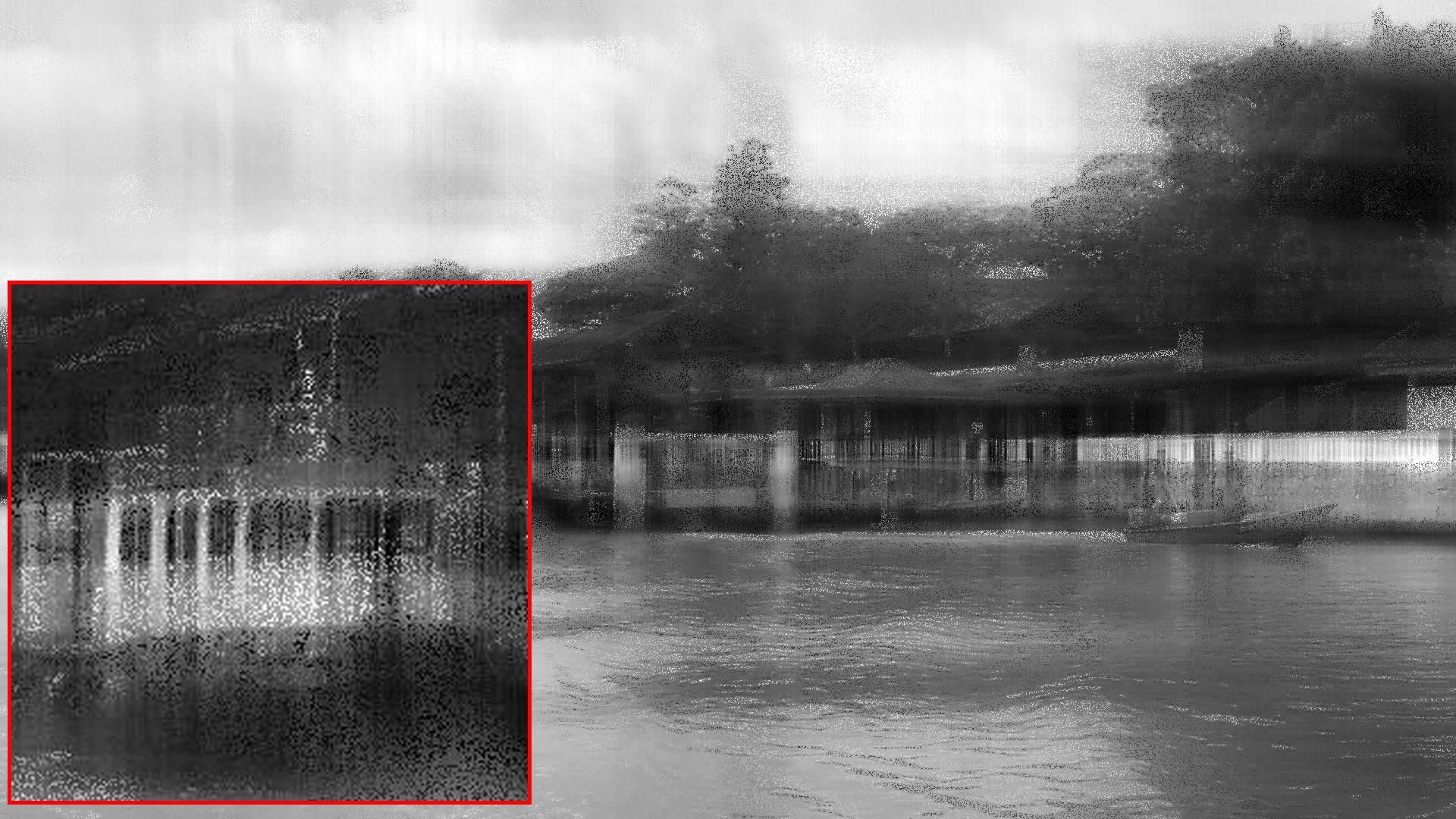}\vskip 2pt
		\includegraphics[width=1\textwidth]{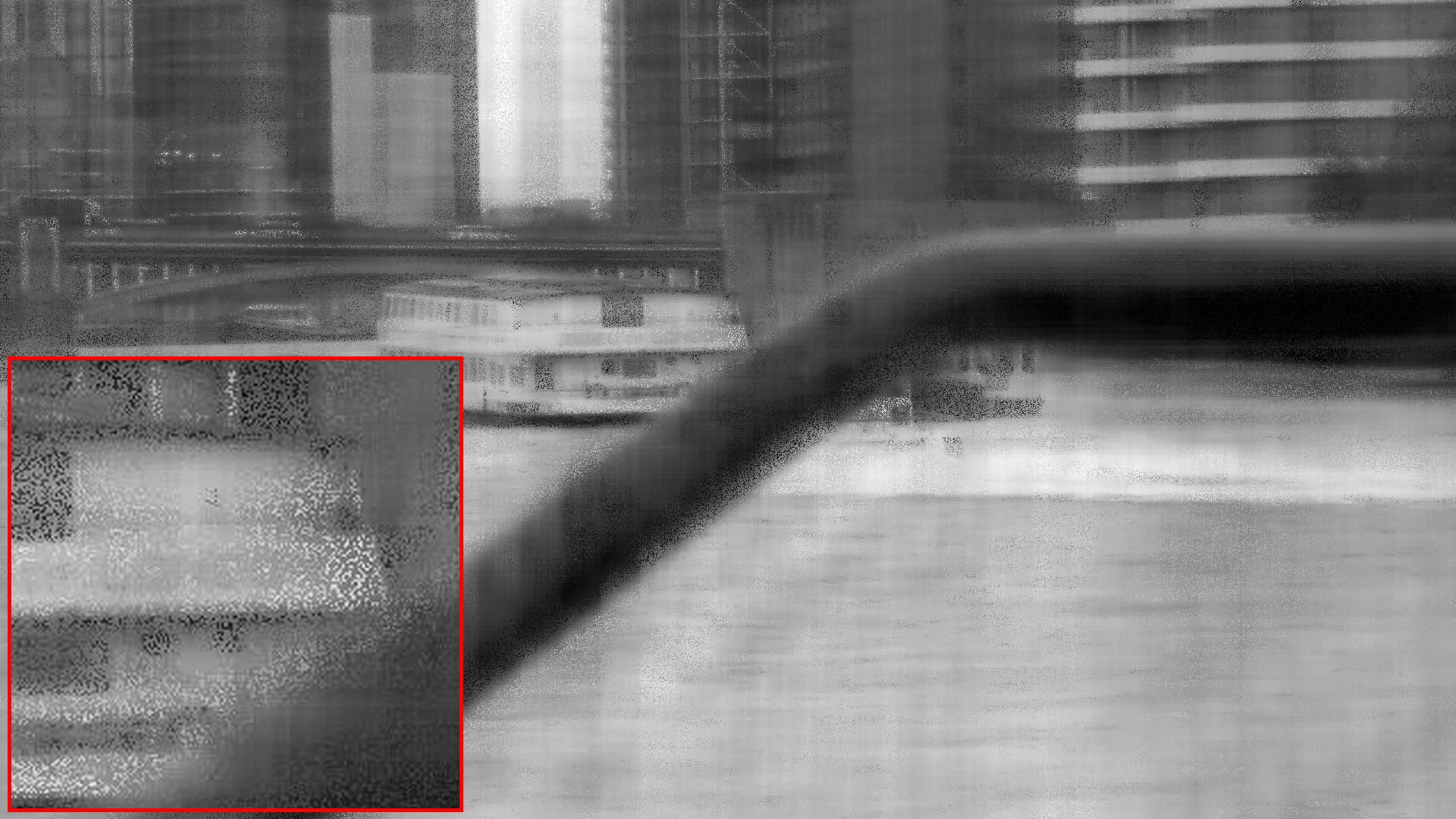}
	\end{minipage}
	\hspace{-12pt}
}
	\subfigure[TC-RE\cite{shi2021robust}]{
		\begin{minipage}[b]{0.12\linewidth}
			\includegraphics[width=1\textwidth]{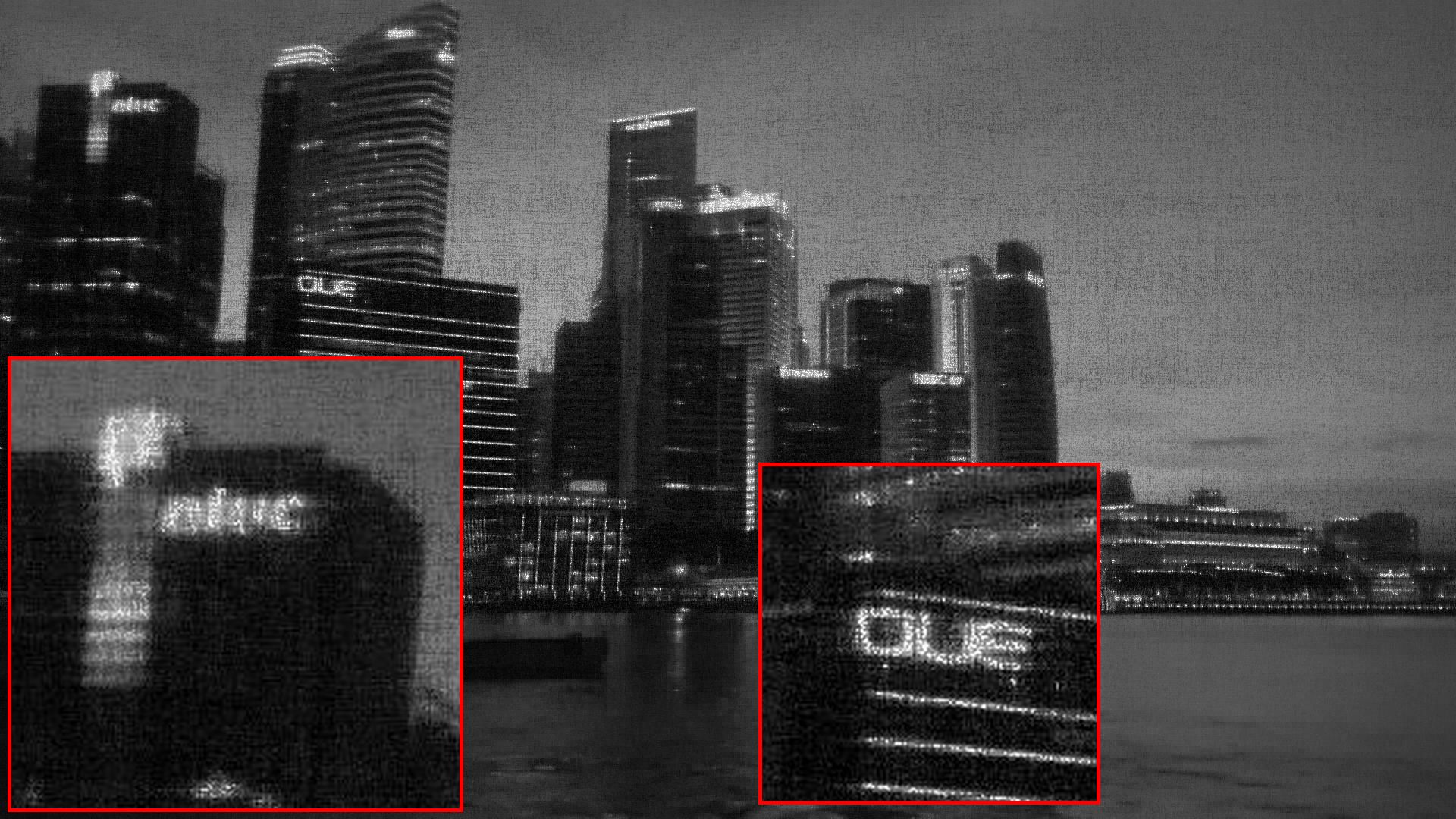}\vskip 2pt
			\includegraphics[width=1\textwidth]{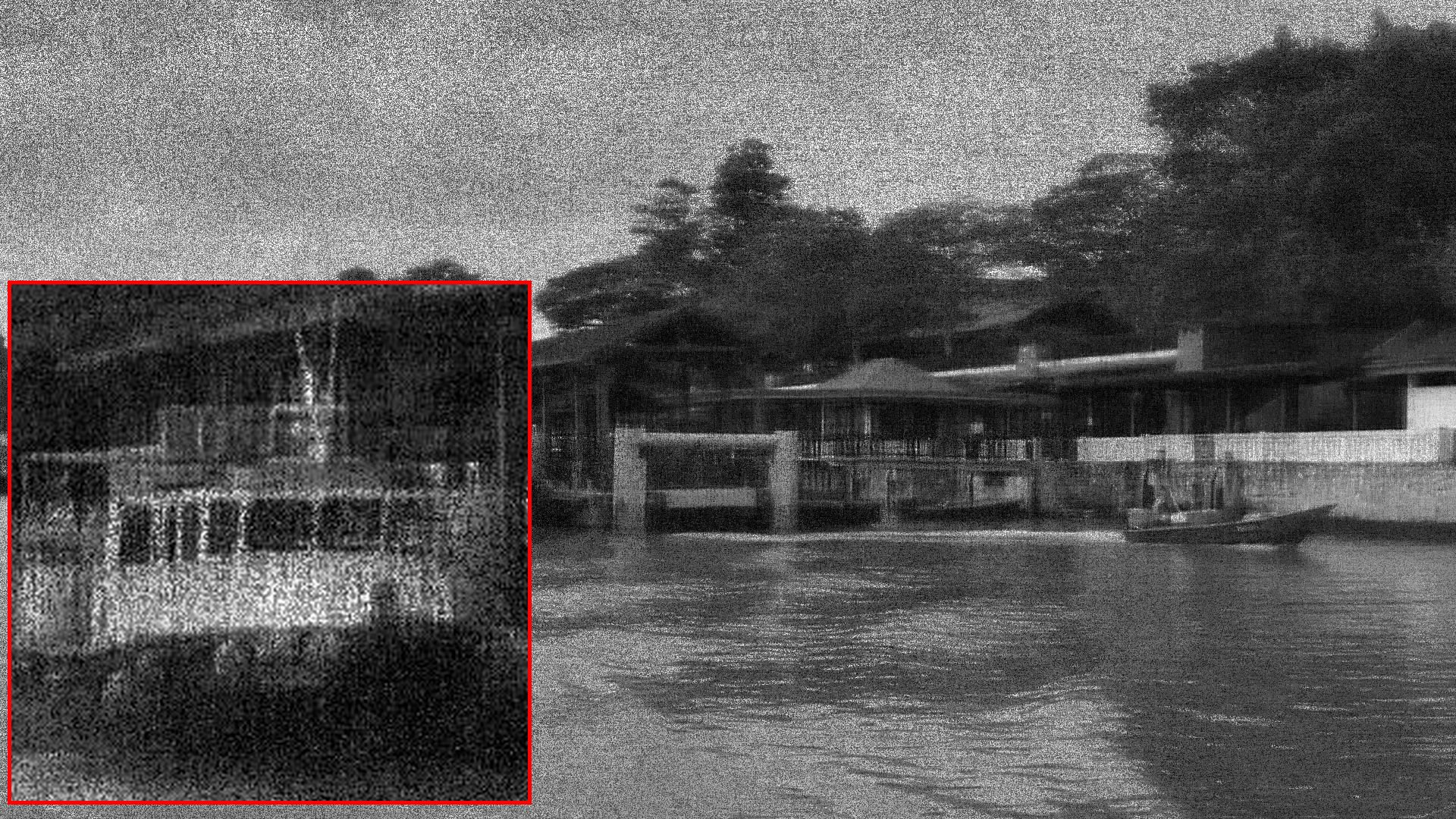}\vskip 2pt
			\includegraphics[width=1\textwidth]{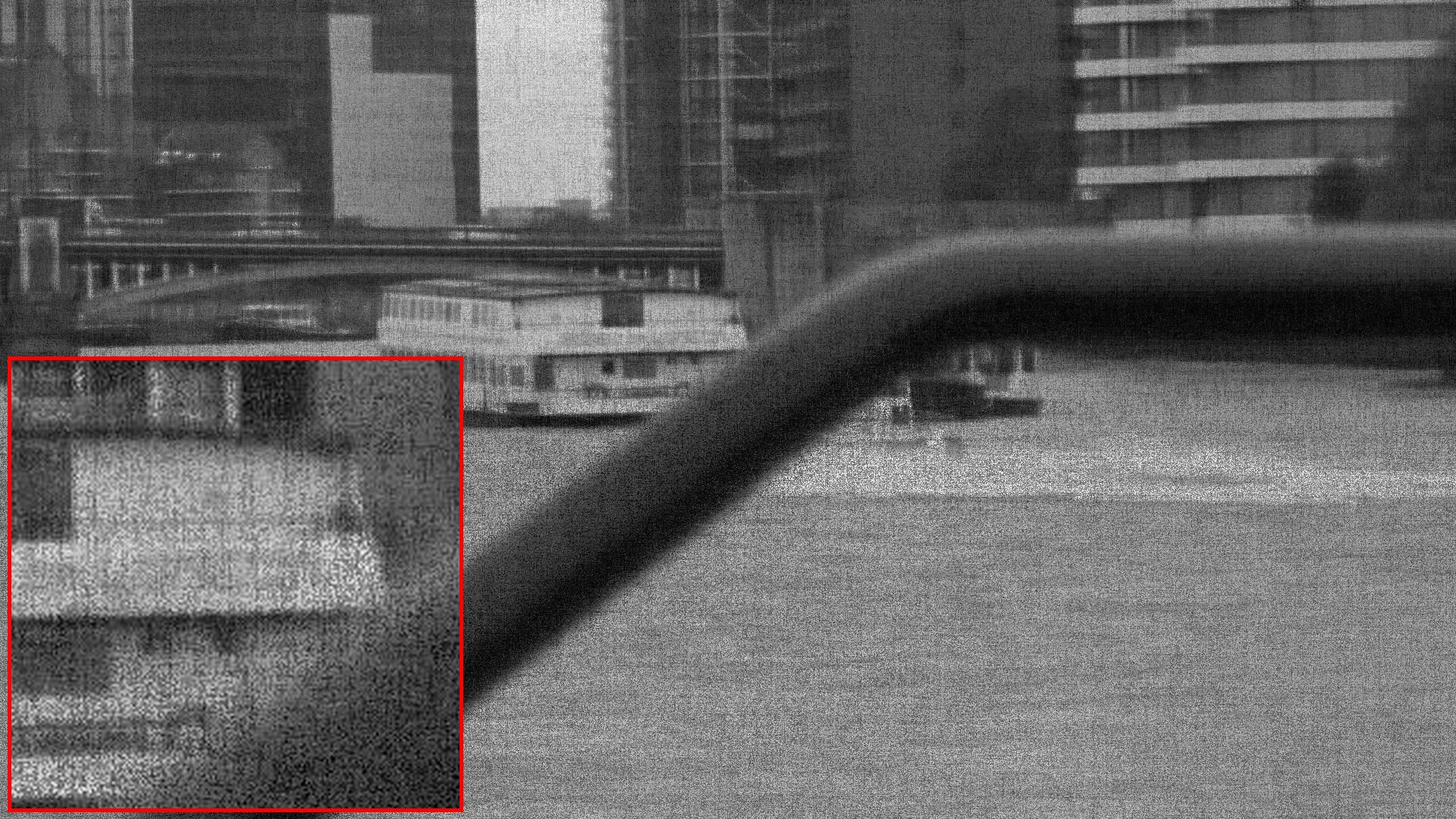}
		\end{minipage}
	\hspace{-12pt}
	} 
\subfigure[PSTNN\cite{jiang2020multi}]{
		\begin{minipage}[b]{0.12\linewidth}
			\includegraphics[width=1\textwidth]{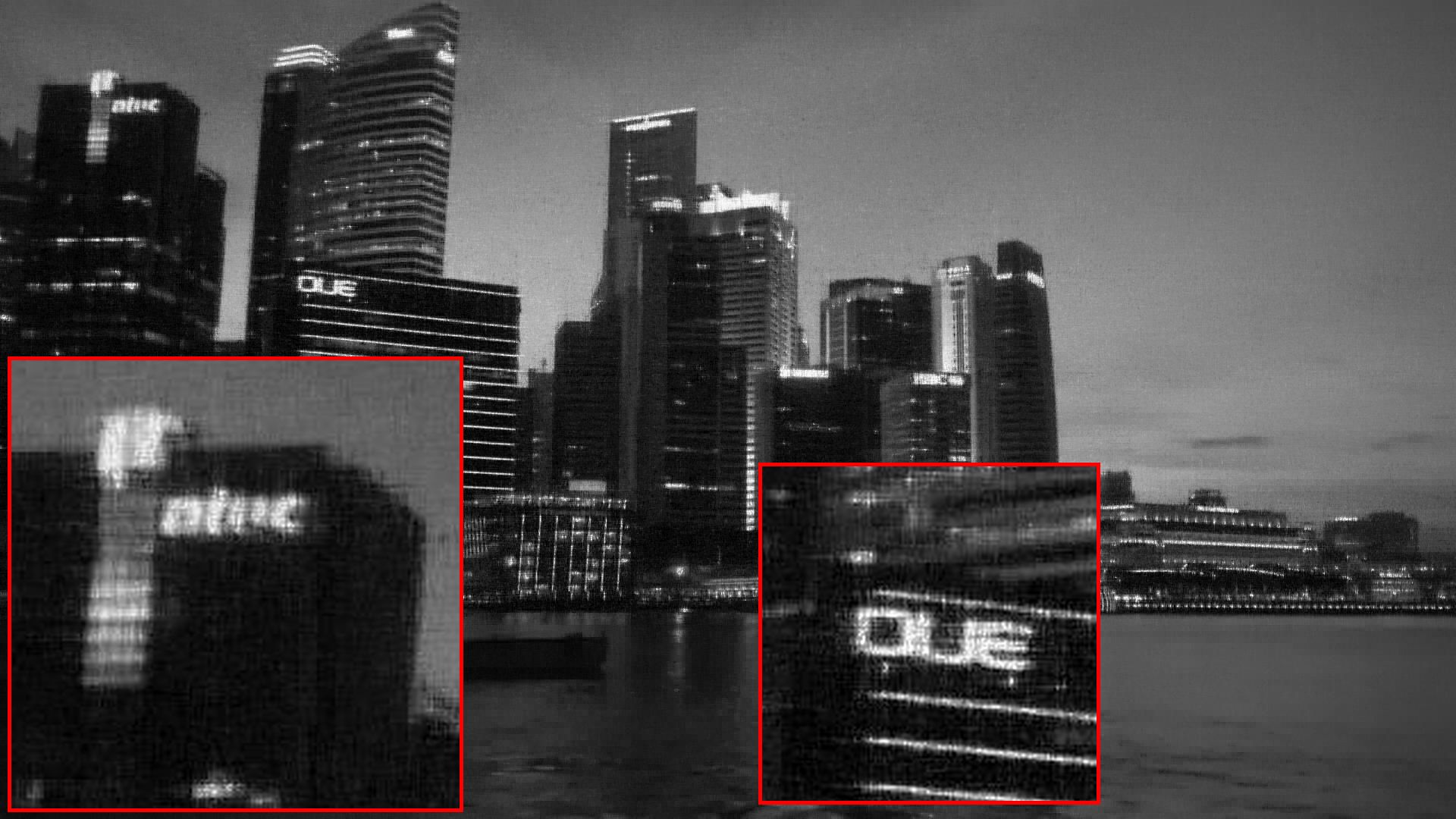}\vskip 2pt
			\includegraphics[width=1\textwidth]{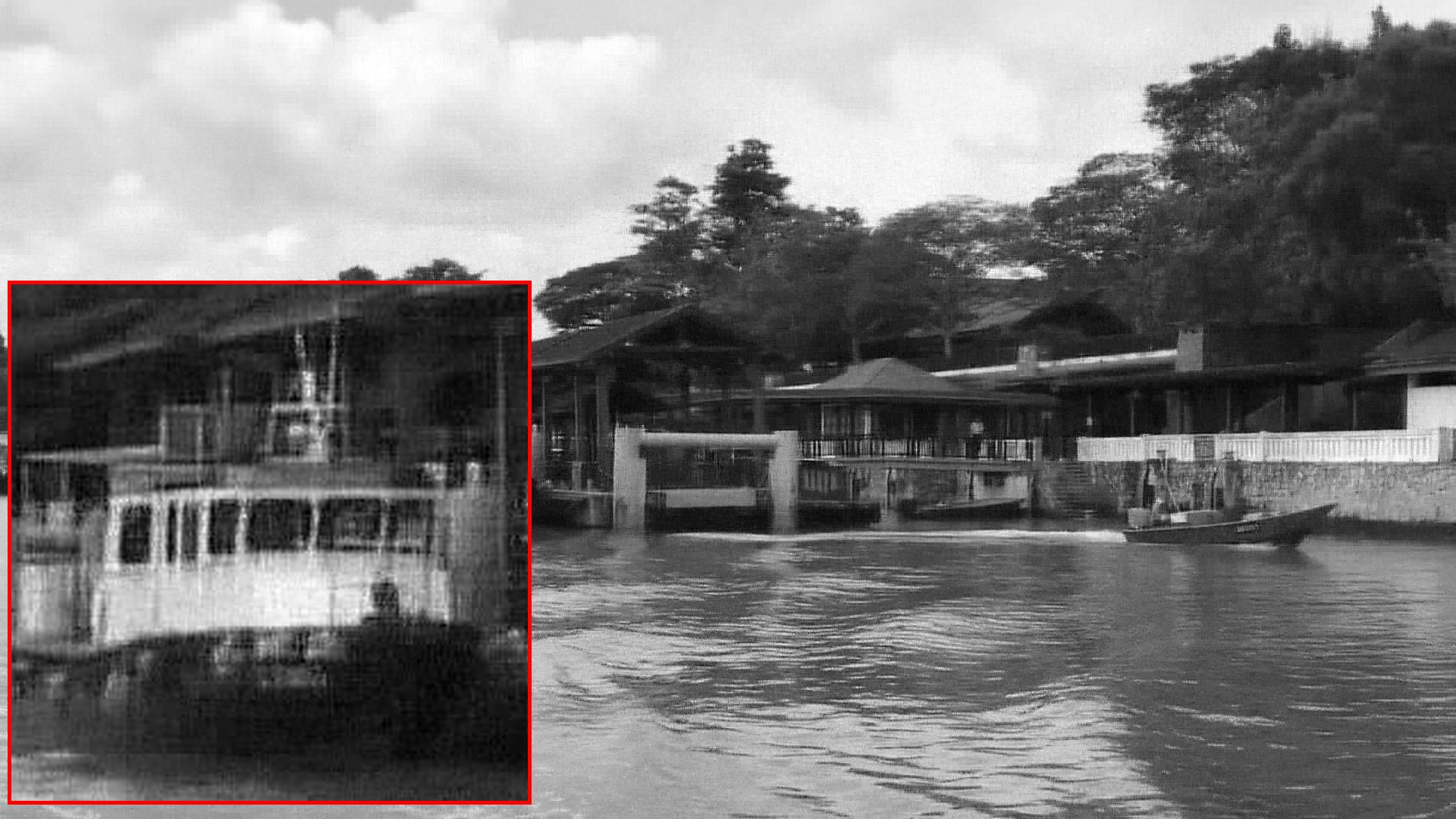}\vskip 2pt
			\includegraphics[width=1\textwidth]{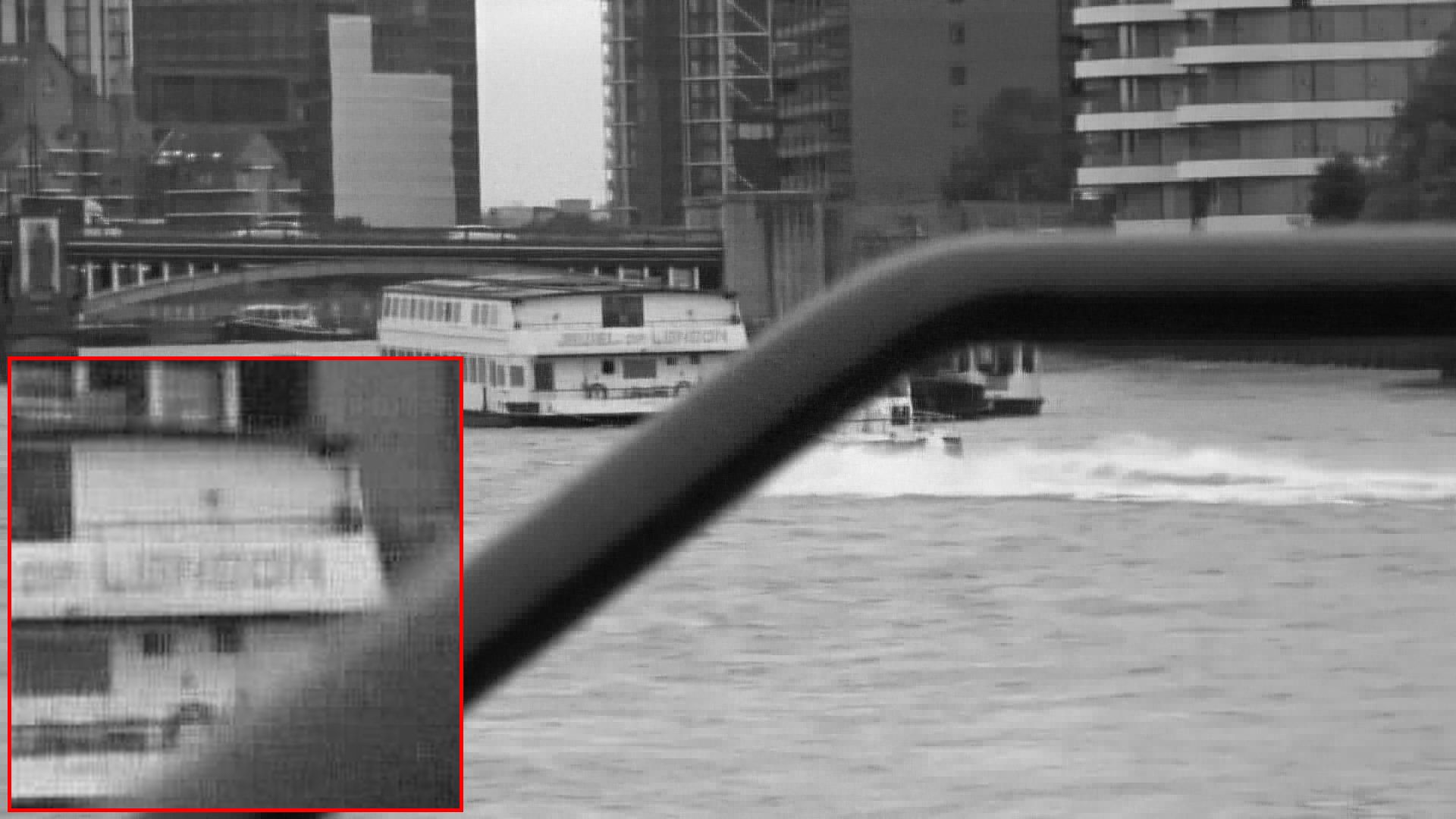}
		\end{minipage}
	\hspace{-12pt}
	}
		\subfigure[TCDLR-RE]{
	\begin{minipage}[b]{0.12\linewidth}
		\includegraphics[width=1\textwidth]{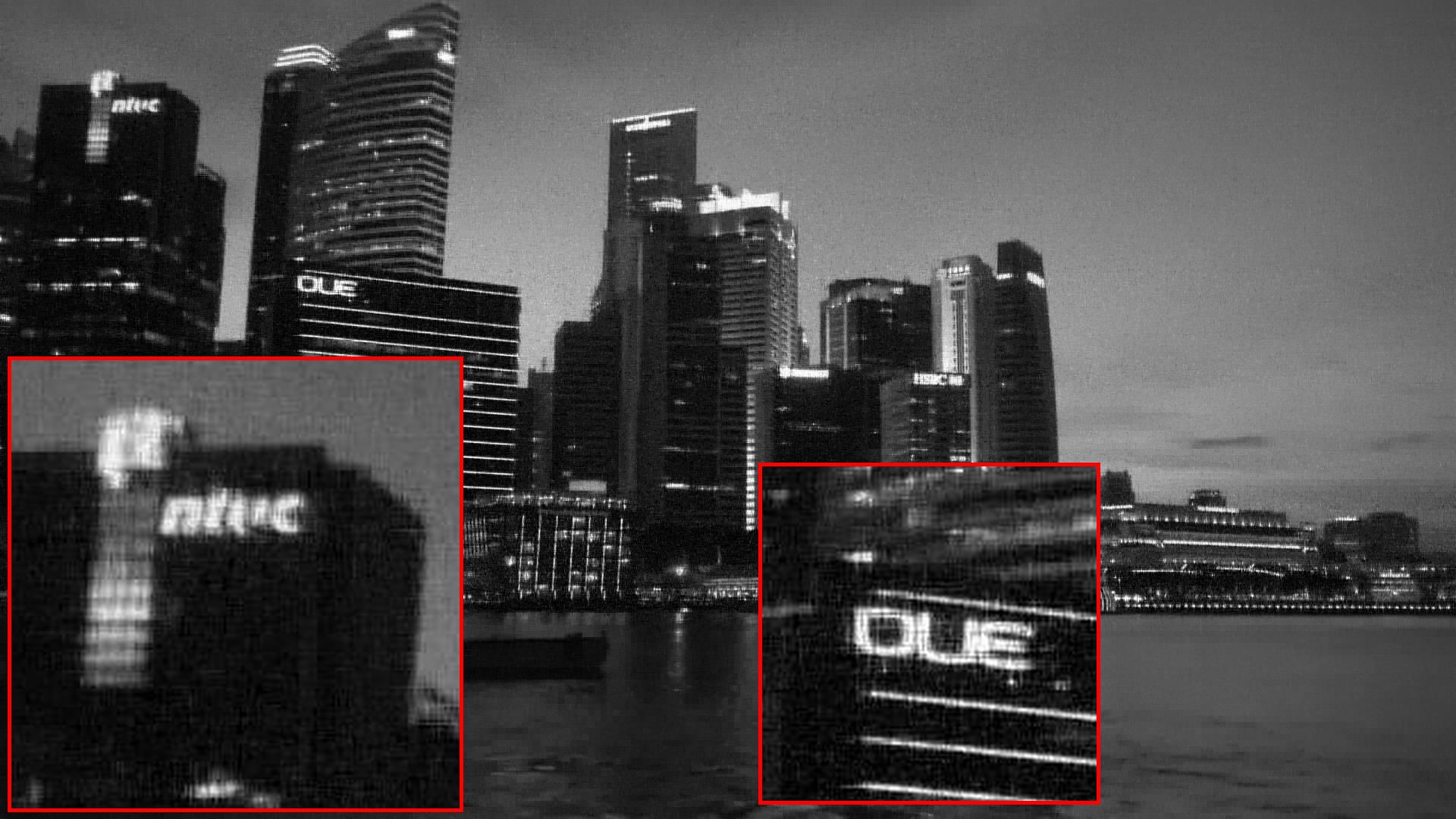}\vskip 2pt
		\includegraphics[width=1\textwidth]{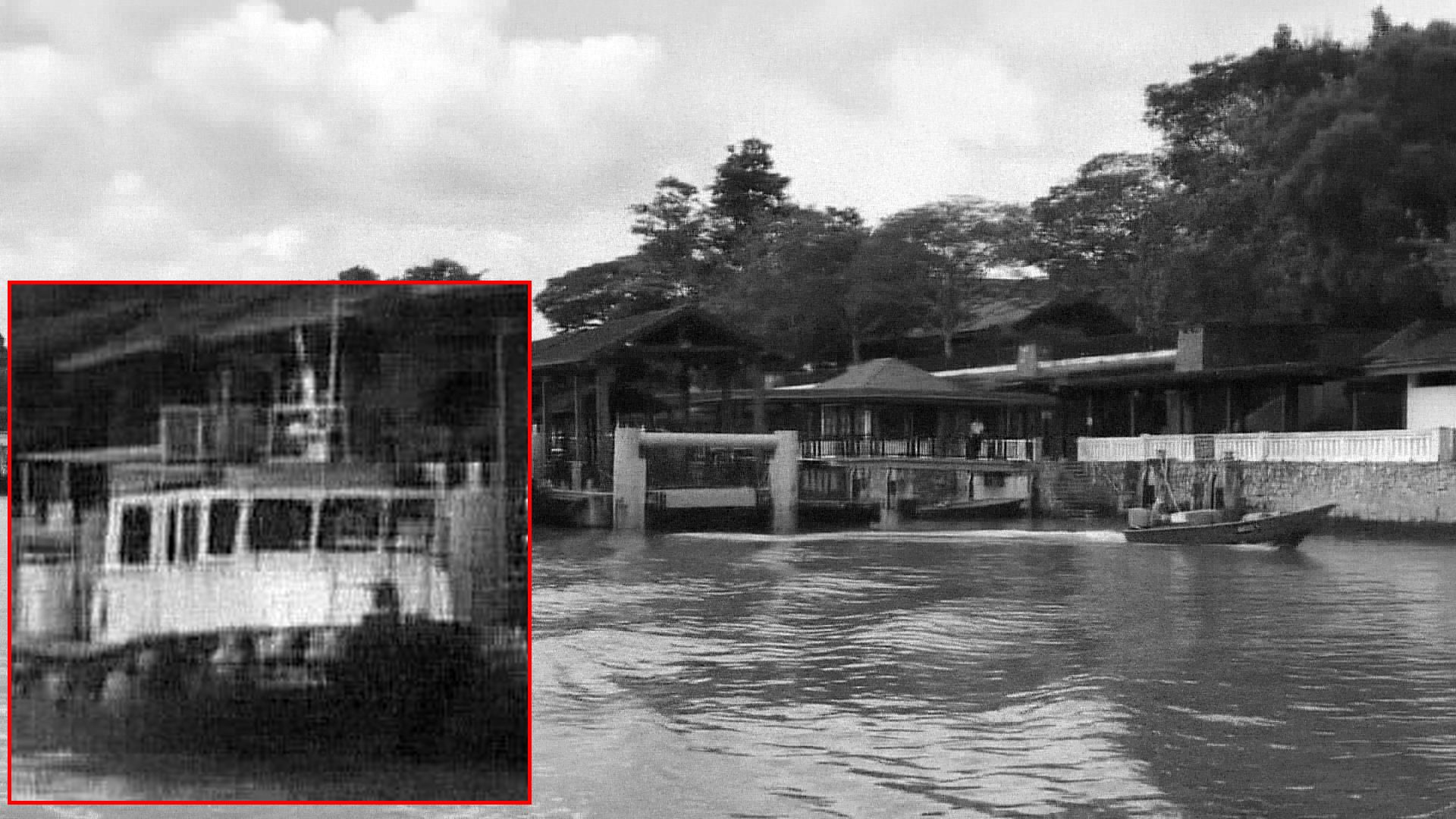}\vskip 2pt
		\includegraphics[width=1\textwidth]{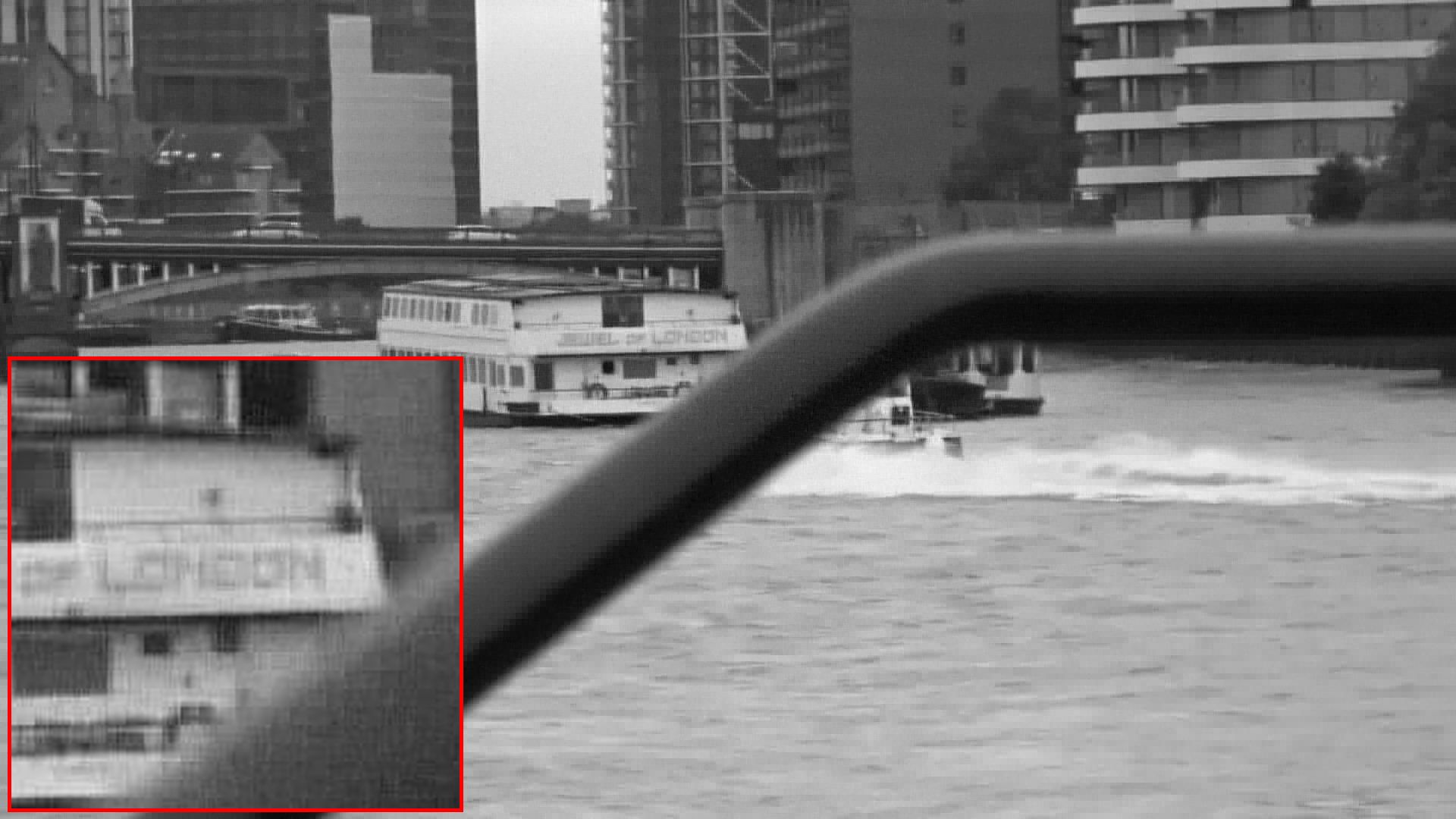}
	\end{minipage}
	\hspace{-12pt}
}
	\caption{The 15th frame of the visual results in the video data. From top to bottom: ``city'', ``Dock'', ``Handrail''.}\label{video_table}
\end{figure*}

All experimental results are given in Table \ref{videos} and Fig. \ref{video_table} to show the effectiveness and efficiency of TCDLR-RE.
From the PSNR results given in Table \ref{videos}, it can be seen that TCDLR-RE performs the best on video inpainting in most of the cases, and it runs the fastest. Especially, compared with TCTF, our method can achieve at least 1.5 times speed-up; compared with other tensor completion methods (including SNN, t-TNN, PSTNN, and TC-RE), our method even achieves more than 10 times speed-up. Fig. \ref{video_table} presents the visual analysis for the three testing videos. The enlarged area of recovery results indicates that TCDLR-RE performs better in restoring the details of letters and ships. These experimental results further demonstrate the superiority of our method for large-scale data.

\begin{table*}[]
\centering
\caption{Comparison of the PSNR and running time (seconds) on videos with a sampling rate of 30\%.}
\resizebox{\textwidth}{!}{%
\begin{tabular}{c|cc|cc|cc|cc|cc|cc|cc}
\hline
\multirow{2}{*}{videos} &
  \multicolumn{2}{c|}{LRMC\cite{36}} &
  \multicolumn{2}{c|}{SNN\cite{12}} &
  \multicolumn{2}{c|}{t-TNN\cite{37}} &
  \multicolumn{2}{c|}{TCTF\cite{29}} & 
  \multicolumn{2}{c|}{PSTNN\cite{jiang2020multi}} &
  \multicolumn{2}{c|}{TC-RE\cite{shi2021robust}} &
  \multicolumn{2}{c}{TCDLR-RE} \\ \cline{2-15} 
 &
  \multicolumn{1}{c|}{PSNR} &
  time(s) &
  \multicolumn{1}{c|}{PSNR} &
  time(s) &
  \multicolumn{1}{c|}{PSNR} &
  time(s) &
  \multicolumn{1}{c|}{PSNR} &
  time(s) & 
  \multicolumn{1}{c|}{psnr} &
  time(s) &
  \multicolumn{1}{c|}{psnr} &
  time(s) &
  \multicolumn{1}{c|}{PSNR} &
  time(s) \\ \hline
\begin{tabular}[c]{@{}c@{}}Dolphin\\ ($1920\times1080\times30$)\end{tabular} &
  \multicolumn{1}{c|}{50.94} &
  2074.26 &
  \multicolumn{1}{c|}{\textbf{53.01}} &
  8563.23 &
  \multicolumn{1}{c|}{50.72} &
  6134.07 &
  \multicolumn{1}{c|}{45.97} &
  728.90 & 
  \multicolumn{1}{c|}{50.87} &
  5431.55 &
  \multicolumn{1}{c|}{18.42} &
  49285.94 &
  \multicolumn{1}{c|}{52.37} &
  \textbf{470.35} \\ \hline
\begin{tabular}[c]{@{}c@{}}City\\ ($1920\times1080\times30$)\end{tabular} &
  \multicolumn{1}{c|}{27.99} &
  2158.47 &
  \multicolumn{1}{c|}{27.83} &
  7483.17 &
  \multicolumn{1}{c|}{27.73} &
  6077.03 &
  \multicolumn{1}{c|}{26.29} &
  744.79 & 
  \multicolumn{1}{c|}{28.01} &
  5482.72 &
  \multicolumn{1}{c|}{23.66} &
  49053.57 &
  \multicolumn{1}{c|}{\textbf{28.91}} &
  \textbf{482.77} \\ \hline
\begin{tabular}[c]{@{}c@{}}Dock\\ ($1920\times1080\times30$)\end{tabular} &
  \multicolumn{1}{c|}{26.87} &
  2064.83 &
  \multicolumn{1}{c|}{27.22} &
  7080.09 &
  \multicolumn{1}{c|}{27.49} &
  5922.21 &
  \multicolumn{1}{c|}{25.34} &
  660.02 & 
  \multicolumn{1}{c|}{27.72} &
  5534.49 &
  \multicolumn{1}{c|}{13.34} &
  50584.37 &
  \multicolumn{1}{c|}{\textbf{28.18}} &
  \textbf{483.94} \\ \hline
\begin{tabular}[c]{@{}c@{}}Ship\\ ($ 1280\times720\times30 $)\end{tabular} &
  \multicolumn{1}{c|}{42.09} &
  713.79 &
  \multicolumn{1}{c|}{45.46} &
  2486.32 &
  \multicolumn{1}{c|}{43.30} &
  1186.47 &
  \multicolumn{1}{c|}{34.97} &
  327.03 & 
  \multicolumn{1}{c|}{43.61} &
  1264.76 &
  \multicolumn{1}{c|}{36.26} &
  10915.48 &
  \multicolumn{1}{c|}{\textbf{46.41}} &
  \textbf{195.41} \\ \hline
\begin{tabular}[c]{@{}c@{}}Handrail\\ ($1920\times1080\times30$)\end{tabular} &
  \multicolumn{1}{c|}{35.44} &
  2019.32 &
  \multicolumn{1}{c|}{36.05} &
  7653.02 &
  \multicolumn{1}{c|}{35.34} &
  6008.16 &
  \multicolumn{1}{c|}{31.32} &
  728.98 & 
  \multicolumn{1}{c|}{35.65} &
  5378.40 &
  \multicolumn{1}{c|}{16.56} &
  48916.63 &
  \multicolumn{1}{c|}{\textbf{37.40}} &
  \textbf{487.75} \\ \hline
\begin{tabular}[c]{@{}c@{}}Penguin\\ ($ 1920\times1080\times30 $)\end{tabular} &
  \multicolumn{1}{c|}{39.92} &
  2062.64 &
  \multicolumn{1}{c|}{42.80} &
  7551.00 &
  \multicolumn{1}{c|}{40.10} &
  6097.18 &
  \multicolumn{1}{c|}{34.83} &
  736.58 & 
  \multicolumn{1}{c|}{40.38} &
  5484.69 &
  \multicolumn{1}{c|}{18.06} &
  49622.41 &
  \multicolumn{1}{c|}{\textbf{42.85}} &
  \textbf{489.57} \\ \hline
\begin{tabular}[c]{@{}c@{}}Leg\\ ($ 1280\times720\times30 $)\end{tabular} &
  \multicolumn{1}{c|}{34.41} &
  680.41 &
  \multicolumn{1}{c|}{37.35} &
  2119.52 &
  \multicolumn{1}{c|}{36.17} &
  1115.00 &
  \multicolumn{1}{c|}{33.94} &
  322.13 & 
  \multicolumn{1}{c|}{36.49} &
  1272.85 &
  \multicolumn{1}{c|}{23.30} &
  10310.20 &
  \multicolumn{1}{c|}{\textbf{38.78}} &
  \textbf{194.84} \\ \hline
\begin{tabular}[c]{@{}c@{}}Chicken\\ ($ 1920\times1080\times30 $)\end{tabular} &
  \multicolumn{1}{c|}{23.89} &
  1937.22 &
  \multicolumn{1}{c|}{\textbf{25.16}} &
  6603.46 &
  \multicolumn{1}{c|}{24.41} &
  5787.27 &
  \multicolumn{1}{c|}{22.02} &
  724.16 & 
  \multicolumn{1}{c|}{24.59} &
  5390.03 &
  \multicolumn{1}{c|}{15.45} &
  49552.65 &
  \multicolumn{1}{c|}{25.07} &
  \textbf{484.86} \\ \hline
\begin{tabular}[c]{@{}c@{}}Bird\\ ($ 1920\times1080\times30 $)\end{tabular} &
  \multicolumn{1}{c|}{27.97} &
  2001.99 &
  \multicolumn{1}{c|}{\textbf{29.19}} &
  6640.75 &
  \multicolumn{1}{c|}{28.37} &
  5795.14 &
  \multicolumn{1}{c|}{26.65} &
  730.30 & 
  \multicolumn{1}{c|}{28.64} &
  5232.23 &
  \multicolumn{1}{c|}{19.24} &
  52635.84 &
  \multicolumn{1}{c|}{29.14} &
  \textbf{487.46} \\ \hline
\begin{tabular}[c]{@{}c@{}}Swan\\ ($ 1280\times720\times30 $)\end{tabular} &
  \multicolumn{1}{c|}{33.95} &
  675.05 &
  \multicolumn{1}{c|}{34.49} &
  2092.67 &
  \multicolumn{1}{c|}{33.82} &
  1133.65 &
  \multicolumn{1}{c|}{25.27} &
  323.54 & 
  \multicolumn{1}{c|}{34.14} &
  1252.79 &
  \multicolumn{1}{c|}{28.24} &
  10113.47 &
  \multicolumn{1}{c|}{\textbf{35.63}} &
  \textbf{193.84} \\ \hline
Average &
  \multicolumn{1}{c|}{34.35} &
  1638.80 &
  \multicolumn{1}{c|}{35.86} &
  5827.32 &
  \multicolumn{1}{c|}{34.74} &
  4525.62 &
  \multicolumn{1}{c|}{30.66} &
  602.64 & 
  \multicolumn{1}{c|}{35.01} &
  4172.45 &
  \multicolumn{1}{c|}{21.25} &
  38099.06 &
  \multicolumn{1}{c|}{\textbf{36.47}} &
  \textbf{397.08} \\ \hline
\end{tabular}\label{videos}
}
\end{table*}


\begin{figure}
	\centering
	
	 	\subfigure[ ]{
		\includegraphics[width=1.5in]{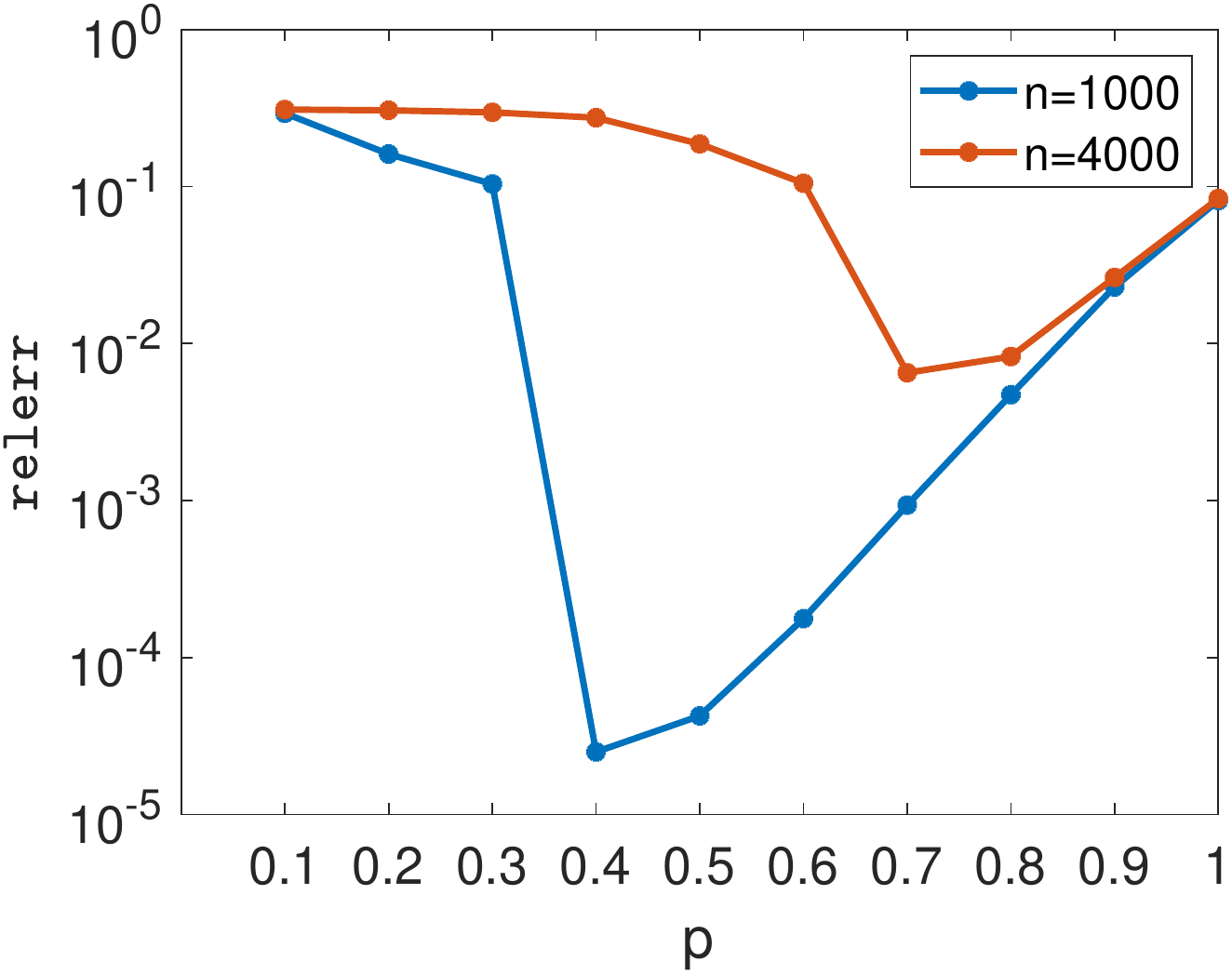}}
	\subfigure[ ]{
		\includegraphics[width=1.5in]{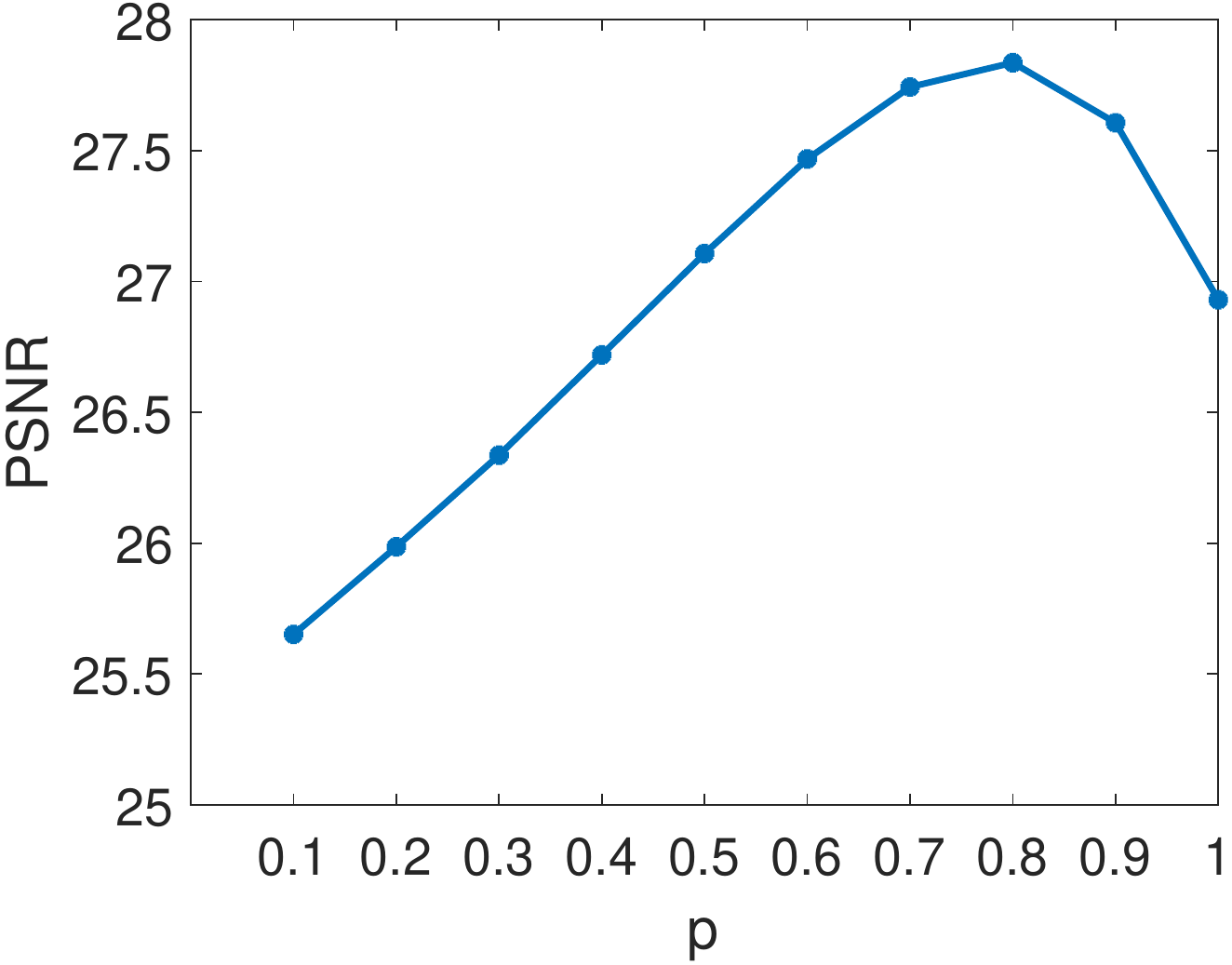}}

\caption{(a) The ${\tt relerr}$ results on the synthetic data with $n\times n \times 3$ for different $p$ when $\mathrm{sampling~rate}=30\%$ and $\bar{r}=0.1n$; (b) The PSNR results on the Berkeley Segmentation Dataset for different $p$ when $\mathrm{sampling~rate}=30\%$. }\label{parameters}
	
\end{figure}
 \subsubsection{Analysis of  Parameters}
 In the proposed method TCDLR-RE, there are two types of parameters: the parameters in the proposed rank estimation (such as $k_{\min}$ and $k_{\max}$), and the parameters in the proposed tensor completion model (TCDLR), where the parameters in the proposed rank estimation can be given by the low-rank prior. Besides, according to the definition of the proposed norm, only one tunable parameter $p$ is involved in TCDLR when $g$ is $\ell_p$.   Therefore, 
in this part, experiments were conducted on both synthetic data and real-world data to investigate the influence of the parameter $p$. All results are presented in Fig. \ref{parameters}, from which it can be seen that our method with $p\in [0.7,0.8]$ achieved more stable performance in tensor recovery.  

\subsection{Results Analysis}


In terms of running time, TCDLR-RE runs much faster than other methods in all cases.  For example, on the GOT-10k video database, TCDLR-RE runs 10 times faster than SNN, t-TNN, PSTNN, and TC-RE, three times faster than LRMC, and 1.5 faster than TCTF. The reasons are analysed as follows.  (1)  Owing to the  proposed dual low-rank constraint, the t-SVD of a smaller tensor  $\mathcal{Z}$ is computed, and the t-SVD operation with large time consumption is avoided, thus reducing the total cost at each iteration of the algorithm from $\mathcal{O}(n_{(1)}n_{(2)}^2n_3+n_1n_2n_3\log n_3)$ to $\mathcal{O}(kn_1n_2n_3+n_1n_2n_3\log n_3)$, where $k$ is an estimation of the tensor rank. This indicates that TCDLR-RE has much lower computation complexity than traditional methods (including LRMC, SNN, t-TNN, and PSTNN) based on t-SVD for the case of data tensors with a low rank (\ie, $k\ll\min(m,n)$). (2)   
 Fig.~\ref{loss} (c) shows that the developed rank estimation method in TCDLR-RE can estimate the tensor rank accurately. The comparison of TCDLR and TCDLR-RE in Table \ref{bgu_table} demonstrates the contribution of the proposed rank estimation method for reducing the running time. (3) Besides, as shown in Fig. \ref{loss} (a)-(b), TCDLR-RE converged to the optimal solution with fewer iterations than most of the comparison methods. 

Except for TCTF, most of the tensor-based methods (including SNN, t-TNN, PSTNN, TC-RE, and TCDLR-RE) achieved a better performance than the matrix-based method (LRMC). This is because tensor-based methods can exploit the low-rank structure in the tensor data and utilize the relationship between different tensor slices well compared with the matrix-based method. Compared with tensor-based methods, TCDLR-RE achieved the best results in real-world applications. Specifically, the gap between the average PSNR values of TCDLR-RE and the other methods is about 1 dB on the BiHID. This is because (1) Compared with the traditional tensor methods, the proposed dual low-rank constraint in our methods (including TCDLR and TCDLR-RE) effectively addresses the issue of over-punishment in t-TNN-based methods and fully utilizes the low-rankness prior in tensor data, which helps to obtain the low-rank estimation more accurately. 
(2) Compared with the existing tensor factorization methods including TCTF and TC-RE, the performance of TCDLR-RE is robust to the tensor rank estimation as proved by   Property \ref{Pn1}, Theorem \ref{th1} and the comparison of TCDLR with other methods in Table \ref{bgu_table}.   
  (3) Compared with the estimation strategy adopted in TCTF, the proposed estimation method can estimate true tubal rank more accurately as proved by the comparison in Fig.~\ref{loss} (c). 



\section{Conclusion} 

 This work presents an efficient and effective tensor completion algorithm. First, aiming at the over-punishment issue in t-TNN-based methods and the difficulty in estimating the true tensor rank for tensor data with a small sampling rate, a novel low-rank tensor completion method with a new tensor norm is proposed, which utilizes the dual low-rank constraint. The proposed tensor norm enables the proposed methods to be more robust to inaccurate rank estimation and recover the low-rank tensor more accurately. Meanwhile, to avoid the high time consumption caused by performing the t-SVD operation on a large tensor, a trick to compute the t-SVD of a smaller tensor $\mathcal{Z} \in \mathbb{R}^{n_1 \times k \times n_3}$ instead of the original tensor of size $\mathbb{R}^{n_1 \times k \times n_3}$  is used to solve TCDLR, thus reducing the total cost at each iteration to $\mathcal{O}(n_1n_2n_3\log n_3 +kn_1n_2n_3)$ from $\mathcal{O}(n_{(1)}n_{(2)}^2n_3+n_1n_2n_3\log n_3)$ achieved by standard t-SVD-based methods. Based on this, a novel estimation method with rank-increasing and decreasing strategies is proposed for estimating the tensor rank $k$. The experimental results demonstrate the high effectiveness and efficiency of the proposed methods. 
 
 It is worth noting that the of TCDLR and TCDLR-RE can be generalized to the case of higher-order tensor easily by utilizing the higher-order tensor product defined in \cite{martin2013order} (as well as the real invertible linear transforms-based tensor product \cite{lu2019low} and its generalization to the higher-order tensor case \cite{liu2022tensor}). Following most works related to tensor product-based methods, we focuses on the case of three-order tensors. Since matrix completion is a special case of tensor completion, the proposed TCDLR-RE can be applied to matrix completion problems. Moreover, the proposed dual low-rank constraint can   be applied to other low-rank recovery problems (such as the case of the tensor data with various types of noise \cite{wright2009robust,zhang2019accurate,zhang2018exploiting} and the low-rank representation problem \cite{liu2010robust}), and it can also be combined with non-local methods \cite{song2018nonlocal,wang2022non} to achieve better performance in image and video recovery.


\section{Appendix}
\subsection{Proofs of Property \ref{Pn1} and Theorem~\ref{th1}}
 
\newtheorem*{property1}{Property~1}
\begin{property1} For $\mathcal{Y}, \mathcal{X}\in \mathbb{R}^{n_1 \times n_2 \times n_3}$, we have 
\begin{itemize}
\item[(1)]  $\|\mathcal{Y}-\mathcal{X}\|^2_F\geq \frac{1}{n_3}\sum_i(\sigma_i(\bar{Y})-\sigma_i(\bar{X}))^2$;
    
    \item[(2)] \textbf{$\|\mathcal{Y}-\mathcal{X}\|^2_F\geq \frac{1}{n_3}\sum_{i=1}^{n_3}  \sum_{\mathrm{rank}(\mathcal{Y})<j\leq \mathrm{rank}(\mathcal{X})}\sigma_j(\bar{X}_i)^2$ if $\mathrm{rank}(\mathcal{Y})<\mathrm{rank}(\mathcal{X})$.} 
\end{itemize}
\end{property1} 
\begin{proof}
  (1) According to $\left<\bar{Y},\bar{X}\right>=\mathrm{Re}(\left<\bar{Y},\bar{X}\right>)$ \cite{Zhang2022Tensor} and $\mathrm{Re}(\mathrm{tr}(\bar{Y}\bar{X}^*))\leq \sum_{i}\sigma_i(\bar{Y})\sigma_i(\bar{X})$ \cite{marshall1979inequalities,komaroff2008enhancements},  $\|\mathcal{Y}-\mathcal{X}\|^2_F=\|\mathcal{Y}\|_F^2+\|\mathcal{X}\|_F^2-2\left<\mathcal{Y},\mathcal{X}\right>=\frac{1}{n_3}(\|\bar{Y}\|_F^2+\|\bar{X}\|_F^2-2 \left<\bar{Y},\bar{X}\right>)=\frac{1}{n_3}(\|\bar{Y}\|_F^2+\|\bar{X}\|_F^2-2 \mathrm{Re}(\left<\bar{Y},\bar{X}\right>))\geq \frac{1}{n_3}\sum_i(\sigma_i(\bar{Y})-\sigma_i(\bar{X}))^2$.  The conclusion holds.
  
(2)   Property \ref{Pn1}  (2) can be concluded from Property \ref{Pn1} (1). 
\end{proof} 
\newtheorem*{theorem1}{Theorem~1}
 \begin{theorem1}  
Let $ \mathring{\mathcal{X}}$,  $\hat{\mathcal{X}}$,  and $ \widetilde{\mathcal{X}}$  be the optimal solutions to \eqref{TClg}, \eqref{TCDLR},  and \eqref{TC}, respectively.  
 Then, we have: 
 \begin{itemize}

      \item[(i)]  $\mathrm{rank}(\mathring{\mathcal{X}})\geq \mathrm{rank}(\widetilde{\mathcal{X}})$ holds;
     \item[(ii)] If $k \geq \mathrm{rank}(\mathring{\mathcal{X}})$,  $\hat{\mathcal{X}}$ is an optimal solution of \eqref{TClg} and  $\mathring{\mathcal{X}}$ is an optimal solution of \eqref{TCDLR}; 
            \item[(iii)]  If $k=\mathrm{rank}(\widetilde{\mathcal{X}})$,      $\hat{\mathcal{X}}$ is an optimal solution of \eqref{TC};  
            \item[(iv)]    $k<\mathrm{rank}(\widetilde{\mathcal{X}})$ holds if and only if    $\|\hat{\mathcal{X}}\|_{*,(k,g)}=\infty$. 
             \end{itemize}
\end{theorem1}
 \begin{proof}
    (i) The conclusion holds by the definitions of  $\hat{\mathcal{X}}$.
    
    (ii) According to the definition of $\mathring{\mathcal{X}}$, $ \|\mathring{\mathcal{X}}\|_{*,g}\leq \|\hat{\mathcal{X}}\|_{*,g} $ and $\mathbf{P}_{\Omega}(\mathcal{M})=\mathbf{P}_{\Omega}(\mathring{\mathcal{X}})$.  It is worth noting that $ \|\mathring{\mathcal{X}}\|_{*,g}\leq \|\mathbf{P}_{\Omega}(\mathcal{M})\|_{*,g}<\infty$ by $\mathbf{P}_{\Omega}(\mathcal{M})=\mathbf{P}_{\Omega}(\mathbf{P}_{\Omega}(\mathcal{M}))$.  Therefore, we have   $\|\mathring{\mathcal{X}}\|_{*,(k,g)}=\|\mathring{\mathcal{X}}\|_{*,g} \leq \|\hat{\mathcal{X}}\|_{*,(k,g)}$ according to $k \geq \mathrm{rank}(\mathring{\mathcal{X}})$.  Thus,    $\mathring{\mathcal{X}}$ is an optimal solution to \eqref{TCDLR}. 

 Meanwhile, according to the definitions of  $\hat{\mathcal{X}}$, we have $\|\hat{\mathcal{X}}\|_{*,(k,g)} \leq \|\mathring{\mathcal{X}}\|_{*,(k,g)}=\|\mathring{\mathcal{X}}\|_{*,g}<\infty$  and $\mathbf{P}_{\Omega}(\mathcal{M})=\mathbf{P}_{\Omega}(\hat{\mathcal{X}})$. Thus, $\|\hat{\mathcal{X}}\|_{*,(k,g)}= \|\hat{\mathcal{X}}\|_{*,g}\leq \|\mathring{\mathcal{X}}\|_{*,g}$ holds, which implies that $\hat{\mathcal{X}}$ is an optimal solution of \eqref{TClg}. 
     
   (iii) According to $\mathrm{rank}(\hat{\mathcal{X}}) \leq k=\mathrm{rank}(\widetilde{\mathcal{X}})$ and $\mathbf{P}_{\Omega}(\mathcal{M})=\mathbf{P}_{\Omega}(\hat{\mathcal{X}})$,  $\hat{\mathcal{X}}$ is an optimal solution of \eqref{TC}.

   (iv)  If $k<\mathrm{rank}(\widetilde{\mathcal{X}})$, it will be proven that $\|\hat{\mathcal{X}}\|_{*,(k,g)}=\infty$ holds by contradiction.   
   Assuming  $\|\hat{\mathcal{X}}\|_{*,(k,g)}<\infty$, it can be concluded that  $\mathrm{rank}(\hat{\mathcal{X}})\leq k$. Therefore, we have $ \mathrm{rank}(\widetilde{\mathcal{X}}) \leq \mathrm{rank}(\hat{\mathcal{X}})\leq k$ according to the definitions of $\widetilde{\mathcal{X}}$, which contracts the fact that $k<\mathrm{rank}(\widetilde{\mathcal{X}})$. 

 Then, if $\|\hat{\mathcal{X}}\|_{*,(k,g)}=\infty$, 
the set $\{\mathcal{X}|\mathbf{P}_{\Omega}(\mathcal{M})=\mathbf{P}_{\Omega}(\mathcal{X})  ~and~  \mathrm{rank}(\mathcal{X}) \leq k \}$ is empty. Thus, $k<\mathrm{rank}(\widetilde{\mathcal{X}})$  holds.


\end{proof}

\subsection{Proof of Theorem~\ref{th2}}
 
 Since $I_{kn_3}=\bar{Q}\mathrm{diag}(\mathrm{fft}(\mathcal{Q}^*,[],3))= \bar{Q}\bar{Q}^*$ if $\mathcal{I}_{k\times k \times n_3 }=\mathcal{Q}\ast\mathcal{Q}^*$, we have $\sigma(\bar{B}\bar{Q})=\sigma(\bar{B})$  and the following conclusions \cite{zhang2017matrix}.  

  \begin{table*}
\caption{ Comparison of the four metrics on  the SBMnet-2016 dataset.} 
\centering
	\scalebox{0.7}{ 
\resizebox{\textwidth}{!}{%
\begin{tabular}{|c|c|c|c|c|c|c|}
\hline
Method  & Average AGE   & Average pCEPS & Average MSSSIM & Average PSNR     & time(s)  \\ \hline
TNN\cite{hu2015new}    & 4.2668          & 0.0311        & 0.9449         & 31.8775      & 120.70 \\ \hline
BRTF\cite{zhao2015bayesian}    & 3.4799            & 0.0073        & 0.9674         & 32.4782       & 113.04 \\ \hline
T-SVD\cite{22} & 4.24            & 0.0300        & 0.9452         & 31.9453       & 604.98  \\ \hline
TCDLR-RE    & 4.3613             & 0.0304        & 0.9451         & 31.7373       & 67.40  \\ \hline
\end{tabular}\label{background}
}}
\end{table*}
\begin{property}\label{P3}
 If $\mathcal{B}\in \mathbb{R}^{n_1\times k \times n_3}$, $\mathcal{Q}\in \mathbb{R}^{k\times n_2 \times n_3}$ and  $\mathcal{Q}\ast\mathcal{Q}^*=\mathcal{I}_{k\times k \times n_3}$, then 
	\begin{itemize}
		\item[(a)] $\|\mathcal{B}\ast\mathcal{Q}\|_F=\|\mathcal{B}\|_F$.
		\item[(b)]  $\|\mathcal{B}\ast\mathcal{Q}\|_{*,g}=\|\mathcal{B}\|_{*,g}$.
	\end{itemize}
\end{property}
\newtheorem*{theorem2}{Theorem~2}
\begin{theorem2} 
Let $ \mathcal{Y}=\mathcal{A}\ast\mathcal{B}$, where $ \mathcal{A} \in \mathbb{R}^{n_1 \times k  \times n_3} $ and $ \mathcal{B} \in \mathbb{R}^{k \times n_2  \times n_3}$.
If $  \mathcal{B}^*=\mathcal{Q}^*\ast\mathcal{R}^*$ is  the QR decomposition  of $\mathcal{B}^*$ and $\mathcal{Z}=\mathcal{A}\ast \mathcal{R}$,  $$ \mathcal{Y}=\mathcal{Z}\ast\mathcal{Q}$$ and
$$\mathbb{S}_{\tau,g}(\mathcal{Y})=\mathbb{S}_{\tau,g}(\mathcal{Z})\ast\mathcal{Q}$$ hold,  where $ \mathcal{Z} \in \mathbb{R}^{n_1 \times k  \times n_3} $, $ \mathcal{Q} \in \mathbb{R}^{k \times n_2  \times n_3}$, and  
$$ \mathbb{S}_{\tau,g}(\mathcal{Y})=\mathop{\arg\min}_{\mathcal{X}} \tau\|\mathcal{X}\|_{*,g}+\frac{1}{2}\|\mathcal{Y}-\mathcal{X}\|_F^2.$$
\end{theorem2}
\begin{proof}
Since $  \mathcal{B}^*=\mathcal{Q}^*\ast\mathcal{R}^*$ is  the QR decomposition  of $\mathcal{B}^*$, $ \mathcal{Q}$  satisfies   $\mathcal{Q}\ast\mathcal{Q}^*=\mathcal{I}_{k\times k \times n_3}$ and  $ \mathcal{Y}=\mathcal{A}\ast\mathcal{R}\ast\mathcal{Q}=\mathcal{Z}\ast\mathcal{Q}$.
From $\mathcal{Q}\ast\mathcal{Q}^*=\mathcal{I}_{k\times k \times n_3}$ and Property \ref{P3} (a), we can get
      \begin{align}\label{pfe1}
		 &\mathbb{S}_{\tau,g}(\mathcal{Z})\ast\mathcal{Q} \notag\\
		  =& \mathop{\arg\min}_{\mathcal{B}}\left(\tau\|\mathcal{B}\|_{*,g}+\frac{1}{2}\|\mathcal{Z}-\mathcal{B}\|_{F}^{2}\right)\ast \mathcal{Q}  \notag\\ 
		=& \mathop{\arg\min}_{\mathcal{B}}\left(\tau\left\| \mathcal{B}\right\|_{*,g}+\frac{1}{2}\|(\mathcal{Z}-\mathcal{B})\ast\mathcal{Q}\|_{F}^{2}\right)\ast\mathcal{Q}.
		\end{align} 
		Since   $\|\mathcal{B}\|_{*,g}=\| \mathcal{B}\ast\mathcal{Q}\|_{*,g}$ by Property \ref{P3} (b),
		\begin{align}\label{pfe2} 
		& ~~~\mathop{\arg\min}_{\mathcal{B}}\left(\tau\left\| \mathcal{B} \right\|_{*,g}+\frac{1}{2}\|(\mathcal{Z}-\mathcal{B})\ast\mathcal{Q}\|_{F}^{2}\right)\ast\mathcal{Q}\notag\\ 
		&=\mathop{\arg\min}_{\mathcal{B}}\left(\tau\|\mathcal{B}\ast\mathcal{Q}\|_{*,g}+\frac{1}{2}\|(\mathcal{Z}-\mathcal{B})\ast\mathcal{Q}\|_{F}^{2}\right)\ast\mathcal{Q}.  
		\end{align} 
			By letting $\mathcal{B}\ast\mathcal{Q}=\mathcal{X}$,
		\begin{align}\label{pfe3} 
		& ~~~\mathop{\arg\min}_{\mathcal{B}}\left(\tau\|\mathcal{B}\ast\mathcal{Q}\|_{*,g}+\frac{1}{2}\|(\mathcal{Z}-\mathcal{B})\ast\mathcal{Q}\|_{F}^{2}\right)\ast\mathcal{Q}\notag\\ 
		&=\mathop{\arg\min}_{\mathcal{X}}\left(\tau\|\mathcal{X}\|_{*,g}+\frac{1}{2}\| \mathcal{Z}\ast\mathcal{Q}-\mathcal{X}\|_{F}^{2}\right) \notag\\
		&=\mathbb{S}_{\tau,g}(\mathcal{ Z}\ast\mathcal{Q})=\mathbb{S}_{\tau,g}(\mathcal{Y}).
		\end{align} 
		Combining \eqref{pfe1}, \eqref{pfe2} and \eqref{pfe3}, we have 
			\[ \mathbb{S}_{\tau,g}(\mathcal{Y})=\mathbb{S}_{\tau,g}(\mathcal{Z})\ast\mathcal{Q}.  \]
\end{proof}

\subsection{Tensor Completion for background initialization}
In this part, 
 TCDLR-RE is compared with several state-of-the-art  methods including TNN\cite{hu2015new}, BRTF\cite{zhao2015bayesian}, and T-SVD\cite{22} in background initialization on the four video sequences from SBMnet-2016\footnote{http://scenebackgroundmodeling.net/}, including  ``Candela\_m1.10'', ``CAVIAR1'', ``CaVignal'', and ``HallAndMonitor''.  

Some implementation details are given below: (1) For our method, the background initialization framework provided in \cite{sobral2017matrix} was used, and the part related to tensor completion was replaced with our method (TCDLR-RE). Considering the strong low-rankness in the background frame, $k_{\max}$ was set to $k_{\max}=0.1\times\min(n_1,n_2)$, and the remaining parameters were the same as those in the experiments of the main body. (2) For TNN and T-SVD, they are free parameters, and the same background initialization framework as ours was adopted. (3) For BRTF, the parameters were set as suggested by the SBMnet-2016's website. (4) Four metrics, including AGE, pCEPS, MSSSIM, and PSNR \cite{sobral2017matrix} \footnote{ The higher MSSIM and PSNR values and the lower AGE and pCEPS values indicate better performance.}, were taken to evaluate the performance of different methods in the background initialization.  

 The average results on the four video sequences are presented in Table \ref{background}. As shown in the table, compared with TNN and T-SVD which adopt the same background initialization framework as ours, TCDLR-RE achieved a comparable performance in background initialization. Also, TCDLR-RE had significantly less running time than other methods.



\bibliography{reference.bib} 

\begin{thebibliography}{10}
\providecommand{\url}[1]{#1}
\csname url@samestyle\endcsname
\providecommand{\newblock}{\relax}
\providecommand{\bibinfo}[2]{#2}
\providecommand{\BIBentrySTDinterwordspacing}{\spaceskip=0pt\relax}
\providecommand{\BIBentryALTinterwordstretchfactor}{4}
\providecommand{\BIBentryALTinterwordspacing}{\spaceskip=\fontdimen2\font plus
\BIBentryALTinterwordstretchfactor\fontdimen3\font minus
  \fontdimen4\font\relax}
\providecommand{\BIBforeignlanguage}[2]{{%
\expandafter\ifx\csname l@#1\endcsname\relax
\typeout{** WARNING: IEEEtran.bst: No hyphenation pattern has been}%
\typeout{** loaded for the language `#1'. Using the pattern for}%
\typeout{** the default language instead.}%
\else
\language=\csname l@#1\endcsname
\fi
#2}}
\providecommand{\BIBdecl}{\relax}
\BIBdecl

\bibitem{1}
A.~Cichocki, D.~Mandic, L.~De~Lathauwer, G.~Zhou, Q.~Zhao, C.~Caiafa, and H.~A.
  Phan, ``Tensor decompositions for signal processing applications: From
  two-way to multiway component analysis,'' \emph{IEEE Signal Processing
  Magazine}, vol.~32, no.~2, pp. 145--163, 2015.

\bibitem{34}
C.~Lu, J.~Feng, Y.~Chen, W.~Liu, Z.~Lin, and S.~Yan, ``Tensor robust principal
  component analysis with a new tensor nuclear norm,'' \emph{IEEE Transactions
  on Pattern Analysis and Machine Intelligence}, vol.~42, no.~4, pp. 925--938,
  2019.

\bibitem{wu2018fused}
Y.~Wu, H.~Tan, Y.~Li, J.~Zhang, and X.~Chen, ``A fused {CP} factorization
  method for incomplete tensors,'' \emph{IEEE Transactions on Neural Networks
  and Learning Systems}, vol.~30, no.~3, pp. 751--764, 2018.

\bibitem{shi2018feature}
Q.~Shi, Y.-M. Cheung, Q.~Zhao, and H.~Lu, ``Feature extraction for incomplete
  data via low-rank tensor decomposition with feature regularization,''
  \emph{IEEE Transactions on Neural Networks and Learning Systems}, vol.~30,
  no.~6, pp. 1803--1817, 2018.

\bibitem{candes2011robust}
E.~J. Cand{\`e}s, X.~Li, Y.~Ma, and J.~Wright, ``Robust principal component
  analysis?'' \emph{Journal of the ACM (JACM)}, vol.~58, no.~3, pp. 1--37,
  2011.

\bibitem{zhao2015bayesian}
Q.~Zhao, G.~Zhou, L.~Zhang, A.~Cichocki, and S.-I. Amari, ``Bayesian robust
  tensor factorization for incomplete multiway data,'' \emph{IEEE Transactions
  on Neural Networks and Learning Systems}, vol.~27, no.~4, pp. 736--748, 2015.

\bibitem{sobral2017matrix}
A.~Sobral and E.-h. Zahzah, ``Matrix and tensor completion algorithms for
  background model initialization: A comparative evaluation,'' \emph{Pattern
  Recognition Letters}, vol.~96, pp. 22--33, 2017.

\bibitem{kajo2018svd}
I.~Kajo, N.~Kamel, Y.~Ruichek, and A.~S. Malik, ``Svd-based tensor-completion
  technique for background initialization,'' \emph{IEEE Transactions on Image
  Processing}, vol.~27, no.~6, pp. 3114--3126, 2018.

\bibitem{kajo2019self}
I.~Kajo, N.~Kamel, and Y.~Ruichek, ``Self-motion-assisted tensor completion
  method for background initialization in complex video sequences,'' \emph{IEEE
  Transactions on Image Processing}, vol.~29, pp. 1915--1928, 2019.

\bibitem{madathil2018twist}
B.~Madathil and S.~N. George, ``Twist tensor total variation
  regularized-reweighted nuclear norm based tensor completion for video missing
  area recovery,'' \emph{Information Sciences}, vol. 423, pp. 376--397, 2018.

\bibitem{gandy2011tensor}
S.~Gandy, B.~Recht, and I.~Yamada, ``Tensor completion and low-n-rank tensor
  recovery via convex optimization,'' \emph{Inverse Problems}, vol.~27, no.~2,
  p. 025010, 2011.

\bibitem{Koldakiers2000towards}
H.~A. Kiers, ``Towards a standardized notation and terminology in multiway
  analysis,'' \emph{Journal of Chemometrics: A Journal of the Chemometrics
  Society}, vol.~14, no.~3, pp. 105--122, 2000.

\bibitem{kolda2009tensor}
T.~G. Kolda and B.~W. Bader, ``Tensor decompositions and applications,''
  \emph{SIAM Review}, vol.~51, no.~3, pp. 455--500, 2009.

\bibitem{12}
J.~Liu, P.~Musialski, P.~Wonka, and J.~Ye, ``Tensor completion for estimating
  missing values in visual data,'' \emph{IEEE Transactions on Pattern Analysis
  and Machine Intelligence}, vol.~35, no.~1, pp. 208--220, 2012.

\bibitem{zhang2019robust}
X.~Zhang, D.~Wang, Z.~Zhou, and Y.~Ma, ``Robust low-rank tensor recovery with
  rectification and alignment,'' \emph{IEEE Transactions on Pattern Analysis
  and Machine Intelligence}, vol.~43, no.~1, pp. 238--255, 2019.

\bibitem{liu2016low}
X.-Y. Liu, S.~Aeron, V.~Aggarwal, and X.~Wang, ``Low-tubal-rank tensor
  completion using alternating minimization,'' \emph{IEEE Transactions on
  Information Theory}, vol.~66, no.~3, pp. 1714--1737, 2019.

\bibitem{guichardet2006symmetric}
A.~Guichardet, \emph{Symmetric Hilbert spaces and related topics: Infinitely
  divisible positive definite functions. Continuous products and tensor
  products. Gaussian and Poissonian stochastic processes}.\hskip 1em plus 0.5em
  minus 0.4em\relax Springer, 2006, vol. 261.

\bibitem{frank1993statistical}
L.~E. Frank and J.~H. Friedman, ``A statistical view of some chemometrics
  regression tools,'' \emph{Technometrics}, vol.~35, no.~2, pp. 109--135, 1993.

\bibitem{geman1992constrained}
D.~Geman and G.~Reynolds, ``Constrained restoration and the recovery of
  discontinuities,'' \emph{IEEE Transactions on Pattern Analysis and Machine
  Intelligence}, vol.~14, no.~3, pp. 367--383, 1992.

\bibitem{trzasko2008highly}
J.~Trzasko and A.~Manduca, ``Highly undersampled magnetic resonance image
  reconstruction via homotopic $\ell_0$ -minimization,'' \emph{IEEE
  Transactions on Medical Imaging}, vol.~28, no.~1, pp. 106--121, 2008.

\bibitem{Malioutov2013Iterative}
D.~Malioutov and A.~Aravkin, ``Iterative log thresholding,'' in \emph{IEEE
  International Conference on Acoustics, Speech and Signal Processing}, 2013,
  pp. 7198--7202.

\bibitem{friedman2012fast}
J.~H. Friedman, ``Fast sparse regression and classification,''
  \emph{International Journal of Forecasting}, vol.~28, no.~3, pp. 722--738,
  2012.

\bibitem{gao2011feasible}
C.~Gao, N.~Wang, Q.~Yu, and Z.~Zhang, ``A feasible nonconvex relaxation
  approach to feature selection,'' in \emph{Proceedings of the AAAI Conference
  on Artificial Intelligence}, vol.~25, no.~1, 2011.

\bibitem{22}
Z.~Zhang and S.~Aeron, ``Exact tensor completion using t-{SVD},'' \emph{IEEE
  Transactions on Signal Processing}, vol.~65, no.~6, pp. 1511--1526, 2016.

\bibitem{37}
C.~Lu, J.~Feng, Z.~Lin, and S.~Yan, ``Exact low tubal rank tensor recovery from
  gaussian measurements,'' \emph{arXiv preprint arXiv:1806.02511}, 2018.

\bibitem{shang2017bilinear}
F.~Shang, J.~Cheng, Y.~Liu, Z.-Q. Luo, and Z.~Lin, ``Bilinear factor matrix
  norm minimization for robust {PCA}: Algorithms and applications,'' \emph{IEEE
  Transactions on Pattern Analysis and Machine Intelligence}, vol.~40, no.~9,
  pp. 2066--2080, 2017.

\bibitem{29}
P.~Zhou, C.~Lu, Z.~Lin, and C.~Zhang, ``Tensor factorization for low-rank
  tensor completion,'' \emph{IEEE Transactions on Image Processing}, vol.~27,
  no.~3, pp. 1152--1163, 2017.

\bibitem{wang2021generalized}
H.~Wang, F.~Zhang, J.~Wang, T.~Huang, J.~Huang, and X.~Liu, ``Generalized
  nonconvex approach for low-tubal-rank tensor recovery,'' \emph{IEEE
  Transactions on Neural Networks and Learning Systems}, 2021.

\bibitem{8913528}
Y.~Su, X.~Wu, and G.~Liu, ``Nonconvex low tubal rank tensor minimization,''
  \emph{IEEE Access}, vol.~7, pp. 170\,831--170\,843, 2019.

\bibitem{Zhang2022Tensor}
X.~Zhang, J.~Zheng, L.~Zhao, Z.~Zhou, and Z.~Lin, ``Tensor recovery with
  weighted tensor average rank,'' \emph{IEEE Transactions on Neural Networks
  and Learning Systems}, pp. 1--15, 2022.

\bibitem{18}
M.~E. Kilmer and C.~D. Martin, ``Factorization strategies for third-order
  tensors,'' \emph{Linear Algebra and its Applications}, vol. 435, no.~3, pp.
  641--658, 2011.

\bibitem{kilmer2013third}
M.~E. Kilmer, K.~Braman, N.~Hao, and R.~C. Hoover, ``Third-order tensors as
  operators on matrices: A theoretical and computational framework with
  applications in imaging,'' \emph{SIAM Journal on Matrix Analysis and
  Applications}, vol.~34, no.~1, pp. 148--172, 2013.

\bibitem{21}
Z.~Zhang, G.~Ely, S.~Aeron, N.~Hao, and M.~Kilmer, ``Novel methods for
  multilinear data completion and de-noising based on tensor-{SVD},'' in
  \emph{Proceedings of the IEEE Conference on Computer Vision and Pattern
  Recognition}, 2014, pp. 3842--3849.

\bibitem{jiang2020multi}
T.-X. Jiang, T.-Z. Huang, X.-L. Zhao, and L.-J. Deng, ``Multi-dimensional
  imaging data recovery via minimizing the partial sum of tubal nuclear norm,''
  \emph{Journal of Computational and Applied Mathematics}, vol. 372, p. 112680,
  2020.

\bibitem{kong2018t}
H.~Kong, X.~Xie, and Z.~Lin, ``t-schatten-$ p $ norm for low-rank tensor
  recovery,'' \emph{IEEE Journal of Selected Topics in Signal Processing},
  vol.~12, no.~6, pp. 1405--1419, 2018.

\bibitem{xu2019laplace}
W.-H. Xu, X.-L. Zhao, T.-Y. Ji, J.-Q. Miao, T.-H. Ma, S.~Wang, and T.-Z. Huang,
  ``Laplace function based nonconvex surrogate for low-rank tensor
  completion,'' \emph{Signal Processing: Image Communication}, vol.~73, pp.
  62--69, 2019.

\bibitem{shi2021robust}
Q.~Shi, Y.-M. Cheung, and J.~Lou, ``Robust tensor svd and recovery with rank
  estimation,'' \emph{IEEE Transactions on Cybernetics}, 2021.

\bibitem{lin2010augmented}
Z.~Lin, M.~Chen, and Y.~Ma, ``The augmented lagrange multiplier method for
  exact recovery of corrupted low-rank matrices,'' \emph{arXiv preprint
  arXiv:1009.5055}, 2010.

\bibitem{vershynin2010introduction}
R.~Vershynin, ``Introduction to the non-asymptotic analysis of random
  matrices,'' \emph{arXiv preprint arXiv:1011.3027}, 2010.

\bibitem{36}
E.~J. Cand{\`e}s and B.~Recht, ``Exact matrix completion via convex
  optimization,'' \emph{Foundations of Computational Mathematics}, vol.~9,
  no.~6, pp. 717--772, 2009.

\bibitem{sun2020nonlocal}
M.~Sun, L.~Zhao, J.~Zheng, and J.~Xu, ``A nonlocal denoising framework based on
  tensor robust principal component analysis with lp norm,'' in \emph{2020 IEEE
  International Conference on Big Data}, 2020, pp. 3333--3340.

\bibitem{32}
D.~Martin, C.~Fowlkes, D.~Tal, and J.~Malik, ``A database of human segmented
  natural images and its application to evaluating segmentation algorithms and
  measuring ecological statistics,'' in \emph{Proceedings Eighth IEEE
  International Conference on Computer Vision.}, vol.~2, 2001, pp. 416--423.

\bibitem{Xia_2018_CVPR}
G.-S. Xia, X.~Bai, J.~Ding, Z.~Zhu, S.~Belongie, J.~Luo, M.~Datcu, M.~Pelillo,
  and L.~Zhang, ``Dota: A large-scale dataset for object detection in aerial
  images,'' in \emph{Proceedings of the IEEE Conference on Computer Vision and
  Pattern Recognition}, 2018, pp. 3974--3983.

\bibitem{Ding_2019_CVPR}
J.~Ding, N.~Xue, Y.~Long, G.-S. Xia, and Q.~Lu, ``Learning roi transformer for
  oriented object detection in aerial images,'' in \emph{Proceedings of the
  IEEE/CVF Conference on Computer Vision and Pattern Recognition}, 2019, pp.
  2849--2858.

\bibitem{Xie_2016_CVPR}
Q.~Xie, Q.~Zhao, D.~Meng, Z.~Xu, S.~Gu, W.~Zuo, and L.~Zhang, ``Multispectral
  images denoising by intrinsic tensor sparsity regularization,'' in
  \emph{Proceedings of the IEEE Conference on Computer Vision and Pattern
  Recognition}, 2016, pp. 1692--1700.

\bibitem{35}
B.~Arad and O.~Ben-Shahar, ``Sparse recovery of hyperspectral signal from
  natural rgb images,'' in \emph{European Conference on Computer Vision}, 2016,
  pp. 19--34.

\bibitem{39}
L.~Huang, X.~Zhao, and K.~Huang, ``Got-10k: A large high-diversity benchmark
  for generic object tracking in the wild,'' \emph{IEEE Transactions on Pattern
  Analysis and Machine Intelligence}, vol.~43, no.~5, pp. 1562--1577, 2021.

\bibitem{martin2013order}
C.~D. Martin, R.~Shafer, and B.~LaRue, ``An order-p tensor factorization with
  applications in imaging,'' \emph{SIAM Journal on Scientific Computing},
  vol.~35, no.~1, pp. A474--A490, 2013.

\bibitem{lu2019low}
C.~Lu, X.~Peng, and Y.~Wei, ``Low-rank tensor completion with a new tensor
  nuclear norm induced by invertible linear transforms,'' in \emph{Proceedings
  of the IEEE/CVF Conference on Computer Vision and Pattern Recognition}, 2019,
  pp. 5996--6004.

\bibitem{liu2022tensor}
Y.~Liu, J.~Liu, Z.~Long, and C.~Zhu, \emph{Tensor computation for data
  analysis}.\hskip 1em plus 0.5em minus 0.4em\relax Springer, 2022.

\bibitem{wright2009robust}
J.~Wright, A.~Ganesh, S.~Rao, Y.~Peng, and Y.~Ma, ``Robust principal component
  analysis: Exact recovery of corrupted low-rank matrices via convex
  optimization,'' \emph{Advances in neural information processing systems},
  vol.~22, 2009.

\bibitem{zhang2019accurate}
L.~Zhang, W.~Wei, Q.~Shi, C.~Shen, A.~van~den Hengel, and Y.~Zhang, ``Accurate
  tensor completion via adaptive low-rank representation,'' \emph{IEEE
  Transactions on Neural Networks and Learning Systems}, vol.~31, no.~10, pp.
  4170--4184, 2019.

\bibitem{zhang2018exploiting}
L.~Zhang, W.~Wei, C.~Bai, Y.~Gao, and Y.~Zhang, ``Exploiting clustering
  manifold structure for hyperspectral imagery super-resolution,'' \emph{IEEE
  Transactions on Image Processing}, vol.~27, no.~12, pp. 5969--5982, 2018.

\bibitem{liu2010robust}
G.~Liu, Z.~Lin, Y.~Yu \emph{et~al.}, ``Robust subspace segmentation by low-rank
  representation.'' in \emph{Icml}, vol.~1.\hskip 1em plus 0.5em minus
  0.4em\relax Citeseer, 2010, p.~8.

\bibitem{song2018nonlocal}
L.~Song, B.~Du, L.~Zhang, L.~Zhang, J.~Wu, and X.~Li, ``Nonlocal patch based
  t-svd for image inpainting: Algorithm and error analysis,'' in
  \emph{Proceedings of the AAAI Conference on Artificial Intelligence},
  vol.~32, no.~1, 2018.

\bibitem{wang2022non}
W.~Wang, J.~Zheng, L.~Zhao, H.~Chen, and X.~Zhang, ``A non-local tensor
  completion algorithm based on weighted tensor nuclear norm,''
  \emph{Electronics}, vol.~11, no.~19, p. 3250, 2022.

\bibitem{marshall1979inequalities}
A.~W. Marshall, I.~Olkin, and B.~C. Arnold, \emph{Inequalities: theory of
  majorization and its applications}.\hskip 1em plus 0.5em minus 0.4em\relax
  Springer, 1979, vol. 143.

\bibitem{komaroff2008enhancements}
N.~Komaroff, ``Enhancements to the von neumann trace inequality,'' \emph{Linear
  algebra and its applications}, vol. 428, no.~4, pp. 738--741, 2008.

\bibitem{zhang2017matrix}
X.-D. Zhang, \emph{Matrix analysis and applications}.\hskip 1em plus 0.5em
  minus 0.4em\relax Cambridge University Press, 2017.

\bibitem{hu2015new}
W.~Hu, D.~Tao, W.~Zhang, Y.~Xie, and Y.~Yang, ``A new low-rank tensor model for
  video completion,'' \emph{arXiv preprint arXiv:1509.02027}, 2015.

\end{thebibliography}
\bibliographystyle{IEEEtran}

\end{document}